\documentclass[10pt,logo,copyright]{nvidiatechreport}
\usepackage{subcaption}
\usepackage{wrapfig}
\usepackage{makecell}
\usepackage{multirow}
\usepackage{float}
\usepackage{placeins}
\usepackage{fontawesome5}
\newcommand{\frozenbranch}{\textcolor[HTML]{1678C8}{\faIcon[regular]{snowflake}}}
\newcommand{\trainedbranch}{\textcolor[HTML]{DC405B}{\faIcon{fire-alt}}}


\usepackage[numbers, sort]{natbib}
\definecolor{citecolor}{HTML}{0071BC}
\definecolor{linkcolor}{HTML}{ED1C24}
\definecolor{unifiedblue}{HTML}{DCECF8}
\definecolor{unifiedred}{HTML}{FADFE0}
\definecolor{baselinegray}{HTML}{F1F3F6}
\definecolor{undaccent}{HTML}{2E75A8}
\definecolor{undcard}{HTML}{F4F9FD}
\definecolor{genaccent}{HTML}{B9555B}
\definecolor{gencard}{HTML}{FFF5F5}
\definecolor{bestgreen}{HTML}{B8DEBA}
\definecolor{secondgreen}{HTML}{DDEED8}
\newcommand{\bestscore}[1]{\cellcolor{bestgreen}\textbf{#1}}
\newcommand{\secondscore}[1]{\cellcolor{secondgreen}#1}

\usepackage{nicefrac}

\usepackage[capitalize]{cleveref}
\crefname{section}{\S}{\S\S}
\crefname{subsection}{\S}{\S\S}
\crefname{table}{\text{Tab.}}{\text{Tab.}}
\Crefname{table}{Table}{Tables}
\crefname{figure}{\text{Fig.}}{\text{Fig.}}
\Crefname{figure}{Figure}{Figures}
\crefname{equation}{\text{Eq}}{\text{Eq}}

\title{PixelUMM: Encoder-Free Unified Image and Video Understanding and Generation}

\newcommand{\pixelummauthor}[2]{\mbox{#1\textsuperscript{#2}}}
\newcommand{\pixelummauthorblock}{%
  \begingroup
  \raggedright\setlength{\parindent}{0pt}\setlength{\parskip}{0pt}%
  \linespread{1}\selectfont
  {\normalfont\bfseries\fontsize{10}{17}\selectfont
    \pixelummauthor{Cong Wei}{1,2}\hspace{1.2em}%
    \pixelummauthor{Xuanchi Ren}{1}\hspace{1.2em}%
    \pixelummauthor{Bryan Chu}{1}\hspace{1.2em}%
    \pixelummauthor{Weiming Ren}{2}\hspace{1.2em}%
    \pixelummauthor{Huan Ling}{1}\hspace{1.2em}%
    \pixelummauthor{Jiahui Huang}{1}\hspace{1.2em}%
    \pixelummauthor{Laura Leal-Taix\'e}{1}\hspace{1.2em}%
    \pixelummauthor{Sanja Fidler}{1}\hspace{1.2em}%
    \pixelummauthor{Wenhu Chen}{2}\hspace{1.2em}%
    \pixelummauthor{Zian Wang}{1}\hspace{1.2em}%
    \pixelummauthor{Jay Zhangjie Wu}{1}\par}
  \vspace{6pt}
  {\normalfont\fontsize{10}{14}\selectfont
    \textsuperscript{1}NVIDIA \quad \textsuperscript{2}University of Waterloo\par}
  \vspace{4pt}
  {\normalfont\small\url{https://nv-tlabs.github.io/PixelUMM/}\par}
  \endgroup
}
\makeatletter
\renewcommand{\@author}{\pixelummauthorblock}
\makeatother

\begin{document}
\maketitle
\begin{abstract}
Unified Multimodal Models (UMMs) often rely on separate visual representations for understanding and generation, increasing visual context length and complicating integration with established vision-language pretraining pipelines.
Recent advances in pixel-space modeling offer an encoder-free alternative, but extending this paradigm from images to videos is non-trivial: video understanding and generation adopt different temporal representations, leaving the design of a unified visual interface an open question.
We present PixelUMM, an encoder-free model for unified image and video understanding and generation directly in pixel space.
PixelUMM represents images as spatial patches and videos as spatiotemporal tubelets, connecting raw pixels to a shared multimodal backbone through single-layer linear projections.
Its Mixture-of-Transformers architecture combines shared attention with task-specific parameters and extends clean-pixel prediction to video generation, jointly supporting autoregressive text prediction and pixel-space flow matching.
Experiments show that PixelUMM achieves competitive performance across image and video understanding and generation tasks.
We further conduct empirical studies of key design choices, including decoder design and spatial-temporal patch size, providing insights for future pixel-space unified multimodal models.
\end{abstract}

\abscontent

\section{Introduction}
\label{sec:intro}

Unified multimodal models aim to bring visual understanding and generation into a single system, allowing the same model to interpret visual inputs, respond in language, and create visual content~\cite{zhou2024transfusion,deng2025bagel,liu2025tuna,tong2026beyond}.
A central question is how to represent vision for these different tasks.
Models such as BAGEL adopt a dual-encoder interface: a vision Transformer (ViT)~\cite{ViT} provides semantic features for understanding, while a variational autoencoder (VAE) provides reconstruction-oriented latents for generation~\cite{deng2025bagel}.
This design supports both capabilities, but leaves them connected to the backbone through different visual representations.

A VAE is not intrinsically required for visual synthesis.
Representation autoencoders (RAEs), for example, replace VAE encoders with pretrained semantic encoders and learned decoders, supporting strong image and text-to-image generation~\cite{zheng2025diffusion,tong2026scalingtexttoimagediffusiontransformers}.
For unified models, however, an important motivation for retaining the VAE is to support editing and reference-based tasks, which often require preserving fine-grained input details.
In these tasks, the generated output must remain faithful to the source or reference image, particularly in regions that should remain unchanged during editing.
The VAE supplies reconstruction-oriented features alongside the ViT's semantic features, allowing the model to retain both kinds of information, as in BAGEL~\cite{deng2025bagel}.

\clearpage
\begin{figure}[H]
  \centering
  \includegraphics[width=\linewidth,height=0.95\textheight,keepaspectratio]{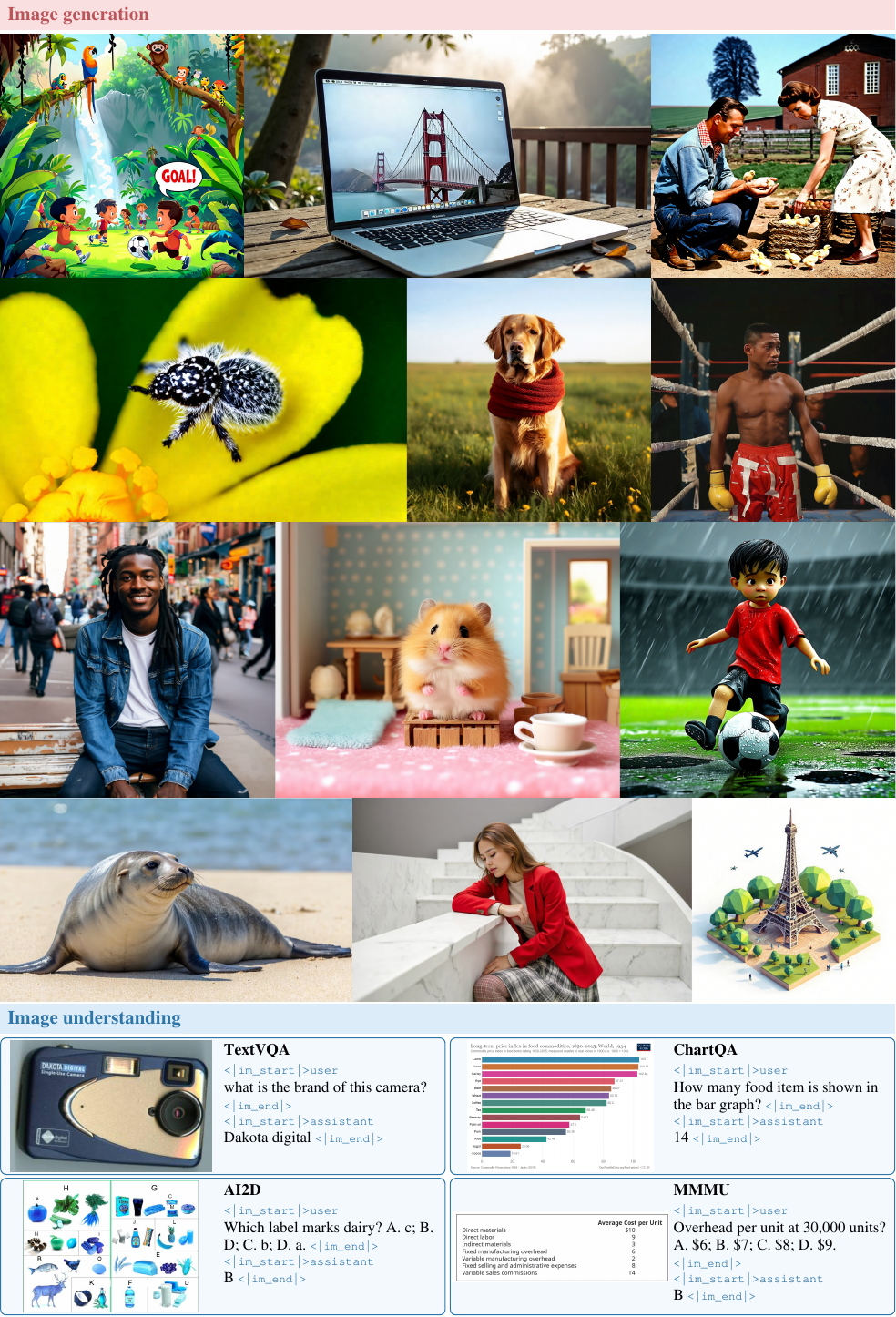}
  \caption{\textbf{Image generation and understanding with PixelUMM.}
  Top: PixelUMM text-to-image generation results. Bottom: benchmark inputs and raw model answers from TextVQA, ChartQA, AI2D, and MMMU.}
  \label{fig:teaser}
\end{figure}
\clearpage

\begin{figure}[H]
  \centering
  \includegraphics[width=\linewidth,height=0.95\textheight,keepaspectratio]{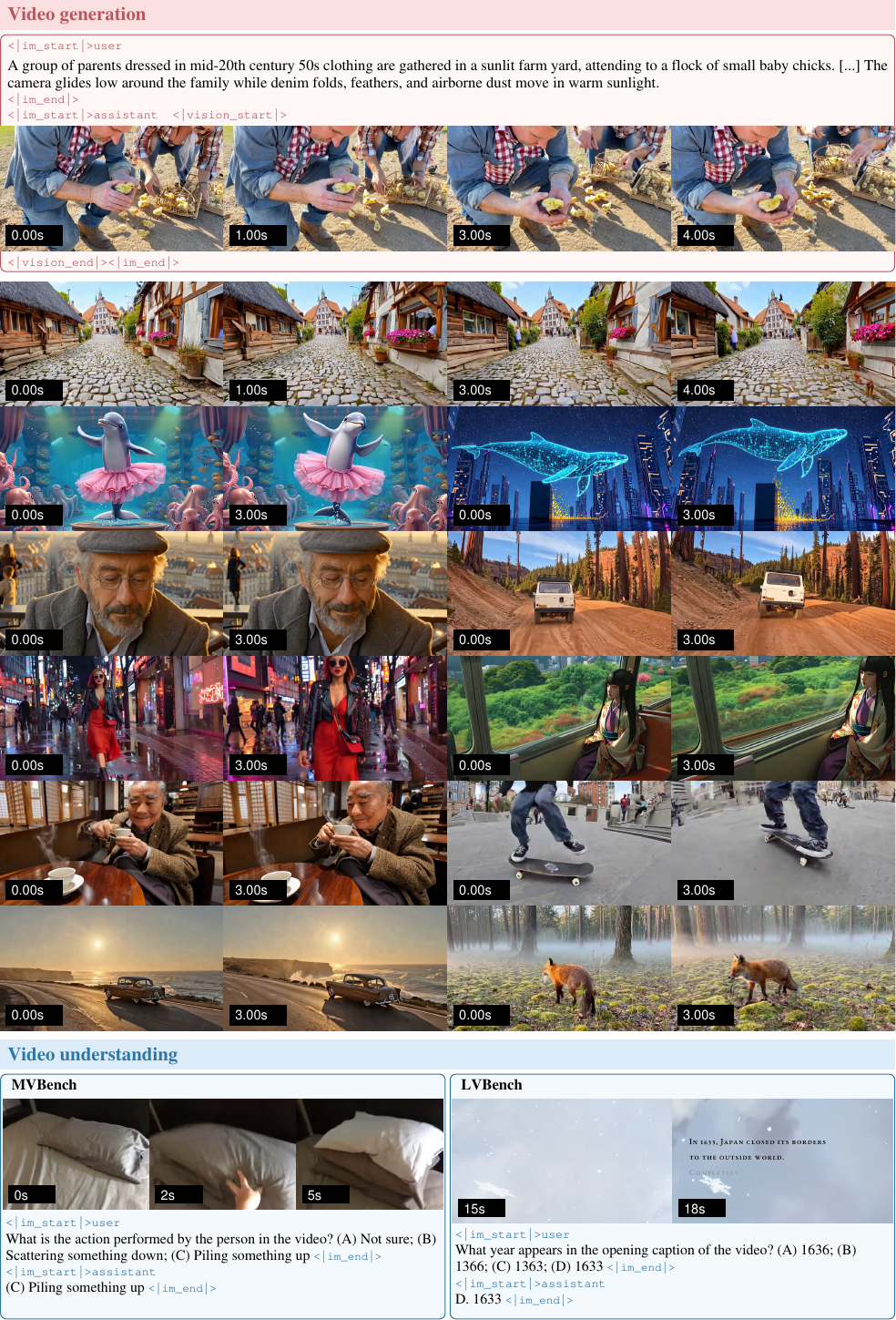}
  \caption{\textbf{Video generation and understanding with PixelUMM.}
  Top: PixelUMM text-to-video generation results. Bottom: PixelUMM responses on MVBench and LVBench.}
  \label{fig:t2v_teaser}
\end{figure}
\clearpage

Maintaining two visual interfaces nevertheless creates a practical burden.
Representing each conditioning image with both ViT and VAE tokens approximately doubles the visual context compared with a single visual stream of similar token resolution, increasing attention and memory costs.
This duplication is particularly undesirable for long-context processing and multi-turn multimodal conversations, where visual tokens accumulate across images, video frames, and dialogue turns.
This dual-stream interface also differs from the single visual stream typically used in vision-language model (VLM) pretraining.
Adding VAE tokens would require substantial changes to existing data pipelines and training recipes.
Requiring these changes at the pretraining stage makes unification less practical.

Recent progress in pixel-space generation offers a route around the separate VAE interface.
JiT~\cite{li2025back} and PixelDiT~\cite{yu2025pixeldit} show that image generation can operate directly on pixels without a pretrained latent autoencoder.
Unlike replacing a VAE with another autoencoder, this approach removes the latent encoding stage itself.
For unified models, it opens the possibility of using raw pixels as the common visual input, rather than carrying both semantic and reconstruction-oriented encodings of the same image.
TUNA-2~\cite{tuna2}, SenseNova-U1~\cite{diao2026sensenovau1}, and SenseNova-U1.5~\cite{diao2026sensenovau15nativeunifiedvisual} develop this direction for unified image understanding and generation with encoder-free, pixel-space interfaces.

In this work, we examine how this paradigm extends to video.
Specifically, we ask whether JiT-style clean-pixel prediction can support video generation, and whether an encoder-free pixel-space model can jointly support image and video understanding and generation.
Extending this paradigm to video is non-trivial, as it introduces new choices in the visual interface.
Video generation often relies on causal 3D VAEs that encode the first frame separately from subsequent frames~\cite{wan2025wan}, whereas video understanding commonly processes sampled frames independently with an image encoder~\cite{cheng2024videollama}.
These different conventions do not naturally yield a shared interface for understanding and generation.
In this work, we redesign the input embedders and output decoders for images and videos to enable a shared pixel-space interface.

We present \textbf{PixelUMM}, an encoder-free model that unifies image and video understanding and generation directly in pixel space.
Images are represented as spatial patches and videos as spatiotemporal tubelets, connected to the backbone through single-layer linear projections.
A Mixture-of-Transformers architecture combines shared multimodal attention with separate understanding and generation parameters.
The model learns autoregressive text prediction and pixel-space flow matching jointly, using clean visual inputs for understanding and noisy visual inputs for generation.
A staged training recipe progressively incorporates image and video tasks and higher spatial resolutions.

Our evaluation shows that PixelUMM achieves competitive performance across image and video understanding and generation tasks.
Alongside these evaluations, we conduct controlled empirical studies of image and video patch sizes and alternative video decoder designs, examining training behavior and spatiotemporal patch artifacts.
These studies provide practical insights into the design of future pixel-space unified multimodal models.

\section{Method}
\label{sec:method}

Extending encoder-free pixel-space modeling to unified image and video
understanding and generation requires an explicit visual interface, rather
than inheriting one from a pretrained vision encoder or video VAE.
We describe PixelUMM's interface and backbone below, followed by its
multimodal sequence construction and joint text and pixel-space objectives.
We examine patch size and decoder alternatives empirically in
Sec.~\ref{sec:findings}.

\subsection{Architecture}
\label{sec:architecture}

\begin{figure}[t]
  \centering
  \includegraphics[width=\linewidth]{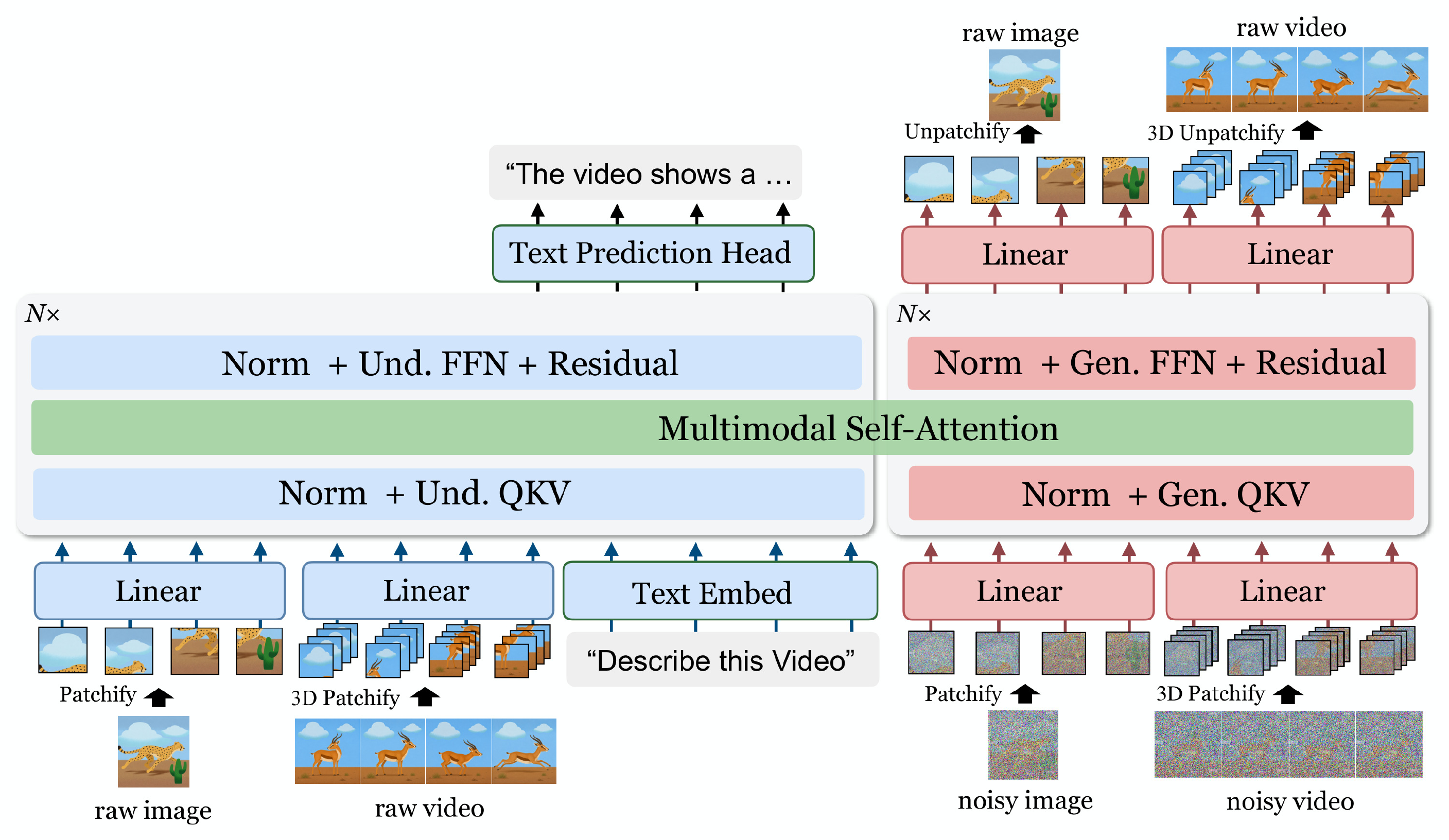}
  \vspace{-1 em}
  \caption{\textbf{PixelUMM architecture.}
  PixelUMM is a native unified multimodal model for image and video understanding and generation.
  It operates directly in pixel space without pretrained visual encoders or VAE-based latent tokenizers.
  PixelUMM adopts a Mixture-of-Transformers (MoT) architecture comprising an understanding expert for text prediction and visual understanding and a generation expert for image and video synthesis.
  We omit explicit timestep embeddings and timestep-conditioned normalization.
  The denoising network receives the noisy pixels without a separate timestep input.
  The two experts have symmetric structures with expert-specific normalization layers, projections, and FFNs, while all tokens interact through shared multimodal self-attention in every Transformer block.}
  \label{fig:architecture}
\end{figure}

\noindent\textbf{Native pixel interface.}
Let an image be $\mathbf{x}^{\mathrm{img}}\in\mathbb{R}^{H\times W\times3}$
and a video be $\mathbf{x}^{\mathrm{video}}\in\mathbb{R}^{F\times H\times W\times3}$.
We partition an image into non-overlapping $p\times p$ patches and a video into
$\tau\times p\times p$ tubelets.  In other words, images undergo 2D patchify,
whereas videos undergo 3D patchify jointly over time, height, and width.
PixelUMM uses $p=16$ and $\tau=4$ for video processing.
Each flattened
patch or tubelet is mapped to the Transformer hidden size $d$ by exactly one
simple linear layer:
\begin{equation}
\begin{aligned}
  \mathbf{x}^{\mathrm{img}}
  &\xrightarrow{\;\operatorname{2D\ Patchify}_{16\times16}\;}
  \mathbb{R}^{N_{\mathrm{img}}\times(16\cdot16\cdot3)}
  \xrightarrow[\texttt{img\_gen\_linear\_proj}]{\texttt{img\_und\_linear\_proj}}
  \mathbb{R}^{N_{\mathrm{img}}\times d},\\[6pt]
  \mathbf{x}^{\mathrm{video}}
  &\xrightarrow{\;\operatorname{3D\ Patchify}_{4\times16\times16}\;}
  \mathbb{R}^{N_{\mathrm{video}}\times(4\cdot16\cdot16\cdot3)}
  \xrightarrow[\texttt{video\_gen\_linear\_proj}]{\texttt{video\_und\_linear\_proj}}
  \mathbb{R}^{N_{\mathrm{video}}\times d}.
\end{aligned}
  \label{eq:pixel_embedding}
\end{equation}
The labels above and below each projection arrow denote the understanding and
generation alternatives, respectively.  Each is a single linear layer with
its own parameters.
Clean images and videos enter through the understanding projections, whereas
corrupted generation targets enter through the generation projections.

At the output, each modality uses a simple linear head, implemented as RMSNorm
followed by a linear projection in the reverse direction:
\begin{equation}
  \mathbb{R}^{d}
  \xrightarrow{\;\operatorname{RMSNorm}+\texttt{img\_linear\_outproj}\;}
  \mathbb{R}^{16\times16\times3},
  \qquad
  \mathbb{R}^{d}
  \xrightarrow{\;\operatorname{RMSNorm}+\texttt{video\_linear\_outproj}\;}
  \mathbb{R}^{4\times16\times16\times3}.
  \label{eq:pixel_decoding}
\end{equation}
Image/video unpatchify then rearranges these predictions into pixels.  The
output projections are zero-initialized before multimodal training.
Consequently, PixelUMM has no vision encoder (VE), variational autoencoder
(VAE), or discrete visual tokenizer: the only transformations between pixels
and the backbone are one-layer linear projections and deterministic
patchify/unpatchify operations.

\Needspace{5\baselineskip}
\noindent\textbf{No explicit timestep conditioning.}
In contrast to BAGEL~\cite{deng2025bagel} and
SenseNova-U1~\cite{diao2026sensenovau1}, PixelUMM omits explicit timestep
embeddings and timestep-conditioned AdaLN, as in MiniT2I~\cite{minit2i2026}.
The denoising network receives noisy pixels without a separate timestep input,
allowing it to infer the corruption level from them. The timestep is still used
to construct training targets and guide the sampling process.

\noindent\textbf{Video understanding modes.}
We use two modes according to the available temporal sampling rate.
In \texttt{dense\_mode}, for input videos at least $4$ FPS, we first apply 3D patchify and
then map each tubelet through \texttt{video\_und\_linear\_proj}, the dedicated linear
video-understanding projection.  We sample at $4$ FPS by default, but the
interface also supports a higher configured sampling rate.
In \texttt{sparse\_mode}, used for input videos below $4$ FPS, we instead sample
at $1$ FPS and encode each frame independently with
\texttt{img\_und\_linear\_proj}.  This avoids duplicating
low-rate frames merely to fill a tubelet; both paths retain temporal order
through the multimodal sequence described in Sec.~\ref{sec:sequence_modeling}.

\noindent\textbf{Mixture-of-Transformers backbone.}
PixelUMM is initialized from a decoder-only Transformer from Qwen3~\cite{yang2025qwen3}
and uses token-level hard routing between understanding and generation
experts.
Each block has route-specific normalization, QKV/output projections,
and FFNs, while all tokens participate in the same multimodal self-attention
operation (Fig.~\ref{fig:architecture}).  Text and clean visual tokens use the
understanding route; noisy visual tokens use the generation route.

\subsection{Unified Multimodal Sequence Modeling}
\label{sec:sequence_modeling}

\noindent\textbf{Conversation serialization.}
We serialize all four tasks in ChatML, as illustrated in
Fig.~\ref{fig:chatml_formats}.  Visual inputs and generation targets occupy
raw-pixel token spans within the conversation.

\begin{figure}[!htb]
\centering
\begin{minipage}[t]{0.493\linewidth}
\begin{mdframed}[
  backgroundcolor=undcard,
  linecolor=undaccent,
  linewidth=0.8pt,
  roundcorner=3pt,
  innerleftmargin=5pt,
  innerrightmargin=5pt,
  innertopmargin=3pt,
  innerbottommargin=3pt,
  skipabove=0pt,
  skipbelow=0pt]
{\fontsize{11}{12}\selectfont\textcolor{undaccent}{\textbf{I2T}}}\hfill
{\fontsize{9}{10}\selectfont\textcolor{black!55}{Image Understanding}}\par\vspace{-4pt}
{\color{undaccent}\rule{\linewidth}{0.45pt}}\vspace{1pt}
{\color{black!88}\ttfamily\fontsize{8.8}{10.0}\selectfont
\detokenize{<|im_start|>system}\par
\detokenize{You are a helpful assistant.<|im_end|>}\par
\detokenize{<|im_start|>user}\par
\detokenize{<|vision_start|>}\par
\detokenize{<image_or_video_frame_patches x777>}\par
\detokenize{<|vision_end|>}\par
\detokenize{Describe this image in detail.<|im_end|>}\par
\detokenize{<|im_start|>assistant}
\par\vspace{38.8pt}}
\end{mdframed}
\end{minipage}\hfill
\begin{minipage}[t]{0.493\linewidth}
\begin{mdframed}[
  backgroundcolor=undcard,
  linecolor=undaccent,
  linewidth=0.8pt,
  roundcorner=3pt,
  innerleftmargin=5pt,
  innerrightmargin=5pt,
  innertopmargin=3pt,
  innerbottommargin=3pt,
  skipabove=0pt,
  skipbelow=0pt]
{\fontsize{11}{12}\selectfont\textcolor{undaccent}{\textbf{V2T}}}\hfill
{\fontsize{9}{10}\selectfont\textcolor{black!55}{Video Understanding}}\par\vspace{-4pt}
{\color{undaccent}\rule{\linewidth}{0.45pt}}\vspace{1pt}
{\color{black!88}\ttfamily\fontsize{8.8}{10.0}\selectfont
\detokenize{<|im_start|>system}\par
\detokenize{You are a helpful assistant.<|im_end|>}\par
\detokenize{<|im_start|>user}\par
\detokenize{<0.0 seconds><|vision_start|>}\par
\detokenize{<video_tube_patches x777><|vision_end|>}\par
\detokenize{<1.0 seconds><|vision_start|>}\par
\detokenize{<video_tube_patches x777><|vision_end|>}\par
\detokenize{<2.0 seconds><|vision_start|>}\par
\detokenize{<video_tube_patches x777><|vision_end|>}\par
\detokenize{Describe this video in details.<|im_end|>}\par
\detokenize{<|im_start|>assistant}
}
\end{mdframed}
\end{minipage}

\vspace{5pt}

\begin{minipage}[t]{0.493\linewidth}
\begin{mdframed}[
  backgroundcolor=gencard,
  linecolor=genaccent,
  linewidth=0.8pt,
  roundcorner=3pt,
  innerleftmargin=5pt,
  innerrightmargin=5pt,
  innertopmargin=3pt,
  innerbottommargin=3pt,
  skipabove=0pt,
  skipbelow=0pt]
{\fontsize{11}{12}\selectfont\textcolor{genaccent}{\textbf{T2I}}}\hfill
{\fontsize{9}{10}\selectfont\textcolor{black!55}{Image Generation}}\par\vspace{-4pt}
{\color{genaccent}\rule{\linewidth}{0.45pt}}\vspace{1pt}
{\color{black!88}\ttfamily\fontsize{8.8}{10.0}\selectfont
\detokenize{<|im_start|>system}\par
\detokenize{You are a helpful assistant.<|im_end|>}\par
\detokenize{<|im_start|>user}\par
\detokenize{Generate a high-quality image based on the}\par
\detokenize{following description:}\par
\detokenize{A cheetah runs across a desert.<|im_end|>}\par
\detokenize{<|im_start|>assistant}\par
\detokenize{<|vision_start|>}\par
\detokenize{<noisy_image_patches x1024>}\par
\detokenize{<|vision_end|>}
}
\end{mdframed}
\end{minipage}\hfill
\begin{minipage}[t]{0.493\linewidth}
\begin{mdframed}[
  backgroundcolor=gencard,
  linecolor=genaccent,
  linewidth=0.8pt,
  roundcorner=3pt,
  innerleftmargin=5pt,
  innerrightmargin=5pt,
  innertopmargin=3pt,
  innerbottommargin=3pt,
  skipabove=0pt,
  skipbelow=0pt]
{\fontsize{11}{12}\selectfont\textcolor{genaccent}{\textbf{T2V}}}\hfill
{\fontsize{9}{10}\selectfont\textcolor{black!55}{Video Generation}}\par\vspace{-4pt}
{\color{genaccent}\rule{\linewidth}{0.45pt}}\vspace{1pt}
{\color{black!88}\ttfamily\fontsize{8.8}{10.0}\selectfont
\detokenize{<|im_start|>system}\par
\detokenize{You are a helpful assistant.<|im_end|>}\par
\detokenize{<|im_start|>user}\par
\detokenize{Generate a high-quality video based on the}\par
\detokenize{following description:}\par
\detokenize{A cheetah chases a gazelle.<|im_end|>}\par
\detokenize{<|im_start|>assistant}\par
\detokenize{<|vision_start|>}\par
\detokenize{<noisy_video_patches x6144>}\par
\detokenize{<|vision_end|>}
}
\end{mdframed}
\end{minipage}
\caption{\textbf{ChatML sequence formats for the four core tasks.}
Top: image understanding (I2T) and video understanding (V2T).
Bottom: text-to-image (T2I) and text-to-video (T2V) generation.
Understanding tasks use clean visual inputs; generation tasks use noisy targets.
Patch placeholders denote raw-pixel token spans, with illustrative token counts.
Timestamps indicate the temporal positions of video-understanding blocks.}
\label{fig:chatml_formats}
\end{figure}

\begin{figure}[!t]
  \centering
  \includegraphics[width=\linewidth]{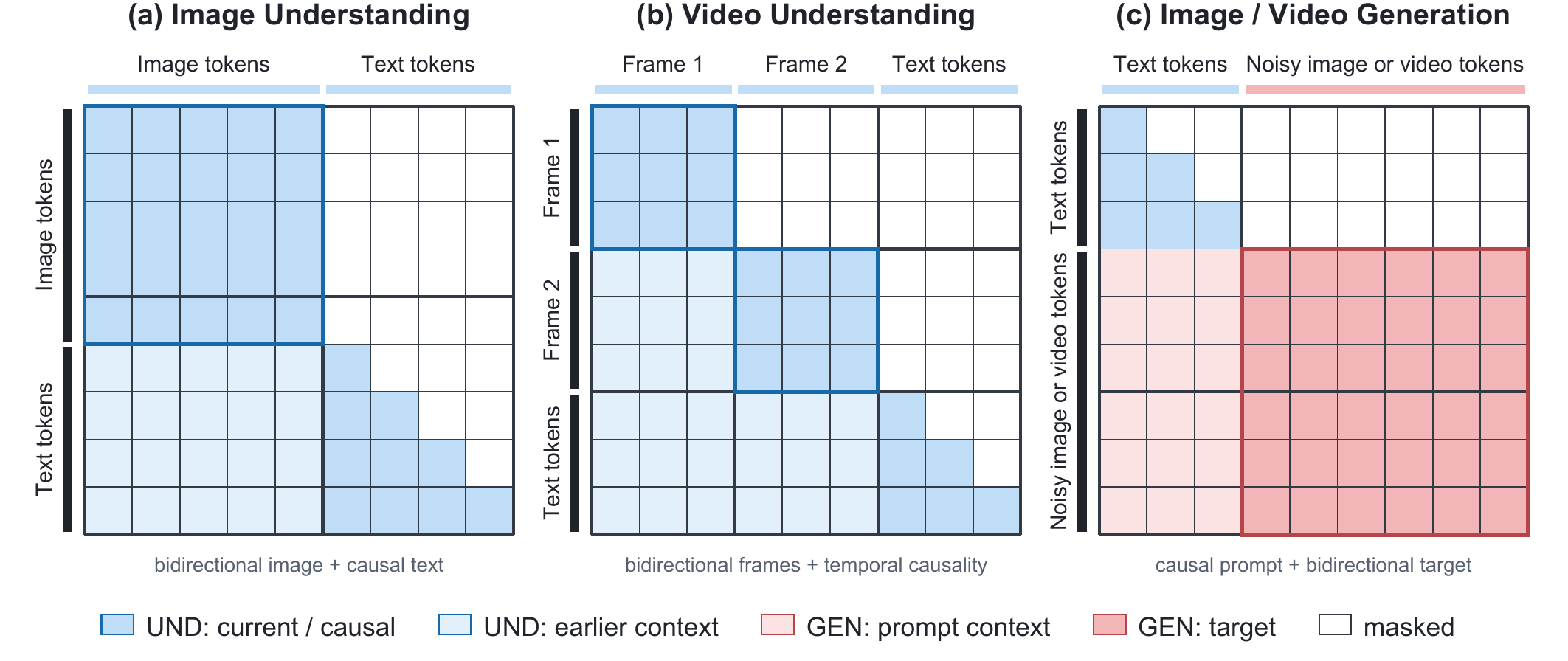}
  \caption{\textbf{Attention patterns used by PixelUMM.}
  Filled cells indicate visible query--key pairs (rows are queries and columns are keys).
  Image tokens form one bidirectional block, while sparse video frames form temporally ordered bidirectional blocks, so later frames can access earlier frames.
  Understanding text is causal and can attend to all preceding visual context.
  For generation, the noisy target attends to the causal prompt and all target tokens.}
  \label{fig:attention_masks}
\end{figure}

\noindent\textbf{Generalized causal attention.}
Following the blockwise view of BAGEL~\cite{deng2025bagel}, we partition the
packed conversation into consecutive text and visual blocks.  A block may
attend to all preceding blocks.  Attention is causal inside text, but
bidirectional inside each visual block.  An image therefore forms one
bidirectional island.  Sparse video frames form temporally ordered islands, so
a later frame can attend to earlier frames but an earlier frame cannot access a
future one; a dense video segment can instead form one tubelet block.  For
generation, the noisy target attends to the causal prompt and bidirectionally
within the complete target block, while preceding clean context cannot attend
back into that target.  Figure~\ref{fig:attention_masks} visualizes these three
cases.  Unlike BAGEL, the sequence contains neither VAE latents nor ViT
features---only text tokens and clean or noisy raw-pixel tokens.

\noindent\textbf{Positional encoding.}
Following the three-axis Native RoPE design of
NEO~\cite{Diao2025NEO} and SenseNova-U1~\cite{diao2026sensenovau1}, we
allocate half of each attention head to the temporal axis and one quarter
each to height and width. Text tokens advance only along the temporal axis
($H=W=0$), while visual tokens also carry spatial grid coordinates. We use
$\theta_H=\theta_W=10^4$ and retain Qwen3's RoPE base of $\theta_T=10^6$
for the temporal axis~\cite{yang2025qwen3}.

\noindent\textbf{Attention implementation.}
For image and video understanding, we implement the generalized causal mask
with FlexAttention~\cite{flexattention}.
For text-to-image and text-to-video generation, we use \texttt{two\_way}
attention with two variable-length FlashAttention calls~\cite{dao2022flashattention,dao2023flashattention2}:
a causal pass processes the text prompt, and a non-causal pass lets noisy
visual tokens attend to both the prompt and the complete target block.
Both implementations preserve the attention patterns in
Fig.~\ref{fig:attention_masks} and isolate samples within a packed sequence.

\subsection{Objectives}
\label{sec:objectives}

\noindent\textbf{Text objective.}
Following BAGEL~\cite{deng2025bagel}, we apply cross-entropy only to assistant
tokens, including the end-of-turn
token, with square-root sequence-length normalization across responses.

\noindent\textbf{Pixel objective.}
Following JiT~\cite{li2025back}, a clean patch or tubelet
$\mathbf{x}$ is corrupted as $\mathbf{z}_t=(1-t)\mathbf{x}
+t\boldsymbol{\epsilon}$, where $t\in[0,1]$ and
$\boldsymbol{\epsilon}\sim\mathcal{N}(\mathbf{0},\mathbf{I})$.  The head
predicts clean pixels $\widehat{\mathbf{x}}_\theta$, while the loss is computed
in velocity space using $\mathbf{v}_\theta=(\mathbf{z}_t-
\widehat{\mathbf{x}}_\theta)/\bar t$ and $\mathbf{v}^{\star}=(\mathbf{z}_t-
\mathbf{x})/\bar t$, with $\bar t=\max(t,0.05)$.  Squared velocity error is
averaged over active pixels, active visual tokens, and then media items,
yielding $\mathcal{L}_{\mathrm{img}}$ and $\mathcal{L}_{\mathrm{vid}}$.

\noindent\textbf{Joint objective.}
The total loss is $\mathcal{L}=\lambda_{\mathrm{CE}}\mathcal{L}_{\mathrm{CE}}
+\lambda_{\mathrm{img}}\mathcal{L}_{\mathrm{img}}
+\lambda_{\mathrm{vid}}\mathcal{L}_{\mathrm{vid}}$.
We combine the text, image, and video losses with stage-specific weights
listed in Table~\ref{tab:training_recipe}.
Only loss terms relevant to each sample are active.

\subsection{Training Details}
\label{sec:exp_setup}

\begin{table}[!t]
\centering
\small
\setlength{\tabcolsep}{4.5pt}
\renewcommand{\arraystretch}{1.15}
\resizebox{\textwidth}{!}{
\begin{tabular}{l|c|c|c|c|c|c}
\toprule
& \makecell{\textbf{Joint}\\\textbf{Stage 1}}
& \makecell{\textbf{Gen}\\\textbf{Stage 1}}
& \makecell{\textbf{Und}\\\textbf{Stage 1}}
& \makecell{\textbf{Und}\\\textbf{Stage 2}}
& \makecell{\textbf{Und}\\\textbf{Stage 3}}
& \makecell{\textbf{Joint}\\\textbf{Stage 2}} \\
\midrule
\multicolumn{7}{l}{\textbf{Module settings}\quad
  {\footnotesize\frozenbranch\ Frozen\qquad\trainedbranch\ Trainable}} \\
Understanding branch
& \trainedbranch & \frozenbranch & \trainedbranch & \trainedbranch & \trainedbranch & \trainedbranch \\
Generation branch
& \trainedbranch & \trainedbranch & \frozenbranch & \frozenbranch & \frozenbranch & \trainedbranch \\
\midrule
\multicolumn{7}{l}{\textbf{Hyperparameters}} \\
Learning rate
& $1\times10^{-4}$ & $1\times10^{-4}$
& $2\times10^{-5}$ & $2\times10^{-5}$ & $2\times10^{-5}$ & $2\times10^{-5}$ \\
LR scheduler
& Constant & Constant & Constant & Constant & Constant & Constant \\
Optimizer
& \multicolumn{6}{c}{AdamW ($\beta_1=0.9,\; \beta_2=0.95,\; \epsilon=10^{-8}$)} \\
Weight decay
& 0.0 & 0.0 & 0.0 & 0.0 & 0.0 & 0.0 \\
Gradient norm clip
& 0.2 & 0.2 & 0.2 & 0.2 & 0.2 & 0.2 \\
Training steps
& 150K & 200K & 100K & 20K & 15K & 20K \\
Warmup steps
& 1000 & 1000 & 1000 & 1000 & 1000 & 1000 \\
\makecell[l]{Loss weight\\(CE : Image MSE : Video MSE)}
& $1:10:0$ & $0:10:30$ & $1:0:0$ & $1:0:0$ & $1:0:0$ & $1:10:30$ \\
Und resolution (image)
& $256^2$ & -- & Native & Native & Native & Native \\
Und resolution (video)
& -- & -- & -- & $224^2$ & $448^2$ & $448^2$ \\
Gen resolution (image)
& $256^2$ & $256^2$ & -- & -- & -- & $512^2$ \\
Gen resolution (video)
& -- & $256^2$ & -- & -- & -- & $512^2$ \\
Gen duration (video)
& \multicolumn{6}{c}{4.0s / 24fps / 96 frames \quad (video generation only)} \\
Seq length
& 60K & 60K & 60K & 80K & 80K & 80K \\
Time shift
& \multicolumn{6}{c}{$s(H,W)=\sqrt{HW/(256\times256)}$ \quad (generation only)} \\
\midrule
\multicolumn{7}{l}{\textbf{Per-step example ratio (unnormalized)}} \\
Text only
& 0.5 & \textcolor{black!40}{\textemdash} & 1 & 1 & 1 & 1 \\
Image understanding (I2T)
& 1 & \textcolor{black!40}{\textemdash} & 10 & 5 & 5 & 5 \\
Image generation (T2I)
& 4 & 1 & \textcolor{black!40}{\textemdash} & \textcolor{black!40}{\textemdash} & \textcolor{black!40}{\textemdash} & 11 \\
Video understanding (V2T)
& \textcolor{black!40}{\textemdash} & \textcolor{black!40}{\textemdash} & \textcolor{black!40}{\textemdash} & 5 & 5 & 5 \\
Video generation (T2V)
& \textcolor{black!40}{\textemdash} & 3 & \textcolor{black!40}{\textemdash} & \textcolor{black!40}{\textemdash} & \textcolor{black!40}{\textemdash} & 33 \\
\bottomrule
\end{tabular}
}
\caption{\textbf{Training recipe of PixelUMM.} Stages are listed in training order. Gen/Und denote generation/understanding. Native denotes native-resolution images, and video resolutions are per frame. CE denotes cross-entropy; zero loss weights indicate inactive objectives. Per-step example ratios specify the relative numbers of training examples consumed by each task in a step and need not sum to one; a gray dash indicates an inactive task (zero ratio). Time shift applies only to generation tasks, with $H$ and $W$ denoting spatial dimensions. Elsewhere, dashes indicate inapplicable settings.}
\label{tab:training_recipe}
\end{table}

\noindent\textbf{Training stages.}
We progressively incorporate image and video tasks through six
training stages (Table~\ref{tab:training_recipe}).
Joint Stage~1 trains text-only tasks, image understanding (I2T), and image generation (T2I) for 150K steps at an image resolution of $256\times256$.
Gen Stage~1 then trains T2I and text-to-video generation (T2V) for 200K steps at $256\times256$.
Und Stage~1 trains text-only tasks and native-resolution image understanding for 100K steps.
Und Stages~2 and~3 retain these tasks and add video understanding at $224\times224$ for 20K steps and $448\times448$ for 15K steps, respectively.
Finally, Joint Stage~2 trains all tasks together for 20K steps: T2I and T2V at $512\times512$, text-only tasks, native-resolution image understanding, and video understanding at $448\times448$.
Video resolutions specify the spatial resolution of each frame.

\Needspace{4\baselineskip}
\noindent\textbf{Datasets.}
We draw image-understanding data from FineVision~\cite{wiedmann2025finevision} and video-understanding data from the VideoChat-Flash training collection~\cite{li2024videochat} and LLaVA-Video-178K~\cite{zhang2024video}.
For image and video generation, we collect public real and synthetic data~\cite{huang2025vipe}.
\section{Experiment}
\label{sec:experiments}

\subsection{Empirical Findings}
\label{sec:findings}

We label the eight experimental families F1--F8 and the configurations
within each family R01, R02, and so on. For example, F1-R01 denotes the first
configuration in family F1. The same identifiers are used in~the~text~and~figures.

\Needspace{0.22\textheight}
\subsubsection{F1: Image Patch Size}
\begin{figure}[!htb]
  \centering
  \includegraphics[width=0.90\linewidth]{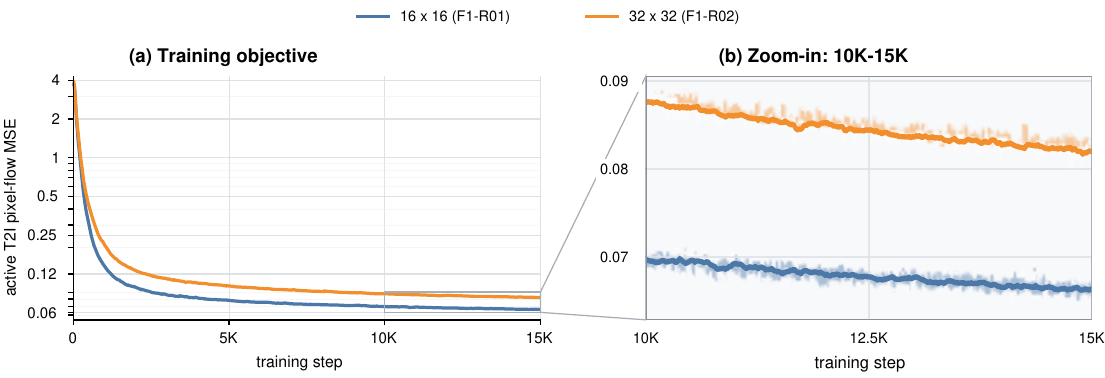}
  \caption{\textbf{Image patch-size ablation.}
  Sample-mean T2I MSE (faint) and a 20-point EMA (opaque).
  Left: the full training trajectory with a logarithmic $y$-axis.
  Right: a zoom-in of 10K--15K steps with a linear $y$-axis.
  The outlined region and connecting lines indicate the enlarged interval.}
  \label{fig:image_patch_size}
\end{figure}

\begin{figure}[!htb]
  \centering
  \includegraphics[width=\linewidth]{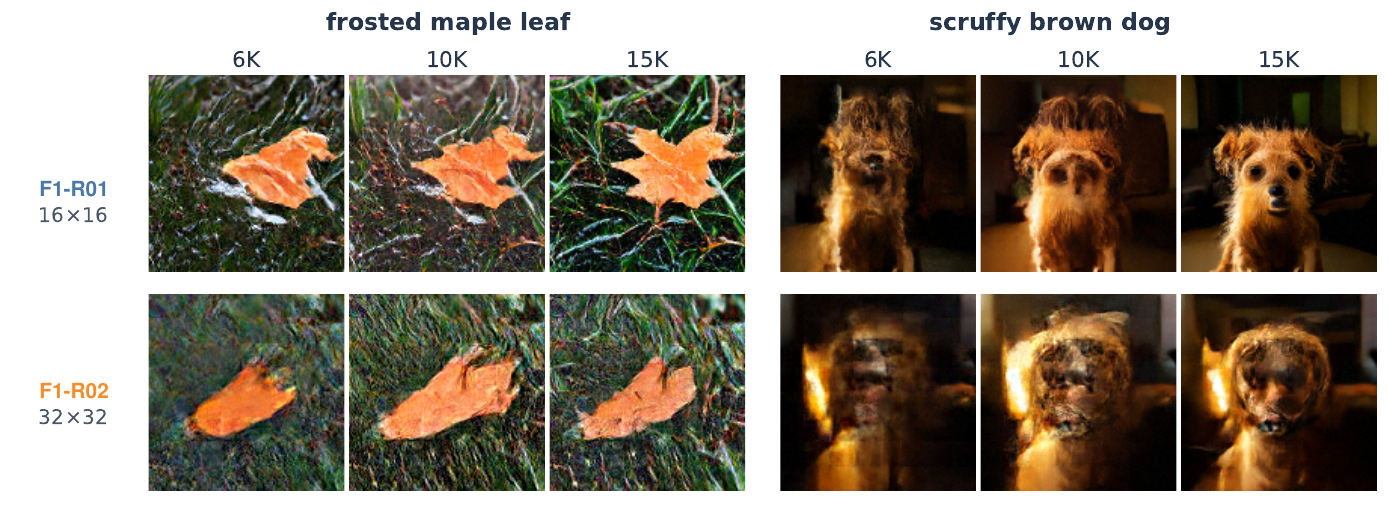}
  \caption{\textbf{Qualitative comparison of image patch sizes.}
  Rows show the $16\!\times\!16$ F1-R01 and $32\!\times\!32$ F1-R02 generation
  interfaces; within each prompt, columns follow training from 6K to 10K to
  15K.  All outputs use the same two prompts, DPM-Solver
  with 50 steps, timestep shift 1, and CFG 3.5 with global renormalization.}
  \label{fig:image_patch_size_qualitative}
\end{figure}

In this experimental family, we examine how image patch size affects convergence.
The $16\!\times\!16$ (F1-R01) and $32\!\times\!32$ (F1-R02)
runs differ only in the generation input projection,
\texttt{img\_gen\_linear\_proj}, and output head,
\texttt{img\_linear\_outproj}. Both initialize the understanding branch from
Joint Stage~1 and the generation branch from scratch; all other training
settings are identical.
With the same sequence-length budget, $32\!\times\!32$ accommodates four times
as many images per step and therefore consumes training data faster.
Figure~\ref{fig:image_patch_size} shows the training pixel-flow MSE,
with the full trajectory on the left and a zoom-in of 10K--15K steps on the right.
Despite this data-consumption advantage, the $32\!\times\!32$ model converges
more slowly. The $16\!\times\!16$ model maintains lower training
MSE throughout the late-training interval.
To examine how the loss differences manifest in generated images,
we compare the two patch sizes on matched prompts at 6K, 10K, and 15K steps
(Fig.~\ref{fig:image_patch_size_qualitative}).

\begin{mdframed}[
  linecolor=genaccent!65!white,
  backgroundcolor=unifiedred!50!white,
  linewidth=0.8pt,
  roundcorner=3pt,
  shadow=true,
  shadowsize=3pt,
  shadowcolor=black!20,
  innerleftmargin=7pt,
  innerrightmargin=7pt,
  innertopmargin=5pt,
  innerbottommargin=5pt,
  skipabove=6pt,
  skipbelow=6pt]
\noindent\textbf{Takeaway 1:}
With the same sequence-length budget, $32\!\times\!32$ patches process four
times as many images per step. Even so, the $16\!\times\!16$ run has lower
image-generation training loss (T2I MSE). This suggests that stronger spatial
compression makes image generation harder to learn.
\end{mdframed}

\Needspace{5\baselineskip}
\subsubsection{F2: Video Patch Size}
\begin{figure}[!htb]
  \centering
  \captionsetup{font=small,skip=3pt}
  \includegraphics[width=\linewidth]{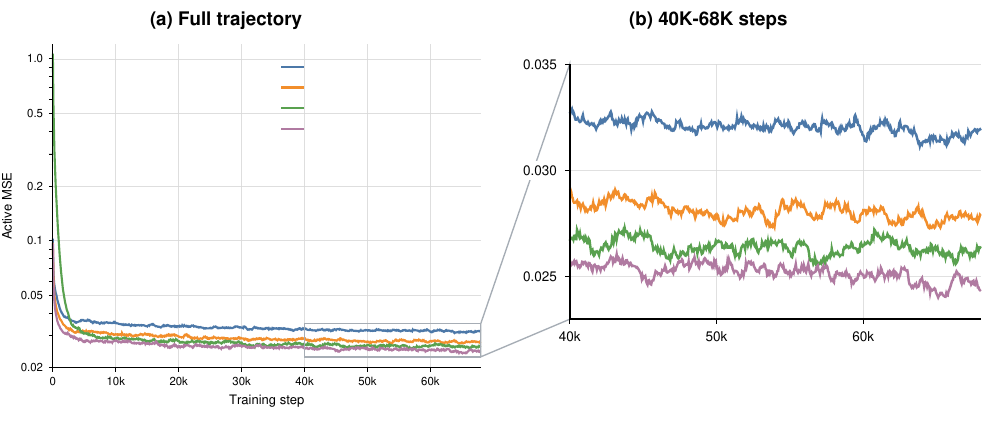}
  \caption{\textbf{Video patch-size ablation.}
  Patch configurations of
  \texttt{video\_gen\_linear\_proj} and \texttt{video\_linear\_outproj}:
  p32/t4 (F2-R01), p32/t2 (F2-R02), p16/t4 (F2-R03), and p32/t1 (F2-R04).
  All runs initialize the understanding branch from Joint Stage~1 and the
  generation branch from scratch; other training settings are identical.
  Faint curves show raw T2V losses; opaque curves show a 20-point EMA.
  The left panel covers 0--68K steps on a logarithmic scale;
  the right panel enlarges 40K--68K on a linear scale.}
  \label{fig:patch_ablation_losses}
\end{figure}

\begin{figure}[H]
  \centering
  \captionsetup{font=small,skip=3pt}
  \includegraphics[width=\linewidth]{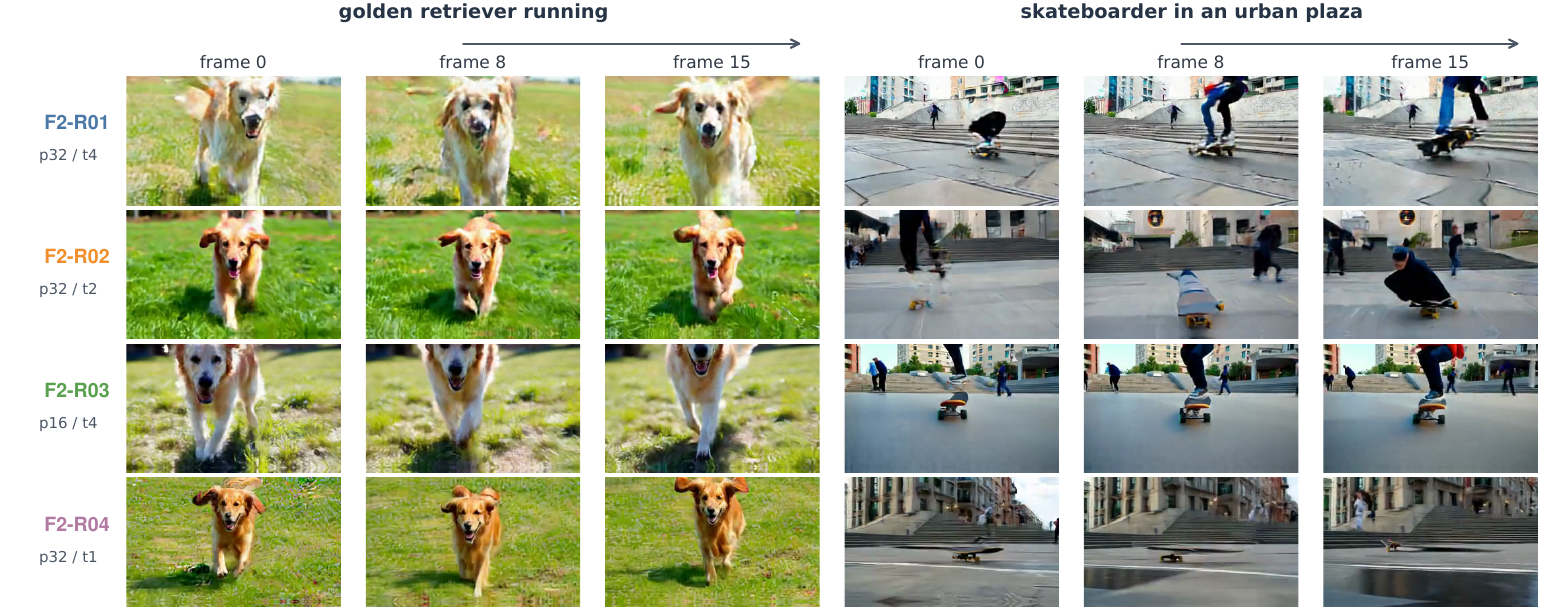}
  \caption{\textbf{Qualitative comparison of video patch configurations.}
  Rows correspond to p32/t4 (F2-R01), p32/t2 (F2-R02), p16/t4 (F2-R03), and
  p32/t1 (F2-R04).  For two matched prompts, columns show
  frames 0, 8, and 15 of each~16-frame,~$192\!\times\!320$~clip.}
  \label{fig:patch_size_t2v_examples}
\end{figure}

In this experimental family, we examine spatial and temporal patch sizes for video generation.
We vary \texttt{video\_gen\_linear\_proj} and
\texttt{video\_linear\_outproj}, the video generation input projection and output head,
to compare four tubelet-to-token mappings, written as spatial
width $\times$ height $\times$ number of frames $\rightarrow$ one LLM hidden token:
F2-R01: $32\!\times\!32\!\times\!4\rightarrow\mathbf{h}$;
F2-R02: $32\!\times\!32\!\times\!2\rightarrow\mathbf{h}$;
F2-R03: $16\!\times\!16\!\times\!4\rightarrow\mathbf{h}$; and
F2-R04: $32\!\times\!32\!\times\!1\rightarrow\mathbf{h}$,
where $\mathbf{h}\in\mathbb{R}^{d}$ is a $d$-dimensional LLM hidden token.
Thus, p32/t4 denotes $32\!\times\!32$ spatial aggregation and $4\!:\!1$
temporal aggregation; the other configurations follow the same convention.
The input projection maps the $3p^{2}t$ RGB values in each tubelet to
$d$ hidden features, and the output head maps each hidden token back to
$3p^{2}t$ output values.
All four runs initialize the understanding branch from Joint Stage~1 and
train a freshly initialized generation branch from scratch.
Only the patch configurations of these two linear layers vary; all other
training settings are identical.
Figure~\ref{fig:patch_ablation_losses} compares their active T2V MSE.
Across these settings, less aggressive spatiotemporal compression generally yields lower T2V loss, with p32/t1 (F2-R04) attaining the lowest loss. For the final model, however, we adopt p16/t4 (F2-R03) to align with the spatiotemporal compression convention used by common video VAEs such as Wan2.2~\cite{Wan2.2}.
Figure~\ref{fig:patch_size_t2v_examples} provides a qualitative comparison at the
68K-step checkpoint, showing three frames from each of two matched prompts.
\begin{mdframed}[
  linecolor=genaccent!65!white,
  backgroundcolor=unifiedred!50!white,
  linewidth=0.8pt,
  roundcorner=3pt,
  shadow=true,
  shadowsize=3pt,
  shadowcolor=black!20,
  innerleftmargin=7pt,
  innerrightmargin=7pt,
  innertopmargin=5pt,
  innerbottommargin=5pt,
  skipabove=11pt,
  skipbelow=6pt]
\noindent\textbf{Takeaway 2:}
Across the tested configurations, spatiotemporal compression
ranges from $1{,}024$ pixels $\rightarrow$ one video token (p16/t4 or p32/t1)
to $4{,}096$ pixels $\rightarrow$ one video token (p32/t4). Stronger
spatiotemporal compression generally yields higher T2V training loss, making
video generation harder to learn.
\end{mdframed}

\Needspace{5\baselineskip}
\FloatBarrier
\subsubsection{F3: Patch Artifacts}
\captionsetup[table]{font={small,stretch=1},skip=5pt}

\begin{figure}[H]
  \centering
  \includegraphics[width=\linewidth]{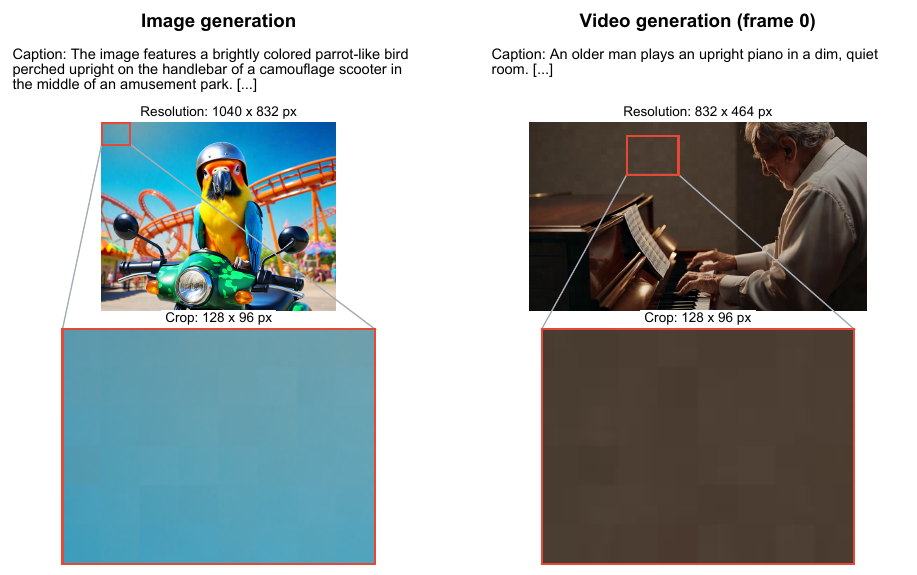}
  \caption{\textbf{Patch artifacts with linear pixel heads.}
  The left column shows an F3-R01 image output above a crop of its smooth sky;
  the right column shows frame 0 of the \emph{Upright piano} video above a crop
  of the smooth wall behind the pianist. Red boxes identify the displayed
  regions. The subtle grid boundaries are clearer when the figure is enlarged.}
  \label{fig:r01_patch_examples}
\end{figure}

Similar to MiniT2I~\cite{minit2i2026}, we observe that patch
artifacts become more pronounced at high classifier-free guidance (CFG)
scales, such as around $6$, particularly in low-texture regions when using
the linear decoder.
Figure~\ref{fig:r01_patch_examples} illustrates these grid-aligned intensity
changes in smooth regions of image and video outputs.

\Needspace{7\baselineskip}

In this experimental family, we test several decoder heads
in a 16-GPU setting and find that convolutional alternatives can reduce these
artifacts. \textbf{Because switching from a linear head to a convolutional head
requires further training, PixelUMM retains the default linear output heads
\texttt{img\_linear\_outproj} and \texttt{video\_linear\_outproj} unless
otherwise specified; this includes the released checkpoint, the model
evaluated in Sec.~\ref{sec:benchmarks}, and qualitative demo figures outside
this decoder ablation.} F3-R01 is this linear-head baseline
(Sec.~\ref{sec:architecture}, Eq.~\ref{eq:pixel_decoding}).

\begin{figure}[H]
  \centering
  \includegraphics[width=\linewidth]{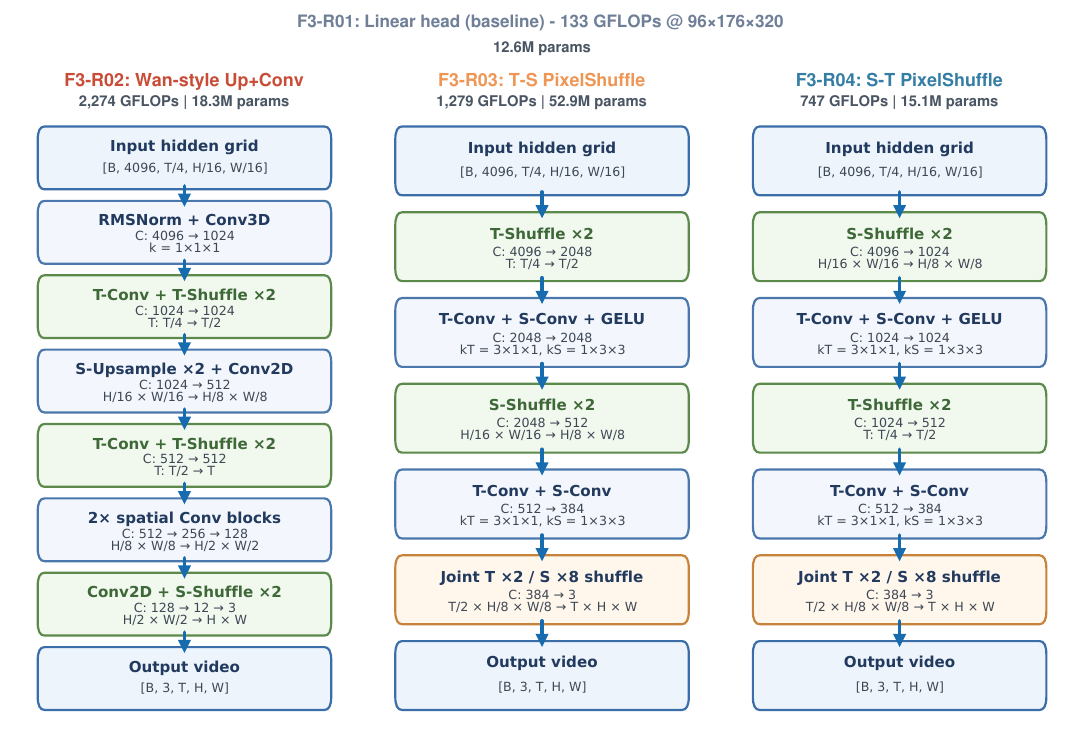}
  \caption{\textbf{Pixel-space video output-head architectures.}
  All three heads map the Transformer hidden grid at
  $T/4\times H/16\times W/16$ to RGB video at $T\times H\times W$.
  F3-R02 uses a Wan-style upsampling-and-convolution stack.
  F3-R03 and F3-R04 are PixelShuffle decoders that apply temporal
  and spatial shuffling in different orders; both end with the same joint
  temporal-spatial shuffle.
  T-Conv and S-Conv denote $3\times1\times1$ temporal and
  $1\times3\times3$ spatial convolutions, respectively.  GFLOPs are the
  convolution-only forward cost for one $96\times176\times320$ clip, using
  one multiply-accumulate as two FLOPs; normalization, activations, and
  parameter-free upsampling or shuffling are excluded.  For reference, the
  F3-R01 linear projection requires 133 GFLOPs under the same accounting.
  At this clip size, F3-R01 has 12.6M parameters. F3-R02,
  F3-R03, and F3-R04 respectively cost 2,274, 1,279, and 747 GFLOPs
  ($17.1\times$, $9.6\times$, and $5.6\times$ the linear head), with 18.3M,
  52.9M, and 15.1M parameters ($1.5\times$, $4.2\times$, and $1.2\times$
  the linear head). Parameter counts are rounded to 0.1M from the shown
  kernel dimensions; bias and normalization terms do not affect this
  precision.}
  \label{fig:f18_decoder_architecture}
\end{figure}

To address these patch artifacts, we investigate three convolutional
decoder heads as alternatives to the linear video output projection
(Fig.~\ref{fig:f18_decoder_architecture}).  F3-R02 is a Wan-style decoder built
from alternating upsampling and convolution blocks.
F3-R03 and F3-R04 use PixelShuffle with temporal-first (T-S)
and spatial-first (S-T) ordering, respectively; both end with the same joint
temporal-spatial shuffle. At $96\times176\times320$, the linear F3-R01 head costs
133 GFLOPs and has 12.6M parameters. The F3-R02, F3-R03, and F3-R04 heads
respectively cost 2,274, 1,279, and 747 GFLOPs ($17.1\times$, $9.6\times$,
and $5.6\times$ the linear head), with 18.3M, 52.9M, and 15.1M parameters
($1.5\times$, $4.2\times$, and $1.2\times$ the linear head).

\begin{figure}[!htb]
  \centering
  \includegraphics[width=\linewidth]{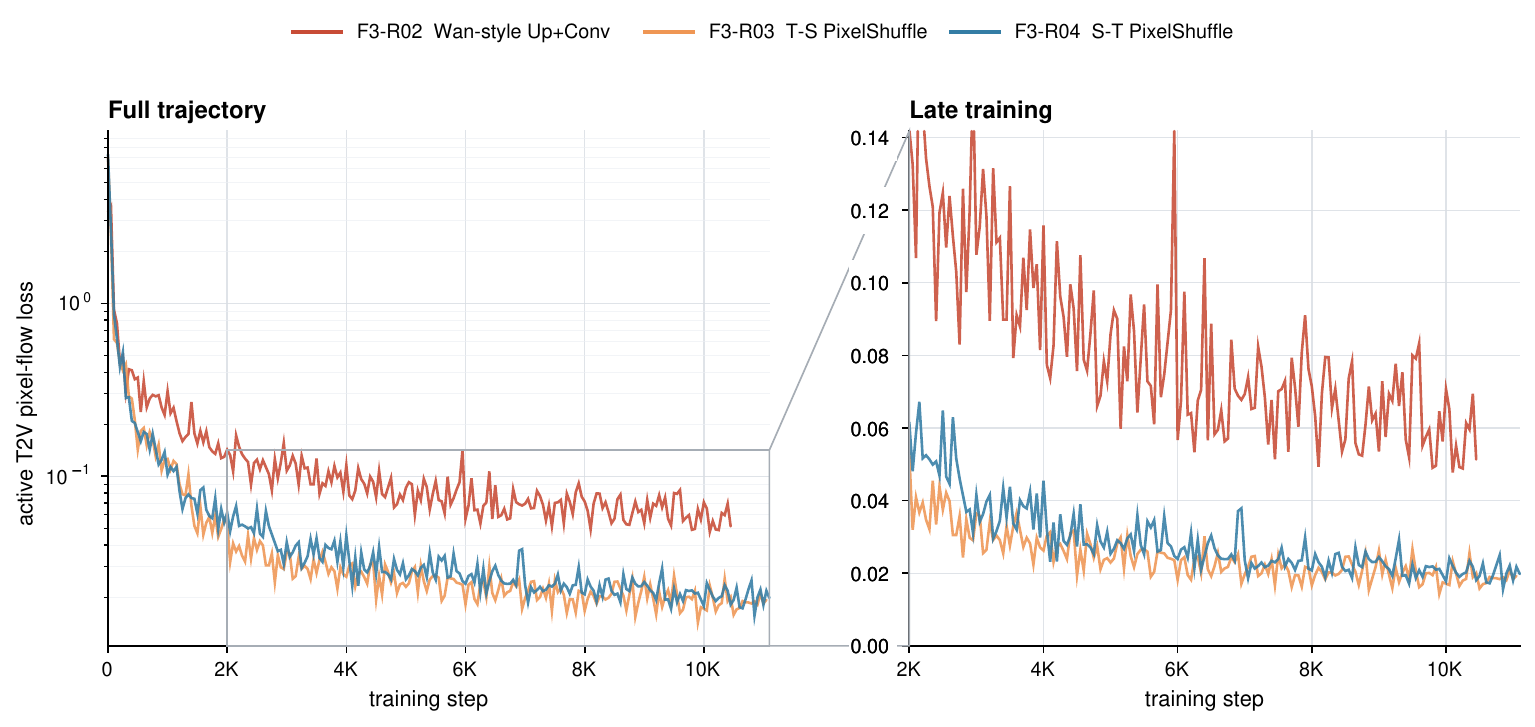}
  \caption{\textbf{Training loss for the pixel-space video output heads.}
  Raw T2V pixel-flow losses for the Wan-style upsample-conv decoder
  (F3-R02), T-S PixelShuffle decoder (F3-R03), and S-T PixelShuffle decoder (F3-R04). The left
  panel shows the complete trajectories on a logarithmic scale; the right
  panel enlarges the interval after 2K training steps.  Curves are shown
  exactly as logged, without EMA or other smoothing.}
  \label{fig:f18_decoder_loss}
\end{figure}

We initialize each model from Gen Stage~1 and discard
\texttt{video\_linear\_outproj}, replacing it with the corresponding decoder head.
We first examine how these decoder choices affect
optimization by comparing their training losses (Fig.~\ref{fig:f18_decoder_loss}).
We report the unweighted T2V pixel-flow loss without smoothing. The
T-S and S-T PixelShuffle heads converge to similar losses near $0.02$,
whereas the Wan-style upsample-conv head remains above $0.05$ over the
observed interval.

\begin{figure}[!htb]
  \centering
  \includegraphics[width=\linewidth]{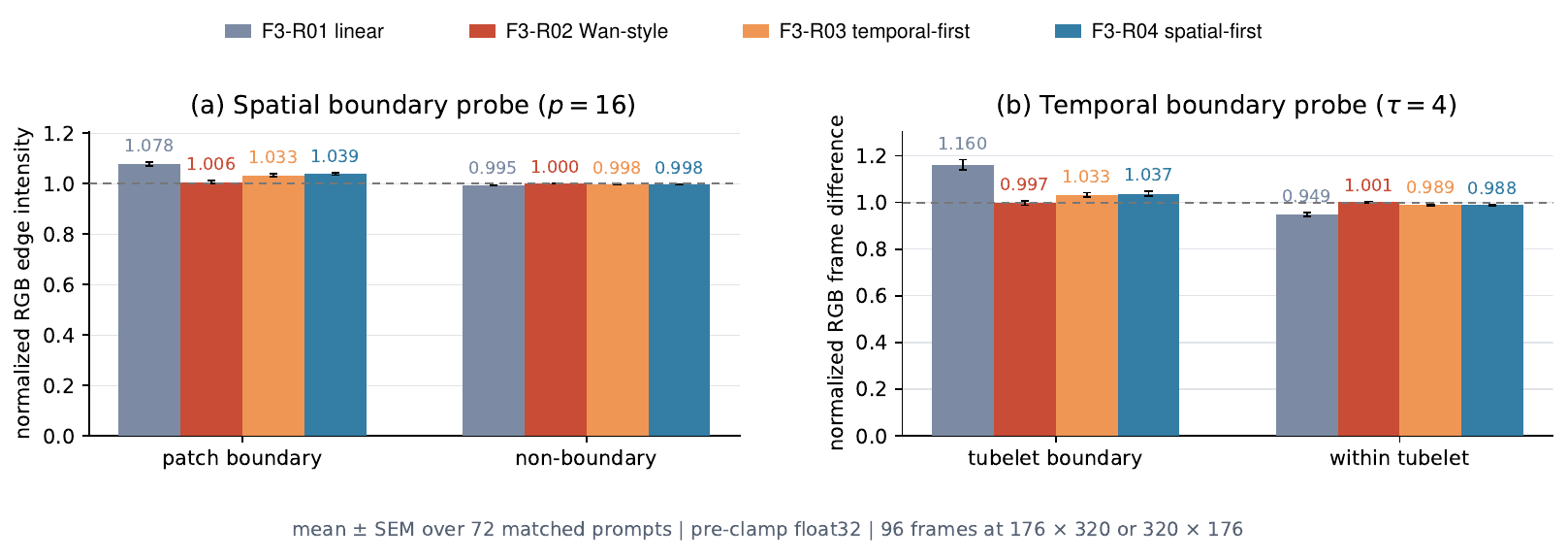}
  \caption{\textbf{Lossless spatial and temporal boundary probes over 72 evaluation prompts.}
  For each output head, we measure all 96 frames and all spatial locations of
  the pre-clamp float32 output at $176\times320$ or $320\times176$.  The spatial probe compares
  RGB edge intensity at 16-pixel patch boundaries with non-boundary edges; the
  temporal probe compares adjacent-frame differences at 4-frame tubelet
  boundaries with within-tubelet transitions.  We normalize each prompt by
  its corresponding video-wide adjacent-sample mean and then aggregate the 72
  evaluation prompts with equal weight.  Bars and error bars report mean $\pm$
  SEM across prompts ($72$ prompts per run; $288$ videos total).}
  \label{fig:lossless_boundary_probe_all_prompts}
\end{figure}

Following MiniT2I~\cite{minit2i2026}, we probe discontinuities at spatial
patch boundaries and extend the analysis to temporal tubelet boundaries.
We evaluate the decoder heads on an evaluation set containing 72 prompts,
measuring boundary effects from their pre-clamp float32 outputs
(Fig.~\ref{fig:lossless_boundary_probe_all_prompts}).
Each prompt is normalized independently before aggregation so that
high-motion videos do not dominate the mean.  F3-R01 shows the strongest spatial
and temporal boundary elevations (1.078 and 1.160); F3-R02 is near the normalized
baseline, while F3-R03 and F3-R04 retain smaller residual elevations.

\begin{figure}[H]
  \centering
  \includegraphics[width=0.85\linewidth]{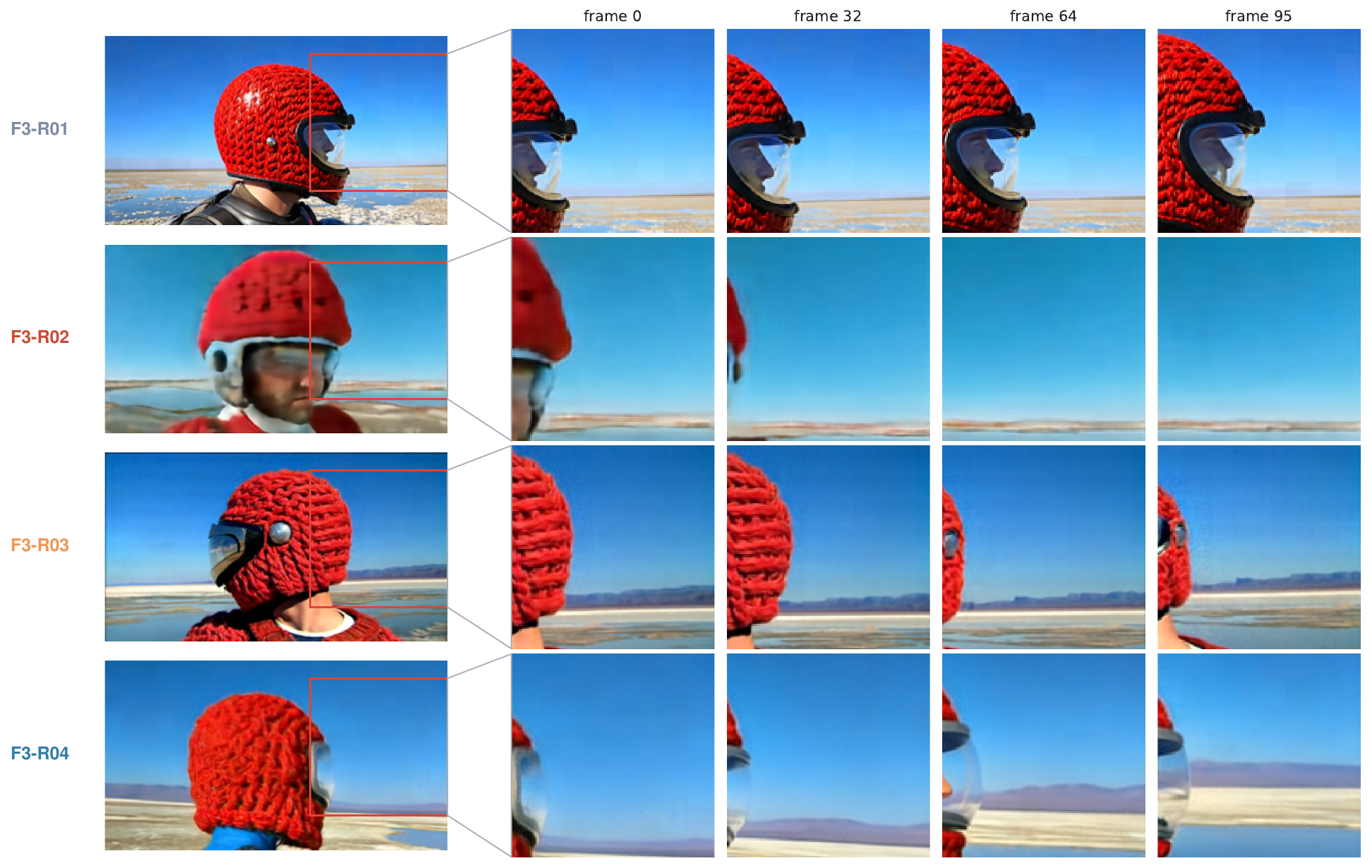}
  \caption{\textbf{Patch artifacts across pixel-space video generation.}
  Each row shows one output-head design under the same standard evaluation prompt:
  a linear baseline (F3-R01), a Wan-style upsample-conv decoder (F3-R02), a T-S
  PixelShuffle decoder (F3-R03), and an S-T PixelShuffle decoder (F3-R04).  The red box in the full frame
  defines a fixed crop.  The four panels to its right show that same spatial
  location in chronological order at frames 0, 32, 64, and 95, spanning the
  complete 96-frame video.  The displayed frames are lossless PNGs captured
  before video encoding; the corresponding pre-clamp float32 tensors are
  retained separately for numerical analysis.}
  \label{fig:patch_artifact}
\end{figure}

\FloatBarrier
We further examine how these boundary effects appear in generated
videos by comparing the three decoder heads with the linear F3-R01 baseline
on the same text-to-video prompt (Fig.~\ref{fig:patch_artifact}).
We use the RGB frames captured after
clamp and uint8 conversion but before MP4/GIF encoding, apply a fixed spatial
crop to expose local patch structure, and inspect the same location at four
time points spanning the \mbox{complete 96-frame generation.}

Overall, convolutional heads suppress the linear baseline's grid artifacts: the Wan-style decoder gives the cleanest boundaries but much higher loss and blurrier outputs, whereas PixelShuffle offers the better trade-off, with T-S (F3-R03) slightly outperforming S-T (F3-R04).

\begin{mdframed}[
  linecolor=genaccent!65!white,
  backgroundcolor=unifiedred!50!white,
  linewidth=0.8pt,
  roundcorner=3pt,
  shadow=true,
  shadowsize=3pt,
  shadowcolor=black!20,
  innerleftmargin=7pt,
  innerrightmargin=7pt,
  innertopmargin=5pt,
  innerbottommargin=5pt,
  skipabove=6pt,
  skipbelow=6pt]
\noindent\textbf{Takeaway 3:}
Convolutional heads reduce patch artifacts. Among these designs, PixelShuffle
offers a better trade-off between training loss and artifact suppression than
the upsampling-and-convolution variant.
\end{mdframed}

\Needspace{5\baselineskip}
\FloatBarrier
\subsubsection{F4: Pixel-Space and VAE-Space Training Dynamics}
\label{sec:pixel_vae_dynamics}
\begin{figure}[H]
  \centering
  \captionsetup{skip=1pt}
  \includegraphics[width=\linewidth]{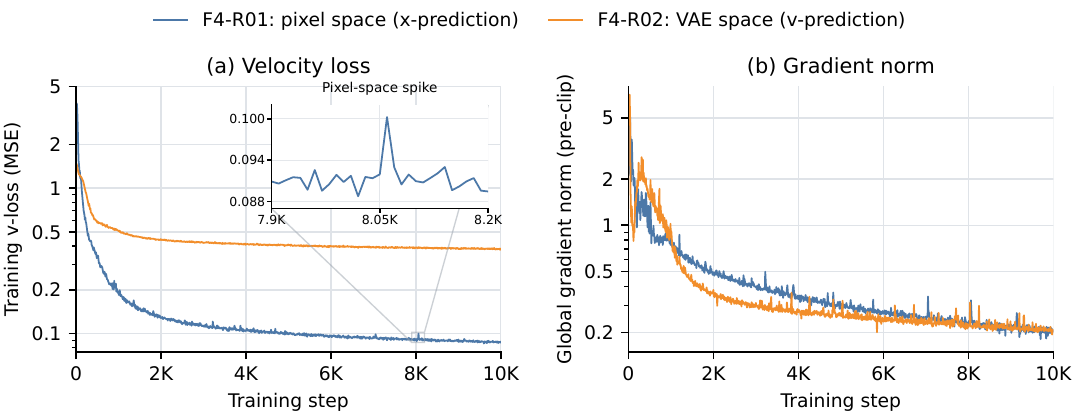}
  \caption{\textbf{Pixel-space versus VAE-space early training dynamics (F4).}
  Left: T2I velocity MSE measured in pixel and VAE latent
  spaces; the difference in loss magnitude alone does not establish relative
  learning speed or generation quality. The inset enlarges a representative
  pixel-space spike at 7.9K--8.2K steps. Right: pre-clip global gradient norm.
  Both panels use logarithmic $y$-axes and show unsmoothed 10-step records.}
  \label{fig:pixel_vae_dynamics}
\end{figure}

In this experimental family, we compare pixel-space and
VAE-space training over the first 10K steps. F4-R01 reuses the
$32\!\times\!32$ pixel-patch configuration of F1-R02. F4-R02 uses a frozen
Wan2.2 VAE~\cite{Wan2.2} with $16\!\times\!16$ spatial compression and $2\!\times\!2$
latent patchification before the transformer, matching the effective
$32\!\times\!32$ spatial token stride of F4-R01. The pixel-space run uses
$x$-prediction with $v$-loss; the VAE-space run uses $v$-prediction with $v$-loss.
Both runs initialize the understanding branch from Joint Stage~1 and the
generation branch from scratch.
Figure~\ref{fig:pixel_vae_dynamics} compares the pixel- and
VAE-space training curves. At 10K steps, the logged VAE-space loss is about
$4.3\times$ the pixel-space loss. Because the losses are measured in different
spaces, this numerical gap does not show that pixel-space training learns
faster or produces better images. The pre-clip global gradient norms are nearly
equal, while the pixel-space loss occasionally spikes.
\enlargethispage{4\baselineskip}
\vspace{-7pt}

\begin{mdframed}[
  linecolor=genaccent!65!white,
  backgroundcolor=unifiedred!50!white,
  linewidth=0.8pt,
  roundcorner=3pt,
  shadow=true,
  shadowsize=3pt,
  shadowcolor=black!20,
  innerleftmargin=7pt,
  innerrightmargin=7pt,
  innertopmargin=5pt,
  innerbottommargin=5pt,
  skipabove=6pt,
  skipbelow=6pt]
\noindent\textbf{Takeaway 4:}
At 10K steps, the VAE-space loss is about $4.3\times$ the
pixel-space loss, but this gap across different prediction spaces does not show
faster pixel-space learning. Their gradient norms are similar, and pixel-space
training shows occasional loss spikes.
\end{mdframed}

\Needspace{5\baselineskip}
\FloatBarrier
\subsubsection{F5: Model Size}
\enlargethispage{2\baselineskip}
In this experimental family, we examine whether increasing model
capacity improves optimization for understanding and generation during joint training.

\begin{figure}[H]
  \centering
  \begin{subfigure}[t]{0.45\textwidth}
    \centering
    \includegraphics[width=\linewidth]{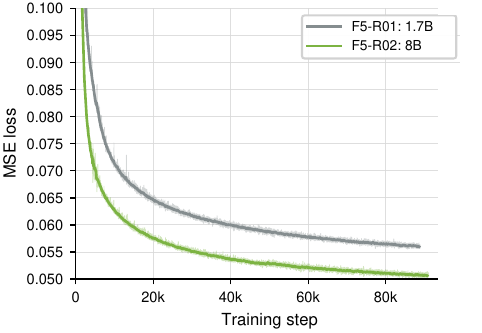}
    \caption{MSE loss.}
  \end{subfigure}\hfill
  \begin{subfigure}[t]{0.45\textwidth}
    \centering
    \includegraphics[width=\linewidth]{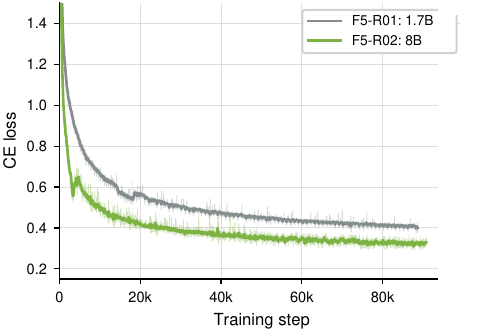}
    \caption{CE loss.}
  \end{subfigure}
  \caption{\textbf{Model-size scaling in Joint Stage~1.}
  We compare the 1.7B (F5-R01) and 8B (F5-R02)
  models using the same training recipe with a global batch size of 256.}
  \label{fig:joint_stage1_model_scaling}
\end{figure}

We compare the 1.7B (F5-R01) and 8B (F5-R02) models using
the same training recipe with a global batch size of 256, tracking both
pixel-flow MSE and text cross-entropy
(Fig.~\ref{fig:joint_stage1_model_scaling}).
The 8B model reaches comparable losses in roughly one-third as many training
steps: MSE $0.060$ at about 14K versus 40K steps, and CE $0.50$ at about
10K versus 31K steps. These estimates from the plotted curves indicate
approximately $3\times$ faster convergence in training steps, rather than
wall-clock time.

\begin{mdframed}[
  linecolor=genaccent!65!white,
  backgroundcolor=unifiedred!50!white,
  linewidth=0.8pt,
  roundcorner=3pt,
  shadow=true,
  shadowsize=3pt,
  shadowcolor=black!20,
  innerleftmargin=7pt,
  innerrightmargin=7pt,
  innertopmargin=5pt,
  innerbottommargin=5pt,
  skipabove=6pt,
  skipbelow=6pt]
\noindent\textbf{Takeaway 5:}
With the same training recipe and global batch size, the 8B model reaches
comparable generation and text losses in roughly $1/3$ as many training
steps as the 1.7B model.
\end{mdframed}

\Needspace{6\baselineskip}
\FloatBarrier
\subsubsection{F6: Compute Scaling}
\begin{figure}[H]
  \centering
  \begin{subfigure}[t]{0.45\textwidth}
    \centering
    \includegraphics[width=\linewidth]{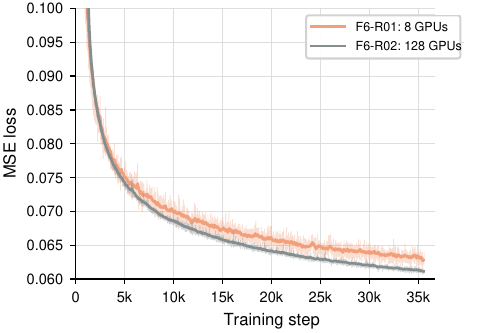}
    \caption{MSE loss.}
  \end{subfigure}\hfill
  \begin{subfigure}[t]{0.45\textwidth}
    \centering
    \includegraphics[width=\linewidth]{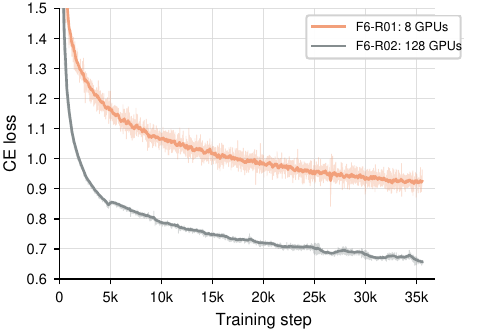}
    \caption{CE loss.}
  \end{subfigure}
  \caption{\textbf{Distributed compute scaling under a matched MidJourney recipe.}
  We compare F6-R01 and F6-R02, which use the same 1.7B model, data mixture, p32 tokenization, pixel embedder and head, loss weights, and optimizer settings.
  Opaque curves are exponential moving averages over 25 logged points; faint curves show raw measurements.}
  \label{fig:midjourney_compute_scaling}
\end{figure}

In this experimental family, we study how increasing distributed
training resources affects optimization at fixed model capacity. We compare
F6-R01 and F6-R02 using the same 1.7B model and matched data, visual interface,
loss weights, and optimizer settings.  We track both generation and text
losses as training progresses (Fig.~\ref{fig:midjourney_compute_scaling}).
Increasing the GPU count from 8 to 128 yields a larger step-efficiency gain
for text than for generation: MSE reaches $0.065$ at about 17K versus 23K
steps ($1.3\times$ faster), whereas CE reaches $1.0$ at about 2K versus
17.5K steps (roughly $9\times$ faster). These curve-based estimates measure
training steps to a fixed loss, not wall-clock speedup or compute efficiency.

\begin{mdframed}[
  linecolor=genaccent!65!white,
  backgroundcolor=unifiedred!50!white,
  linewidth=0.8pt,
  roundcorner=3pt,
  shadow=true,
  shadowsize=3pt,
  shadowcolor=black!20,
  innerleftmargin=7pt,
  innerrightmargin=7pt,
  innertopmargin=5pt,
  innerbottommargin=5pt,
  skipabove=6pt,
  skipbelow=6pt]
\noindent\textbf{Takeaway 6:}
Increasing the GPU count from 8 to 128 lets CE reach a loss of
1.0 in about 2K rather than 17.5K training steps (roughly $9\times$ fewer).
MSE reaches 0.065 in about 17K rather than 23K steps (roughly $1.3\times$
fewer). Thus, increasing the GPU count benefits CE convergence substantially
more than MSE convergence in terms of training steps.
\end{mdframed}

\Needspace{5\baselineskip}
\FloatBarrier
\subsubsection{F7: Multimodal Context Conditioning}
\label{sec:condition_routing}
In this experimental family, we investigate whether clean visual
conditions routed through the understanding expert can guide synthesis by the
generation expert.
SenseNova-U1 introduced encoder-free visual conditioning that routes clean
image inputs through the understanding expert and synthesizes targets through
the generation expert~\cite{diao2026sensenovau1}.
We extend this design to video, supporting image-to-video generation and
video editing with the same conditioning interface.

\begin{figure}[H]
  \centering
  \includegraphics[width=\linewidth]{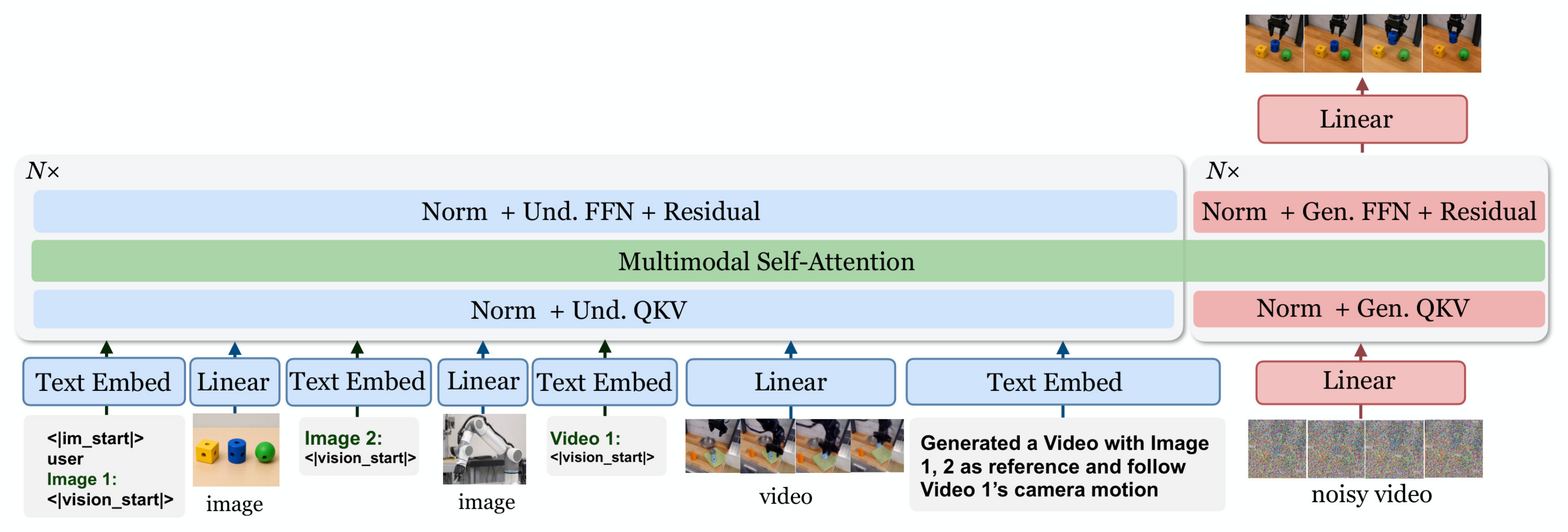}
  \caption{\textbf{Multimodal context conditioning.} Text, two reference images, and an input video enter the
  understanding expert, while noisy target-video tokens enter the generation
  expert. Both routes interact through shared multimodal self-attention.}
  \label{fig:context_conditioning}
\end{figure}

\begin{figure}[H]
  \centering
  \includegraphics[width=0.90\linewidth]{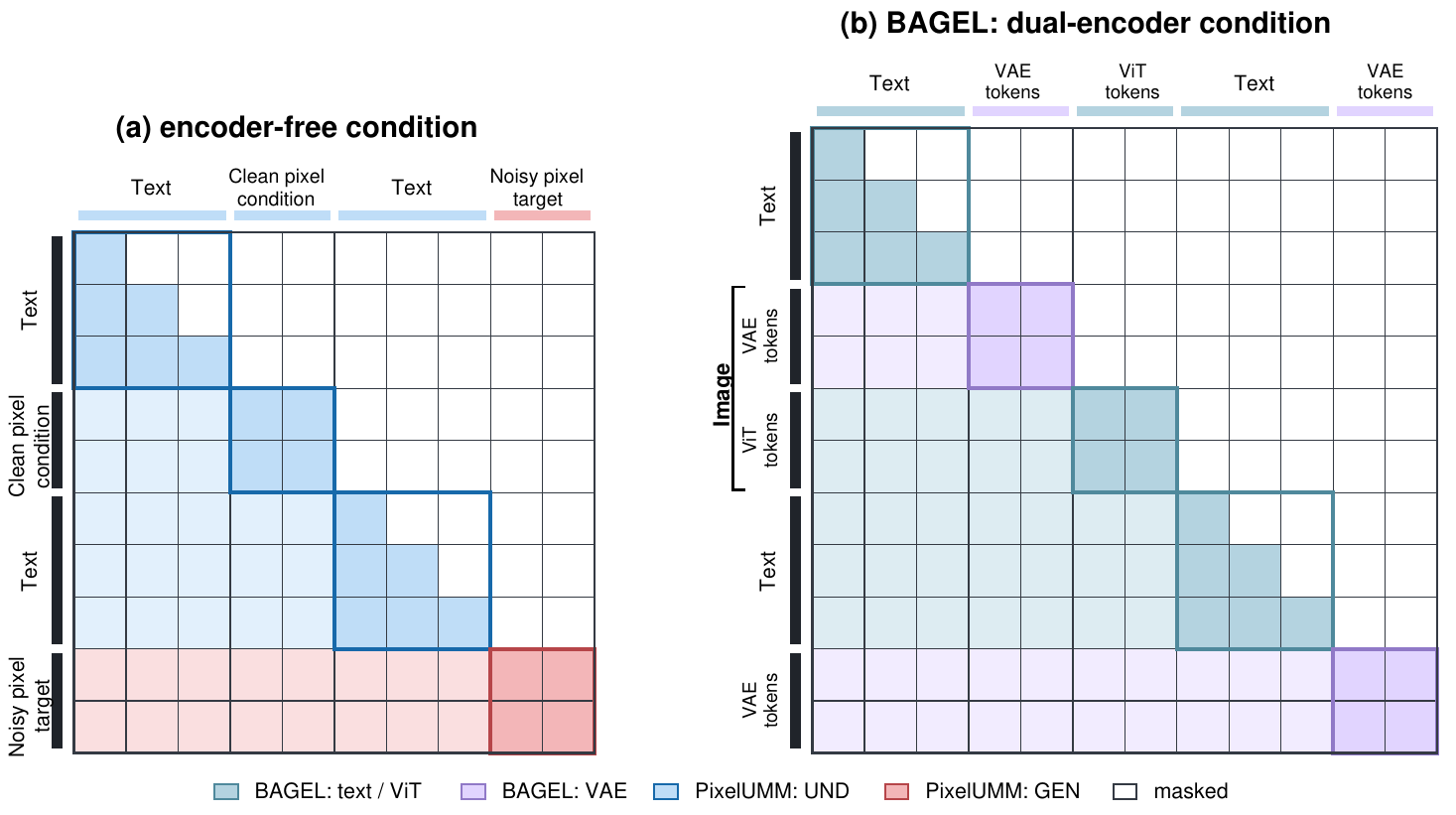}
  \caption{\textbf{Dual-encoder and encoder-free visual conditioning.}
  (a) Clean visual conditions enter only
  through the understanding expert, while noisy targets enter through the
  generation expert. (b) BAGEL uses dual ViT/VAE representations of a clean
  image~\cite{deng2025bagel}.
  SenseNova-U1 introduced this encoder-free conditioning design for
  images~\cite{diao2026sensenovau1}; PixelUMM extends it to image-to-video
  generation and video editing.
  Rows are queries and columns are keys; colored cells indicate visible
  attention and white cells are masked. Block sizes are schematic.}
  \label{fig:condition_routing}
\end{figure}

Figure~\ref{fig:context_conditioning} illustrates PixelUMM's video-conditioning
interface: interleaved text instructions and clean image/video conditions enter
through the understanding expert, while noisy target-video tokens enter the
generation expert and interact with the conditions through shared attention.
Figure~\ref{fig:condition_routing} contrasts this single-stream
interface with BAGEL's dual ViT/VAE representation~\cite{deng2025bagel}.
In BAGEL's interleaved-generation setting, each conditioning image contributes
both ViT and clean VAE tokens to the context, increasing the visual-token count
and KV-cache memory as the visual context grows. PixelUMM instead routes clean
visual conditions through the understanding expert without a separate
VAE-token stream.

To examine whether learning visual
conditioning can harm understanding, we introduce a multi-task fine-tuning
stage (Table~\ref{tab:joint_stage3_condition}). The model starts from an
intermediate Joint Stage~2 checkpoint and trains on eight tasks for 15K steps,
including image and video editing alongside understanding and generation.
Table~\ref{tab:editing_understanding} compares understanding performance
before this stage (F7-R01) and after it (F7-R02). Image understanding shows mixed
changes: for example, BLINK improves by 2.37 points and CV-Bench by 2.27,
whereas SEED-I decreases by 5.41 points. All four video benchmarks improve,
by 0.89--3.49 points in \texttt{dense\_mode} and
1.00--3.49 points in \texttt{sparse\_mode}.
These results suggest that acquiring visual
conditioning capabilities through multi-task fine-tuning need not cause a
uniform loss of understanding ability, although individual image benchmarks
can regress.

\enlargethispage{2\baselineskip}
\FloatBarrier
\begin{table}[H]
\centering
\footnotesize
\renewcommand{\arraystretch}{1.18}
\setlength{\tabcolsep}{0pt}
\begin{tabular*}{\linewidth}{@{\extracolsep{\fill}}lr@{\hspace{2em}}lr@{\hspace{2em}}lr@{}}
\toprule
\multicolumn{2}{@{}l}{\textbf{Optimization}} &
\multicolumn{2}{l}{\textbf{Data and objectives}} &
\multicolumn{2}{l@{}}{\textbf{Per-step example ratio}} \\
\cmidrule(r{1.1em}){1-2}\cmidrule(lr{1.1em}){3-4}\cmidrule(l){5-6}
Understanding branch & \trainedbranch & Loss weight$^{\dagger}$ & $1:10:30$ & Text only & 1 \\
Generation branch & \trainedbranch & Seq length$^{\ddagger}$ & 60K / 32,768 & Image understanding (I2T) & 2 \\
Learning rate & $2\times10^{-5}$ & Time shift & 1 & Image generation (T2I) & 2 \\
LR scheduler & Constant & Und resolution (image) & Native & Video understanding (V2T) & 2 \\
Optimizer & Fused Adam & Und resolution (video) & $448^2$ & Video generation (T2V) & 3 \\
$(\beta_1,\beta_2,\epsilon)$ & $(0.9,\,0.99,\,10^{-8})$ & Gen resolution (image) & $256^2$ & Image-to-video (I2V) & 3 \\
Weight decay & 0.0 & Gen resolution (video) & $256^2$ & Image any-to-any & 2 \\
Gradient norm clip & 0.2 & Gen duration (video) & 4\,s / 96 frames / 24\,FPS & Video any-to-any & 3 \\
Training steps & 15K & & & & \\
Warmup steps & 1000 & & & & \\
\bottomrule
\end{tabular*}
\par\vspace{3pt}
{\footnotesize\raggedright
\trainedbranch\ Trainable.
$^{\dagger}$CE : Image MSE : Video MSE.
$^{\ddagger}$Visual / text token caps.\par}
\caption{\textbf{Multi-task fine-tuning recipe.}
The model is initialized from an intermediate checkpoint in Joint Stage~2.
Video resolutions are per frame. Per-step example ratios specify relative numbers of examples consumed per step and need not sum to one.}
\label{tab:joint_stage3_condition}
\end{table}

\begin{table}[H]
  \newcommand{\deltapos}[1]{\textcolor[HTML]{287D60}{\ensuremath{+#1}}}
  \newcommand{\deltaneg}[1]{\textcolor[HTML]{B6535C}{\ensuremath{-#1}}}
  \centering

  \footnotesize
  \setlength{\tabcolsep}{1pt}
  \renewcommand{\arraystretch}{1.35}
  \begin{tabular*}{\linewidth}{@{\extracolsep{\fill}}l*{11}{r}@{}}
    \toprule
    Checkpoint & MMMU & MMStar & RWQA & SEED-I & AI2D & DocVQA & ChartQA & InfoVQA & TextVQA & OCRBench & MME \\
    \midrule
    F7-R01 & 40.44 & 55.81 & 70.98 & 78.10 & 80.18 & 90.21 & 82.32 & 64.59 & 78.72 & 77.60 & 1752.67 \\
    F7-R02 & 41.44 & 55.43 & 69.28 & 72.69 & 79.40 & 90.35 & 81.64 & 63.40 & 78.99 & 77.30 & 1805.29 \\
    $\Delta$ & \deltapos{1.00} & \deltaneg{0.38} & \deltaneg{1.70} & \deltaneg{5.41} & \deltaneg{0.78} & \deltapos{0.14} & \deltaneg{0.68} & \deltaneg{1.19} & \deltapos{0.27} & \deltaneg{0.30} & \deltapos{52.62} \\
    \bottomrule
  \end{tabular*}
  \par\vspace{4pt}
  \begin{tabular*}{\linewidth}{@{\extracolsep{\fill}}l*{10}{r}@{}}
    \toprule
    Checkpoint & GQA & MMVP & SEED2+ & CV-Bench & CountBench & PixMo-Count & V* & MMMU-Pro & BLINK & MuirBench \\
    \midrule
    F7-R01 & 62.10 & 81.00 & 64.82 & 77.03 & 93.48 & 74.16 & 75.92 & 27.40 & 49.41 & 35.58 \\
    F7-R02 & 62.53 & 81.00 & 64.73 & 79.30 & 92.87 & 73.22 & 74.35 & 28.15 & 51.78 & 36.58 \\
    $\Delta$ & \deltapos{0.43} & \textcolor{black!60}{$0.00$} & \deltaneg{0.09} & \deltapos{2.27} & \deltaneg{0.61} & \deltaneg{0.94} & \deltaneg{1.57} & \deltapos{0.75} & \deltapos{2.37} & \deltapos{1.00} \\
    \bottomrule
  \end{tabular*}
  \par\vspace{4pt}
  \begin{tabular*}{\linewidth}{@{\extracolsep{\fill}}l*{8}{r}@{}}
    \toprule
    & \multicolumn{4}{c}{Video: \texttt{dense\_mode}}
    & \multicolumn{4}{c}{Video: \texttt{sparse\_mode}} \\
    \cmidrule(lr){2-5}\cmidrule(lr){6-9}
    Checkpoint & MVBench & Video-MME & LVB & LVBench
    & MVBench & Video-MME & LVB & LVBench \\
    \midrule
    F7-R01 & 65.60 & 53.74 & 56.10 & 37.31 & 65.47 & 54.30 & 56.39 & 37.31 \\
    F7-R02 & 67.65 & 54.96 & 56.99 & 40.80 & 68.10 & 55.30 & 57.74 & 40.80 \\
    $\Delta$ & \deltapos{2.05} & \deltapos{1.22} & \deltapos{0.89} & \deltapos{3.49} & \deltapos{2.63} & \deltapos{1.00} & \deltapos{1.35} & \deltapos{3.49} \\
    \bottomrule
  \end{tabular*}
\par
  \caption{\textbf{Understanding after
  multi-task fine-tuning with visual conditioning.}
  F7-R01: before multi-task fine-tuning;
  F7-R02: after multi-task fine-tuning;
  $\Delta=\text{F7-R02}-\text{F7-R01}$. Video: \texttt{dense\_mode} ($\leq$384 frames)
  and \texttt{sparse\_mode} ($\leq$96 frames), without subtitles for Video-MME.
  LVB denotes LongVideoBench; scores retain their original scales.}
  \label{tab:editing_understanding}
\end{table}

\begin{figure}[H]
  \centering
  \captionsetup{font=small,skip=3pt}
  \includegraphics[width=\linewidth]{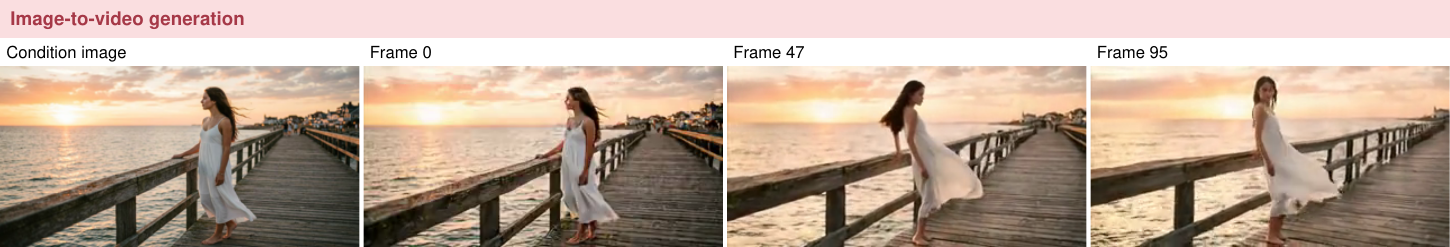}
  \caption{\textbf{Image-to-video conditioning through the understanding expert.}
  The clean first-frame condition is followed by frames 0, 47, and 95 of the
  generated 96-frame clip, in which the woman turns toward the camera and
  gives a small wave on a seaside pier at sunset.  The output is sampled with DPM-Solver for 50
  steps, timestep shift 10, text CFG 6, image-reference scale 1, and no CFG renormalization.}
  \label{fig:condition_i2v}
\end{figure}

\begin{figure}[H]
  \centering
  \captionsetup{font=small,skip=3pt}
  \includegraphics[width=0.85\linewidth]{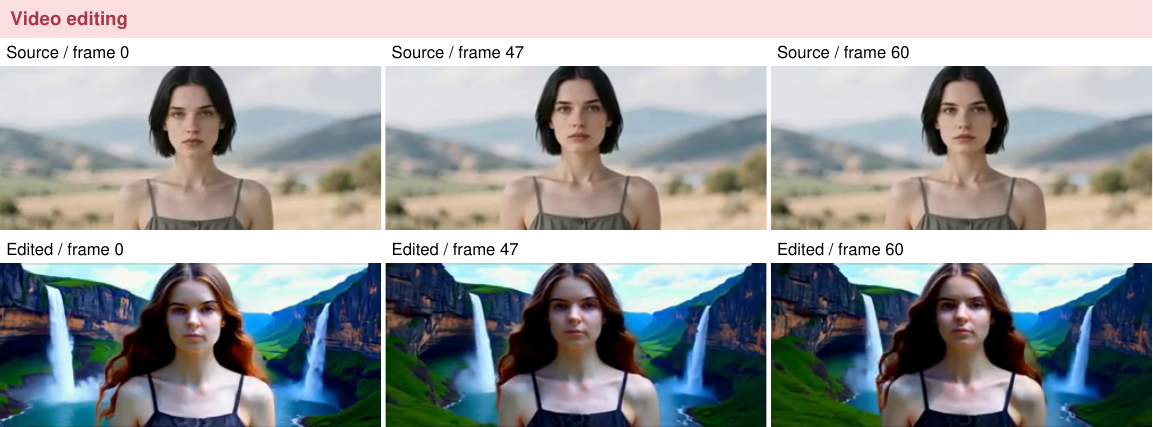}
  \caption{\textbf{Video editing through understanding-expert conditioning.}
  The source and edited rows show frames 0, 47, and 60.  The instruction
  changes the background to a waterfall valley with cliffs and replaces the
  short bob with long wavy hair while retaining temporal consistency.
  Sampling uses DPM-Solver for 50 steps,
  timestep shift 10, text CFG 6, video-reference scale 1, and no CFG
  renormalization.}
  \label{fig:condition_video_edit}
\end{figure}

\begin{mdframed}[
  linecolor=genaccent!65!white,
  backgroundcolor=unifiedred!50!white,
  linewidth=0.8pt,
  roundcorner=3pt,
  shadow=true,
  shadowsize=3pt,
  shadowcolor=black!20,
  innerleftmargin=7pt,
  innerrightmargin=7pt,
  innertopmargin=5pt,
  innerbottommargin=5pt,
  skipabove=6pt,
  skipbelow=6pt]
\noindent\textbf{Takeaway 7:}
PixelUMM can route clean visual conditions through the understanding expert
without systematically degrading its understanding capability.
\end{mdframed}

\FloatBarrier
\Needspace{10\baselineskip}
\subsubsection{F8: Video Understanding Interfaces}
\label{sec:video_understanding_interfaces}
In this experimental family, we compare two video-understanding
interfaces: \texttt{dense\_mode}, which can sample at 4 FPS and form temporal
tubelets, and \texttt{sparse\_mode}, which samples at 1 FPS and projects frames
independently, as described under ``Video understanding modes'' in
Sec.~\ref{sec:architecture}. F8-R01 is the checkpoint obtained after 9K steps
of Und Stage~3 training and is evaluated with both interfaces
(Table~\ref{tab:video_understanding_interfaces}). During training,
videos of at most 96 seconds with a source frame rate of at least 4 FPS use
the two interfaces in a 1:1 ratio.
Let $D$ denote video duration in seconds and $r$ the source frame rate in FPS.
The \texttt{dense\_mode} evaluation protocol uses three cases:
(i) for $D\leq96$ and $r\geq4$, sample at a strict 4 FPS, up to 384 frames,
and apply temporal tubelets with $\tau=4$;
(ii) for $D\leq96$ and $1\leq r<4$, fall back to strict 1 FPS sampling
and project each frame independently;
(iii) for $D>96$ or $r<1$, uniformly sample up to 96 frames over the video
and use the same frame-based interface.
In contrast, \texttt{sparse\_mode} uses strict 1 FPS sampling (up to 96
frames) whenever $D\leq96$ and $r\geq1$, and uniform sampling of up to
96 frames otherwise; it always projects frames independently.

Thus, the evaluation protocols differ only for videos of at
most 96 seconds with a source frame rate of at least 4 FPS. Both use the
same native-aspect-ratio resolution budget of $448^2$ pixels per frame.
Despite the higher sampling rate, we do not observe a consistent improvement
from \texttt{dense\_mode}: MVBench changes by only $+0.07$ points,
Video-MME and LongVideoBench decrease by $0.22$ and $0.68$ points,
respectively, and LVBench is unchanged. Under this setting, higher-FPS
tubelet input offers no clear advantage over~sparse~frame-based~input.

\begin{table}[H]
  \centering

  \small
  \setlength{\tabcolsep}{4pt}
  \renewcommand{\arraystretch}{1.08}
  \begin{tabular*}{\linewidth}{@{\extracolsep{\fill}}lrrrr@{}}
    \toprule
    Interface & MVBench & Video-MME & LongVideoBench & LVBench \\
    \midrule
    \texttt{sparse\_mode} & 71.15 & 57.89 & 59.39 & 41.38 \\
    \texttt{dense\_mode} & 71.22 & 57.67 & 58.71 & 41.38 \\
    $\Delta$ & \textcolor[HTML]{287D60}{$+0.07$}
    & \textcolor[HTML]{B6535C}{$-0.22$}
    & \textcolor[HTML]{B6535C}{$-0.68$}
    & \textcolor{black!60}{$0.00$} \\
    \bottomrule
  \end{tabular*}
\par
  \caption{\textbf{Video understanding interfaces at F8-R01.}
  Video-MME is evaluated without subtitles.
  $\Delta=\texttt{dense\_mode}-\texttt{sparse\_mode}$.}
  \label{tab:video_understanding_interfaces}
\end{table}

\begin{mdframed}[
  linecolor=genaccent!65!white,
  backgroundcolor=unifiedred!50!white,
  linewidth=0.8pt,
  roundcorner=3pt,
  shadow=true,
  shadowsize=3pt,
  shadowcolor=black!20,
  innerleftmargin=7pt,
  innerrightmargin=7pt,
  innertopmargin=5pt,
  innerbottommargin=5pt,
  skipabove=6pt,
  skipbelow=6pt]
\noindent\textbf{Takeaway 8:}
For video understanding, higher-FPS sampling (4 FPS) with temporal
compression performs comparably to 1-FPS sampling without temporal
compression.
\end{mdframed}

\FloatBarrier
\subsection{Benchmark Results}
\label{sec:benchmarks}
We report PixelUMM's results after Joint Stage~2 on image understanding
(Table~\ref{tab:image_understanding}), video understanding
(Table~\ref{tab:video_understanding}), image generation
(Table~\ref{tab:image_generation}), and video generation
(Tables~\ref{tab:video_generation} and~\ref{tab:video_generation_part2}),
where PixelUMM achieves overall performance comparable to the baselines.
Since training data differ across models, these results cannot establish
which architecture is superior or which model converges faster.

\begin{table}[H]
  \captionsetup{font={normalsize,stretch=1.15},skip=10pt}
  \centering
  \setlength{\tabcolsep}{2.6pt}
  \renewcommand{\arraystretch}{1.17}
  \resizebox{\textwidth}{!}{%
  \begin{tabular}{@{}l*{4}{>{\columncolor{unifiedblue}}c}*{5}{c}@{}}
    \toprule
    & \multicolumn{4}{c}{\cellcolor{unifiedblue}\textbf{Unified Models}}
    & \multicolumn{5}{c}{\cellcolor{baselinegray}\textbf{VLM}} \\
    \cmidrule(lr){2-5}\cmidrule(lr){6-10}
    \cellcolor{baselinegray}\textbf{Benchmark}
    & \makecell{\textbf{PixelUMM}\\[-1pt]\scriptsize 8B MoT}
    & \makecell{\textbf{BAGEL}\\[-1pt]\scriptsize 7B MoT}
    & \makecell{\textbf{TUNA}\\[-1pt]\scriptsize 7B+5B}
    & \makecell{\textbf{TUNA-2}\\[-1pt]\scriptsize 7B+5B}
    & \makecell{\textbf{Qwen2.5-VL}\\[-1pt]\scriptsize 7B}
    & \makecell{\textbf{LLaVA-OV-1.5}\\[-1pt]\scriptsize 8B}
    & \makecell{\textbf{LLaVA-OV-2}\\[-1pt]\scriptsize 8B}
    & \makecell{\textbf{Qwen3-VL-Inst.}\\[-1pt]\scriptsize 8B}
    & \makecell{\textbf{NEO-ov}\\[-1pt]\scriptsize 8B} \\
    \midrule
    MMMU        & 41.67 & 55.30 & 49.80 & 50.70 & 51.30 & 55.40 & --    & \bestscore{69.60} & \secondscore{68.10} \\
    MMStar      & 53.99 & --    & 61.20 & --    & 62.50 & \secondscore{67.70} & 64.30 & \bestscore{70.90} & 67.30 \\
    RWQA        & \secondscore{71.63} & \bestscore{72.80} & 66.10 & 67.70 & 68.50 & 68.10 & 69.70 & 71.50 & 67.80 \\
    SEED-I      & 70.39 & -- & 74.70 & -- & \bestscore{77.50} & \secondscore{77.30} & -- & -- & 76.60 \\
    AI2D        & 80.12 & \bestscore{89.20} & 79.30 & 79.60 & 82.60 & 84.20 & 84.30 & \secondscore{85.70} & 85.40 \\
    DocVQA      & 90.42 & -- & -- & -- & 94.90 & 95.00 & \secondscore{95.20} & \bestscore{96.10} & 91.90 \\
    ChartQA     & 82.96 & 78.50 & 85.80 & 85.60 & 84.10 & \secondscore{86.50} & 85.90 & \bestscore{89.60} & 86.20 \\
    InfoVQA     & 63.67 & -- & -- & -- & \secondscore{81.70} & 78.40 & 74.40 & \bestscore{83.10} & -- \\
    TextVQA     & \secondscore{78.76} & -- & -- & -- & \bestscore{84.90} & -- & -- & -- & 78.50 \\
    OCRBench    & 78.00 & 73.30 & 74.30 & 79.70 & \secondscore{84.20} & 82.90 & 78.20 & \bestscore{89.60} & 81.60 \\
    MME         & 1809.54 & \bestscore{2388.00} & -- & -- & \secondscore{2347.00} & -- & -- & -- & -- \\
    GQA         & 61.90 & \bestscore{66.40} & 63.90 & \secondscore{65.00} & 60.70 & -- & -- & -- & -- \\
    MMVP        & \secondscore{80.33} & \bestscore{85.00} & 70.70 & 77.30 & 78.00 & -- & -- & -- & -- \\
    SEED2+      & 64.51 & \bestscore{71.90} & 52.70 & 61.10 & \secondscore{70.90} & 69.20 & -- & -- & -- \\
    CV-Bench    & 77.60 & -- & -- & -- & \secondscore{80.00} & \bestscore{80.70} & -- & -- & -- \\
    CountBench  & \bestscore{94.30} & 82.50 & 73.50 & 81.70 & 86.40 & 88.20 & 89.00 & \secondscore{89.80} & -- \\
    PixMo-Count & \bestscore{73.03} & -- & -- & -- & 63.30 & 62.20 & \secondscore{64.00} & 62.40 & -- \\
    V$^*$       & 76.96 & 70.20 & 52.40 & 59.20 & 77.00 & 78.00 & \bestscore{85.90} & \secondscore{85.30} & -- \\
    MMMU-Pro    & 27.63 & -- & -- & -- & \secondscore{36.30} & \bestscore{37.40} & -- & -- & -- \\
    BLINK       & 53.46 & -- & -- & -- & 56.40 & 48.30 & \secondscore{63.50} & \bestscore{69.10} & 62.80 \\
    MuirBench   & 38.31 & -- & -- & -- & \secondscore{59.60} & -- & -- & \bestscore{64.40} & 58.20 \\
    \bottomrule
  \end{tabular}}
\par
  \caption{\textbf{Image understanding benchmarks.} Unified models are grouped first. PixelUMM is evaluated using the official LMMS-Eval protocol (64,750 generations over 21 tasks). Dark and light green denote the best and second-best results, respectively.}
  \label{tab:image_understanding}
\end{table}

\begin{table}[H]
  \captionsetup{font={normalsize,stretch=1.15},skip=10pt}

  \centering
  \setlength{\tabcolsep}{3.0pt}
  \renewcommand{\arraystretch}{1.12}
  \resizebox{\textwidth}{!}{%
  \begin{tabular}{@{}l*{4}{>{\columncolor{unifiedblue}}c}*{6}{c}@{}}
    \toprule
    & \multicolumn{4}{c}{\cellcolor{unifiedblue}\textbf{Unified Models}}
    & \multicolumn{6}{c}{\cellcolor{baselinegray}\textbf{VLM}} \\
    \cmidrule(lr){2-5}\cmidrule(lr){6-11}
    \cellcolor{baselinegray}\textbf{Benchmark}
    & \makecell{\textbf{PixelUMM}\\[-1pt]\scriptsize 8B MoT}
    & \makecell{\textbf{Show-o2}\\[-1pt]\scriptsize 1.5B+0.5B}
    & \makecell{\textbf{TUNA}\\[-1pt]\scriptsize 1.5B+?}
    & \makecell{\textbf{Lance}\\[-1pt]\scriptsize 3B MoT}
    & \makecell{\textbf{LLaVA-OV-2}\\[-1pt]\scriptsize 8B}
    & \makecell{\textbf{Qwen3-VL}\\[-1pt]\scriptsize 8B}
    & \makecell{\textbf{Keye-VL-1.5}\\[-1pt]\scriptsize 8B}
    & \makecell{\textbf{InternVL-3.5}\\[-1pt]\scriptsize 8B}
    & \makecell{\textbf{PLM}\\[-1pt]\scriptsize 8B}
    & \makecell{\textbf{LLaVA-OV-1.5}\\[-1pt]\scriptsize 8B} \\
    \midrule
    MVBench                    & 70.53 & 49.80 & 54.40 & 62.00 & 66.20 & 69.00 & 56.90 & \secondscore{72.10} & \bestscore{77.10} & 51.20 \\
    Video-MME (w/o sub.)       & 57.33 & 48.00 & 49.10 & --    & \secondscore{71.90} & 71.40 & \bestscore{73.00} & 65.90 & 60.50 & 61.10 \\
    LongVideoBench             & 59.61 & 49.20 & 49.70 & --    & \secondscore{66.90} & \bestscore{68.00} & 66.00 & 62.40 & 59.60 & 56.20 \\
    LVBench                    & 40.41 & --    & 27.40 & --    & \secondscore{55.50} & \bestscore{58.00} & 42.80 & 46.70 & 44.50 & 40.10 \\
    \bottomrule
  \end{tabular}}
\par
  \caption{\textbf{Video understanding benchmarks.} Unified models are grouped first. PixelUMM is evaluated in \texttt{sparse\_mode} with up to 96 sampled frames. Video-MME is evaluated without subtitles. Published models retain their original evaluation protocols. Dark and light green denote the best and second-best results, respectively.}
  \label{tab:video_understanding}
\end{table}

\begin{table}[H]
  \captionsetup{font={normalsize,stretch=1.15},skip=10pt}

  \centering
  \footnotesize
  \setlength{\tabcolsep}{0.7pt}
  \renewcommand{\arraystretch}{1.0}
  \begin{tabular*}{\textwidth}{@{\extracolsep{\fill}}l|c|ccccccc|cccccc@{}}
    \toprule
    \multirow{2}{*}{\textbf{Model}} & \multirow{2}{*}{\textbf{Size}}
    & \multicolumn{7}{c|}{\textbf{GenEval}}
    & \multicolumn{6}{c}{\textbf{DPG-Bench}} \\
    \cmidrule(lr){3-9}\cmidrule(lr){10-15}
    & & 1-Obj. & 2-Obj. & Count & Colors & Position & Col. Attr. & Overall
      & Global & Entity & Attribute & Relation & Other & Overall \\
    \midrule
    \rowcolor{baselinegray}
    \multicolumn{15}{c}{\cellcolor{baselinegray}\textbf{T2I Models}} \\
    SD3-M & 2B
      & 0.99 & 0.94 & 0.72 & 0.89 & 0.33 & 0.60 & 0.74
      & 87.90 & 91.01 & 89.96 & 80.70 & 88.68 & 84.08 \\
    FLUX.1 [dev]$^\dagger$ & 12B
      & 0.98 & 0.93 & 0.75 & 0.93 & 0.68 & 0.65 & 0.82
      & 82.10 & 89.50 & 88.70 & 91.10 & 89.40 & 84.00 \\
    LongCat-Image & 6B
      & 0.99 & 0.98 & 0.86 & 0.86 & 0.75 & 0.73 & 0.87
      & 89.10 & 92.54 & 92.00 & 93.28 & 87.50 & 86.80 \\
    Qwen-Image & 20B
      & 0.99 & 0.92 & 0.89 & 0.88 & 0.76 & 0.77 & 0.87
      & 91.32 & 91.56 & 92.02 & 94.31 & 92.73 & 88.32 \\
    Seedream 3.0 & --
      & 0.99 & 0.96 & 0.91 & 0.93 & 0.47 & 0.80 & 0.84
      & 94.31 & 92.65 & 91.36 & 92.78 & 88.24 & 88.27 \\
    Z-Image-Turbo & 6B
      & 1.00 & 0.95 & 0.77 & 0.89 & 0.65 & 0.68 & 0.82
      & 91.29 & 89.59 & 90.14 & 92.16 & 88.68 & 84.86 \\
    \midrule
    \rowcolor{unifiedred}
    \multicolumn{15}{c}{\cellcolor{unifiedred}\textbf{Unified Models}} \\
    Tar & 7B
      & \secondscore{0.99} & 0.92 & \secondscore{0.83} & 0.85 & 0.80 & 0.65 & 0.84
      & 83.98 & 88.62 & 88.05 & \bestscore{93.98} & 84.86 & 84.19 \\
    BLIP3-o & 8B
      & -- & -- & -- & -- & -- & -- & 0.84
      & -- & -- & -- & -- & -- & 81.60 \\
    UniWorld-V1$^\dagger$ & 12B
      & 0.98 & 0.93 & 0.81 & 0.89 & 0.74 & 0.71 & 0.84
      & 83.64 & 88.39 & 88.44 & 89.27 & 87.22 & 81.38 \\
    OmniGen2$^\dagger$ & 7B
      & \secondscore{0.99} & \secondscore{0.96} & 0.74 & \bestscore{0.98} & 0.71 & 0.75 & 0.86
      & 88.81 & 88.83 & 90.18 & 89.37 & 90.27 & 83.57 \\
    MUSE-VL & 7B
      & -- & -- & -- & -- & -- & -- & 0.57
      & -- & -- & -- & -- & -- & -- \\
    Transfusion & 7B
      & -- & -- & -- & -- & -- & -- & 0.63
      & -- & -- & -- & -- & -- & -- \\
    Emu3 & 8B
      & -- & -- & -- & -- & -- & -- & 0.66
      & -- & -- & -- & -- & -- & 81.60 \\
    Show-o2 & 7B+0.5B
      & \bestscore{1.00} & 0.87 & 0.58 & 0.92 & 0.52 & 0.62 & 0.76
      & 89.00 & 91.78 & 89.96 & 91.81 & \secondscore{91.64} & 86.14 \\
    Janus-Pro & 7B
      & \secondscore{0.99} & 0.89 & 0.59 & 0.90 & 0.79 & 0.66 & 0.80
      & 86.90 & 88.90 & 89.40 & 89.32 & 89.48 & 84.19 \\
    HBridge$^\dagger$ & 7B
      & \bestscore{1.00} & \secondscore{0.96} & 0.80 & 0.94 & 0.77 & 0.78 & 0.87
      & \bestscore{91.78} & \bestscore{91.82} & 90.23 & 90.06 & 88.42 & 85.23 \\
    Ming-UniVision & 16B
      & \bestscore{1.00} & 0.93 & 0.59 & 0.93 & \bestscore{0.92} & 0.70 & 0.85
      & -- & -- & -- & -- & -- & 82.12 \\
    BAGEL$^\dagger$ & 7B MoT
      & 0.98 & 0.95 & \bestscore{0.84} & 0.95 & 0.78 & 0.77 & 0.88
      & 88.94 & 90.37 & \secondscore{91.29} & 90.82 & 88.67 & 85.07 \\
    Mogao & 7B
      & \bestscore{1.00} & \bestscore{0.97} & \secondscore{0.83} & 0.93 & 0.84 & 0.80 & \secondscore{0.89}
      & 82.37 & 90.03 & 88.26 & 93.18 & 85.40 & 84.33 \\
    UniVideo$^\dagger$ & 7B+13B
      & 0.98 & 0.85 & 0.36 & 0.90 & 0.45 & 0.64 & 0.69
      & -- & -- & -- & -- & -- & -- \\
    Lance & 3B MoT
      & \bestscore{1.00} & 0.94 & \bestscore{0.84} & \secondscore{0.97} & 0.87 & \secondscore{0.81} & \bestscore{0.90}
      & 83.89 & 91.07 & 89.36 & 93.38 & 80.80 & 84.67 \\
    TUNA & 7B+5B
      & \bestscore{1.00} & \bestscore{0.97} & 0.81 & 0.91 & \secondscore{0.88} & \bestscore{0.83} & \bestscore{0.90}
      & \secondscore{90.42} & 91.68 & 90.94 & 91.87 & 90.73 & \bestscore{86.76} \\
    TUNA-R$^\dagger$ & 7B+5B
      & \bestscore{1.00} & 0.95 & 0.82 & 0.89 & 0.86 & 0.79 & 0.88
      & 86.00 & \secondscore{91.80} & 91.03 & \secondscore{93.48} & 84.89 & 86.35 \\
    TUNA-2 & 7B+5B
      & \secondscore{0.99} & \secondscore{0.96} & 0.80 & 0.91 & 0.84 & 0.76 & 0.87
      & 89.50 & 91.40 & \bestscore{92.07} & 91.91 & 88.81 & \secondscore{86.54} \\
    \midrule
    \textbf{PixelUMM} (orig.) & 8B MoT
      & \secondscore{0.99} & 0.95 & 0.65 & 0.84 & 0.55 & 0.64 & 0.77
      & 88.87 & 88.92 & 90.00 & 92.53 & \bestscore{92.62} & 85.74 \\
    \textbf{PixelUMM}$^\dagger$ & 8B MoT
      & 0.98 & 0.95 & 0.72 & 0.92 & 0.71 & 0.72 & 0.83
      & -- & -- & -- & -- & -- & -- \\
    \bottomrule
  \end{tabular*}
  \par\smallskip
  {\scriptsize Size notation: MoT denotes the backbone/expert scale;
  additive sizes separate the backbone and generation head. Counts are rounded;
  pretrained visual encoders/tokenizers are excluded. ``--'' denotes an unknown
  or undisclosed size.}
\par
  \caption{\textbf{Image generation benchmarks.} Results on GenEval and DPG-Bench. All multimodal systems are grouped as \emph{Unified Models}; generation-only systems are grouped as \emph{T2I Models}. ``Col. Attr.'' denotes color attribute, and $^\dagger$ marks reported use of an LLM prompt rewriter for GenEval. Baseline scores are taken directly from the original papers; the absence of $^\dagger$ does not confirm that prompt rewriting was not used. GenEval uses both the original prompts and BAGEL-long rewritten prompts ($^\dagger$), with 553 prompts and four samples per prompt. DPG-Bench follows the official 1,065-prompt protocol with four samples per prompt. Dark and light green denote the best and second-best results among~Unified~Models,~respectively.}
  \label{tab:image_generation}
\end{table}

\begin{table}[H]

  \centering
  \setlength{\tabcolsep}{1.4pt}
  \renewcommand{\arraystretch}{1.0}
  \footnotesize\linespread{1}\selectfont
  \begin{tabular*}{\textwidth}{@{\extracolsep{\fill}}l|c|cccccccccc@{}}
    \toprule
    \textbf{Model} & \textbf{Size}
      & \makecell{\textbf{Quality}\\\textbf{Score}}
      & \makecell{\textbf{Semantic}\\\textbf{Score}}
      & \makecell{\textbf{Subj.}\\\textbf{Consist.}}
      & \makecell{\textbf{Bkg.}\\\textbf{Consist.}}
      & \makecell{\textbf{Temp.}\\\textbf{Flicker}}
      & \makecell{\textbf{Motion}\\\textbf{Smooth.}}
      & \makecell{\textbf{Dynamic}\\\textbf{Degree}}
      & \makecell{\textbf{Aesthetic}\\\textbf{Quality}}
      & \makecell{\textbf{Imaging}\\\textbf{Quality}}
      & \makecell{\textbf{Object}\\\textbf{Class}} \\
    \midrule
    \rowcolor{baselinegray}
    \multicolumn{12}{c}{\cellcolor{baselinegray}\textbf{T2V Models}} \\
    ModelScope & 1.7B & 78.05 & 66.54 & 89.87 & 95.29 & 98.28 & 95.79 & 66.39 & 52.06 & 58.57 & 82.25 \\
    LaVie & 3B & 78.78 & 70.31 & 91.41 & 97.47 & 98.30 & 96.38 & 49.72 & 54.94 & 61.90 & 91.82 \\
    Show-1 & 6B & 80.42 & 72.98 & 95.53 & 98.02 & 99.12 & 98.24 & 44.44 & 57.35 & 58.66 & 93.07 \\
    AnimateDiff-V2 & 1.3B & 82.90 & 69.75 & 95.30 & 97.68 & 98.75 & 97.76 & 40.83 & 67.16 & 70.10 & 90.90 \\
    VideoCrafter-2.0 & 1.4B & 82.20 & 73.42 & 96.85 & 98.22 & 98.41 & 97.73 & 42.50 & 63.13 & 67.22 & 92.55 \\
    CogVideoX & 5B & 82.75 & 77.04 & 96.23 & 96.52 & 98.66 & 96.92 & 70.97 & 61.98 & 62.90 & 85.23 \\
    Kling & -- & 83.39 & 75.68 & 98.33 & 97.60 & 99.30 & 99.40 & 46.94 & 61.21 & 65.62 & 87.24 \\
    Open-Sora-2.0 & 11B & 82.10 & 80.14 & 98.75 & 98.00 & 99.40 & 99.49 & 20.74 & 64.33 & 65.62 & 94.50 \\
    Gen-3 & -- & 84.11 & 75.17 & 97.10 & 96.62 & 98.61 & 99.23 & 60.14 & 63.34 & 66.82 & 87.81 \\
    Step-Video-T2V & 30B & 84.46 & 71.28 & 98.05 & 97.67 & 99.40 & 99.08 & 53.06 & 61.23 & 70.63 & 80.56 \\
    HunyuanVideo & 13B & 85.07 & 76.88 & 97.22 & 97.60 & 99.39 & 99.05 & 71.94 & 60.28 & 67.24 & 83.48 \\
    Wan2.1-T2V & 14B & 85.59 & 76.11 & 97.52 & 98.09 & 99.46 & 98.30 & 65.46 & 66.07 & 69.43 & 86.28 \\
    \midrule
    \rowcolor{unifiedred}
    \multicolumn{12}{c}{\cellcolor{unifiedred}\textbf{Unified Models}} \\
    HaploOmni & 7B & -- & -- & 96.40 & \secondscore{97.60} & -- & 96.80 & 65.30 & -- & -- & -- \\
    Emu3 & 8B & -- & -- & 95.32 & \bestscore{97.69} & -- & \secondscore{98.93} & \bestscore{79.27} & 59.64 & -- & 86.17 \\
    VILA-U & 7B & 76.26 & 65.04 & -- & -- & -- & -- & -- & -- & -- & -- \\
    Show-o2 & 1.5B+0.5B & 82.10 & 78.31 & \bestscore{97.28} & 96.78 & 97.68 & 98.25 & 40.83 & 65.15 & 67.06 & 94.81 \\
    TUNA & 1.5B+? & 84.32 & \secondscore{83.04} & 95.99 & 96.72 & 98.02 & 98.33 & 69.39 & \secondscore{65.88} & 66.83 & 95.41 \\
    UniVideo$^\dagger$ & 7B+13B & \secondscore{84.36} & 79.96 & \secondscore{96.57} & 96.66 & 99.29 & \bestscore{99.15} & 49.72 & \bestscore{68.68} & \secondscore{67.32} & 94.59 \\
    Lance$^\dagger$ & 3B MoT & \bestscore{85.14} & \bestscore{84.96} & 94.52 & 94.28 & \bestscore{99.66} & 95.93 & \secondscore{75.83} & 64.33 & 66.78 & \bestscore{96.58} \\
    \midrule
    \textbf{PixelUMM}$^\dagger$ & 8B MoT & 84.10 & 79.80 & 95.05 & 97.15 & \secondscore{99.45} & 98.70 & 64.44 & 61.58 & \bestscore{67.63} & \secondscore{95.57} \\
    \bottomrule
  \end{tabular*}
  \par\smallskip
  {\scriptsize Size notation: MoT denotes the backbone/expert scale;
  additive sizes separate the backbone and generation head. Counts are rounded;
  pretrained visual encoders/tokenizers are excluded. ``--'' denotes an unknown
  or undisclosed size.}
\par
  \caption{\textbf{Video generation benchmarks on VBench Part 1.} Generation-only systems are grouped as \emph{T2V Models}; multimodal systems are grouped as \emph{Unified Models}. Baseline scores are taken directly from the original papers; the absence of $^\dagger$ does not confirm that prompt rewriting was not used. UniVideo and PixelUMM both use VBench GPT-enhanced prompts. Dark and light green denote the best and second-best results among Unified Models, respectively.}
  \label{tab:video_generation}
\end{table}

\begin{table}[H]

  \centering
  \setlength{\tabcolsep}{1.4pt}
  \renewcommand{\arraystretch}{1.0}
  \footnotesize\linespread{1}\selectfont
  \begin{tabular*}{\textwidth}{@{\extracolsep{\fill}}l|c|ccccccccc@{}}
    \toprule
    \textbf{Model} & \textbf{Size}
      & \makecell{\textbf{Multi.}\\\textbf{Objects}}
      & \makecell{\textbf{Human}\\\textbf{Action}}
      & \textbf{Color}
      & \makecell{\textbf{Spatial}\\\textbf{Relation}}
      & \textbf{Scene}
      & \makecell{\textbf{Appear.}\\\textbf{Style}}
      & \makecell{\textbf{Temp.}\\\textbf{Style}}
      & \makecell{\textbf{Overall}\\\textbf{Consist.}}
      & \makecell{\textbf{Total}\\\textbf{Score}$\uparrow$} \\
    \midrule
    \rowcolor{baselinegray}
    \multicolumn{11}{c}{\cellcolor{baselinegray}\textbf{T2V Models}} \\
    ModelScope & 1.7B & 38.98 & 92.40 & 81.72 & 33.68 & 39.26 & 23.39 & 25.37 & 25.67 & 75.75 \\
    LaVie & 3B & 33.32 & 96.80 & 86.39 & 34.09 & 52.69 & 23.56 & 25.93 & 26.41 & 77.08 \\
    Show-1 & 6B & 45.47 & 95.60 & 86.35 & 53.50 & 47.03 & 23.06 & 25.28 & 27.46 & 78.93 \\
    AnimateDiff-V2 & 1.3B & 36.88 & 92.60 & 87.47 & 34.60 & 50.19 & 22.42 & 26.03 & 27.04 & 80.27 \\
    VideoCrafter-2.0 & 1.4B & 40.66 & 95.00 & 92.92 & 35.86 & 55.29 & 25.13 & 25.84 & 28.23 & 80.44 \\
    CogVideoX & 5B & 62.11 & 99.40 & 82.81 & 66.35 & 53.20 & 24.91 & 25.38 & 27.59 & 81.61 \\
    Kling & -- & 68.05 & 93.40 & 89.90 & 73.03 & 50.86 & 19.62 & 24.17 & 26.42 & 81.85 \\
    Open-Sora-2.0 & 11B & 77.72 & 95.40 & 85.98 & 76.18 & 52.71 & 22.98 & 25.91 & 27.57 & 81.71 \\
    Gen-3 & -- & 53.64 & 96.40 & 80.90 & 65.09 & 54.57 & 24.31 & 24.71 & 26.69 & 82.32 \\
    Step-Video-T2V & 30B & 50.55 & 94.00 & 88.25 & 71.47 & 24.38 & 23.17 & 26.01 & 27.12 & 81.83 \\
    HunyuanVideo & 13B & 66.71 & 94.40 & 89.79 & 72.13 & 54.46 & 22.21 & 24.52 & 26.95 & 83.43 \\
    Wan2.1-T2V & 14B & 69.58 & 95.40 & 88.59 & 75.39 & 45.75 & 22.64 & 23.19 & 25.91 & 83.69 \\
    \midrule
    \rowcolor{unifiedred}
    \multicolumn{11}{c}{\cellcolor{unifiedred}\textbf{Unified Models}} \\
    HaploOmni & 7B & -- & -- & -- & -- & 34.60 & -- & -- & -- & 78.10 \\
    Emu3 & 8B & 44.64 & 77.71 & -- & 68.73 & 37.11 & 20.92 & -- & -- & 80.96 \\
    VILA-U & 7B & -- & -- & -- & -- & -- & -- & -- & -- & 74.01 \\
    Show-o2 & 1.5B+0.5B & 76.01 & 95.20 & 80.89 & 62.61 & 57.67 & 23.29 & 25.27 & 27.00 & 81.34 \\
    TUNA & 1.5B+? & \secondscore{92.31} & \secondscore{97.50} & \secondscore{87.67} & \secondscore{78.12} & \secondscore{58.59} & 23.18 & 24.68 & \secondscore{27.71} & \secondscore{84.06} \\
    UniVideo$^\dagger$ & 7B+13B & 82.06 & 97.40 & 82.83 & 76.43 & 48.74 & \bestscore{24.92} & 25.06 & 26.38 & 83.48 \\
    Lance$^\dagger$ & 3B MoT & \bestscore{93.86} & \bestscore{97.80} & \bestscore{92.61} & \bestscore{93.61} & \bestscore{64.75} & 23.14 & \secondscore{25.53} & 27.04 & \bestscore{85.11} \\
    \midrule
    \textbf{PixelUMM}$^\dagger$ & 8B MoT & 76.75 & 97.40 & 84.26 & 68.11 & 53.94 & \secondscore{23.68} & \bestscore{25.84} & \bestscore{27.90} & 83.24 \\
    \bottomrule
  \end{tabular*}
  \par\smallskip
  {\scriptsize Size notation: MoT denotes the backbone/expert scale;
  additive sizes separate the backbone and generation head. Counts are rounded;
  pretrained visual encoders/tokenizers are excluded. ``--'' denotes an unknown
  or undisclosed size.}
\par
  \caption{\textbf{Video generation benchmarks on VBench Part 2.} Generation-only systems are grouped as \emph{T2V Models}; multimodal systems are grouped as \emph{Unified Models}. Baseline scores are taken directly from the original papers; the absence of $^\dagger$ does not confirm that prompt rewriting was not used. UniVideo and PixelUMM both use VBench GPT-enhanced prompts. Dark and light green denote the best and second-best results among Unified Models, respectively.}
  \label{tab:video_generation_part2}
\end{table}

\FloatBarrier
\section{Related Work}
\label{sec:related}

\noindent\textbf{Unified multimodal models.}
Native multimodal understanding models train on mixed-modal inputs while producing text~\cite{li2025ariaopenmultimodalnative}.
Discrete-token models unify language and vision through autoregressive sequence modeling~\cite{aghajanyan2022cm3causalmaskedmultimodal,io2,Chameleon,wang2024emu3,zhang2026nextflow,lu2022unifiediounifiedmodelvision,cui2025emu35nativemultimodalmodels,geng2025x,li2025onecat,chern2024anole,team2026longcat,wu2024liquid,wu2024vila}.
Continuous-representation models also support both multimodal understanding and image generation~\cite{emu1,cao2025hunyuanimage,fan2025unified}.
Modular models connect pretrained understanding and generation components through learned interfaces~\cite{dream-llm,emu2,ge2024seed,tong2024metamorph,pan2025transfer,wunext,zheng2023minigpt,wu2025openuni}.
Layerwise fusion connects understanding and generation experts at intermediate representations~\cite{wang2025lightfusion,xu2025tbacuniimageunifiedunderstandinggeneration,wang2025hbridge}.
Hybrid models pair language autoregression with visual diffusion or flow matching~\cite{zhou2024transfusion,xie2024show,xie2025show,ma2024janusflow,deng2025bagel,liu2025tuna,tong2026beyond,liao2025mogao,wang2025ovis,xiao2025mindomniunleashingreasoninggeneration,shen2025mammothmoda2}.
Unified architectures vary in their visual representations and the sharing of model parameters~\cite{wu2024janus,chen2025janus,MoT,MOMA,shi2024llamafusion,hao2025uni,li2025synergen,he2025emma,tian2026internvludemocratizingunifiedmultimodal,ai2025ming}.
Diffusion- and flow-based unified models provide alternatives to autoregressive text decoding~\cite{yang2025mmada,xin2025lumina,you2026llada,nguyen2025oneflowconcurrentmixedmodalinterleaved,li2025dual,luo2025next,shi2025muddit}.
Recent studies directly test transfer between understanding and generation~\cite{niu2025does,zhang2026crosstaskgeneralizationunderstandinggeneration,shi2025realunify}.
Dedicated evaluations probe knowledge and reasoning in visual generation and unified models~\cite{niu2025wise,li2026uevalbenchmarkunifiedmultimodal,yang2026ureasonbenchmarkingreasoningtogenerationalignment}.

\noindent\textbf{Visual representations.}
Encoder-based vision-language models couple pretrained visual representations to language models~\cite{flamingo,blip2,liu2023llava,Qwen2.5-VL,chen2024internvl}.
Native-resolution encoders use patch packing to accommodate variable image sizes and aspect ratios~\cite{dehghani2023patch}.
Language-supervised visual encoders provide semantic interfaces for understanding~\cite{radford2021learning,VLP:SigLIP,tschannen2025siglip,sun2023eva,sun2024evaclip18bscalingclip18,xu2025demystifyingclipdata}.
Self-supervised objectives learn visual representations without paired language supervision~\cite{chen2020simpleframeworkcontrastivelearning,he2020momentumcontrastunsupervisedvisual,caron2021emerging,he2022masked,oquab2023dinov2,simeoni2025dinov3,fan2023motionguidedmaskingspatiotemporalrepresentation,fan2025scalinglanguagefreevisualrepresentation,xu2025nextembeddingpredictionmakesstrong}.
Discrete visual tokenizers compress images into codes for generative modeling~\cite{VQ_VAE,razavi2019generatingdiversehighfidelityimages,lee2022autoregressive,esser2021taming,yu2022vectorquantized,mentzer2023finite}.
Variational autoencoders instead provide continuous latent spaces for image synthesis~\cite{VAE,LatentDiffusion}.
Semantic discrete tokenizers align visual codes with language while retaining reconstructable image information~\cite{qu2025tokenflow,ma2025unitok,TokLIP,QLIP,Dualtoken,ge2023making,xie2024muse,han2025vision,wang2024illume,huang2025illume+}.
Shared continuous visual representations support both understanding and generation~\cite{yue2025uniflow,tang2025unilip,fan2025prism,zhang2026openvision,huang2025ming,wu2025harmonizing}.
Representation alignment and semantic autoencoders support generation from pretrained visual features~\cite{yu2024representation,chen2025vugenvisualunderstandingpriors,zheng2025diffusion,tong2026scalingtexttoimagediffusiontransformers,AlignTok,shi2025latent,chen2025blip3o,lin2025uniworld}.
Alignment and denoising objectives improve latent representations for visual generation~\cite{REPA-E,vavae,yang2025latent}.
Reconstruction-based supervision connects visual understanding with generation~\cite{wang2024reconstructive,wang2025autoregressive,ma2025genhancer,xie2025reconstruction}.
TUNA and Beyond Language Modeling examine representation choice within joint multimodal pretraining~\cite{liu2025tuna,tong2026beyond}.

\noindent\textbf{Pixel-space modeling.}
Encoder-free vision-language models directly embed image patches for language modeling~\cite{VLM:Fuyu-8b,diao2024EVE,diao2025EVEv2,Diao2025NEO,VLM:SOLO,VLM:Mono-InternVL-1.5,diao2026neov}.
Other monolithic designs integrate visual processing through shared modules, visual experts, learned queries, or adapted language-model blocks~\cite{VLM:Mono-InternVL,VLM:BREEN,VLM:HaploVL,VLM:HoVLE,VLM:SAIL,VLM:VoRA}.
Pixel-space diffusion and flow models remove the need for latent image compression~\cite{DDPM,PixelFlow,wang2025pixnerd,Dip,yu2025pixeldit,li2025back}.
These contrast with synthesis in learned latent spaces~\cite{LatentDiffusion,DiT,SD3,SDXL,flux2024,xie2025sana15efficientscaling}.
Recent empirical work studies transferring latent-space generative priors to pixel-space text-to-image models during post-training~\cite{jiang2026empiricalpixelspace}.
TUNA-2 and the NEO-unify/SenseNova series extend pixel-space learning to unified understanding and generation~\cite{tuna2,sensenova2026neounify,diao2026sensenovau1,diao2026sensenovau15nativeunifiedvisual}.
HiDream-O1-Image also integrates image generation and editing through a pixel-level architecture~\cite{cai2026hidream}.

\noindent\textbf{Video generation and unification.}
Video diffusion and flow models provide task-specific visual generators~\cite{VideoDiffusionModels,blattmann2023stable,CogVideoX,kong2024hunyuanvideo,MovieGen,wan2025wan}.
Cosmos and Cosmos-Predict2 develop video foundation models for physical-world generation~\cite{agarwal2025cosmos,nvidia2025cosmospredict2}.
Video-language models encode temporal visual inputs for language-based understanding~\cite{lin2024videollava,cheng2024videollama}.
Omni-Video, UniVid, and UniVideo combine video understanding with generation~\cite{tan2025omni,luo2025univid,wei2025univideo}.
Cosmos~3 unifies multimodal understanding and generation within a mixture-of-transformers architecture~\cite{nvidia2026cosmos3}.
Joint-pretraining models extend unified learning across images and videos~\cite{liu2025tuna,xie2025show,fu2026lance,tong2026beyond,wu2024vila,luo2025next}.

Bringing together these directions in pixel-space modeling and
image-video unification, we investigate a shared raw-pixel interface for
joint understanding and generation. PixelUMM represents images as spatial
patches and videos as spatiotemporal tubelets, connecting both directly to
a shared multimodal backbone without a pretrained visual encoder or VAE.

\section{Conclusion}
\label{sec:conclusion}

We presented PixelUMM, an encoder-free unified model for image and video understanding and generation directly in pixel space.
By connecting spatial image patches and spatiotemporal video tubelets to a shared multimodal backbone through lightweight linear projections, PixelUMM extends pixel-space unified modeling to video without pretrained visual encoders or VAE-based latent tokenizers.
Experiments show performance comparable to state-of-the-art baselines across image and video understanding and generation tasks.
Our experiments suggest that less aggressive spatial and temporal
token compression generally lowers generation loss, while convolutional heads
can suppress patch artifacts. PixelUMM still uses separate understanding and
generation Transformer experts with shared attention, rather than fully
unified parameters. Future work could investigate MoE architectures that more
fully integrate these capabilities and test whether pixel-space modeling can
support larger spatial patches ($p=64$) and longer temporal tubelets
($\tau=8$) without compromising training or generation quality.

{
\small
\linespread{1.0}\selectfont
\setlength{\bibsep}{0.5pt plus .5pt minus .5pt}

}

\begin{thebibliography}{100}

\bibitem{aghajanyan2022cm3causalmaskedmultimodal}
A.~Aghajanyan, B.~Huang, C.~Ross, V.~Karpukhin, H.~Xu, N.~Goyal, D.~Okhonko,
  M.~Joshi, G.~Ghosh, M.~Lewis, and L.~Zettlemoyer.
\newblock Cm3: A causal masked multimodal model of the internet.
\newblock {\em arXiv preprint arXiv:2201.07520}, 2022.

\bibitem{ai2025ming}
I.~AI, B.~Ma, C.~Zou, C.~Yan, C.~Jin, C.~Shen, C.~Lian, D.~Zheng, F.~Wang,
  F.~Xu, et~al.
\newblock Ming-flash-omni: A sparse, unified architecture for multimodal
  perception and generation.
\newblock {\em arXiv preprint arXiv:2510.24821}, 2025.

\bibitem{flamingo}
J.-B. Alayrac, J.~Donahue, P.~Luc, A.~Miech, I.~Barr, Y.~Hasson, K.~Lenc,
  A.~Mensch, K.~Millican, M.~Reynolds, et~al.
\newblock Flamingo: a visual language model for few-shot learning.
\newblock {\em Advances in Neural Information Processing Systems},
  35:23716--23736, 2022.

\bibitem{Qwen2.5-VL}
S.~Bai, K.~Chen, X.~Liu, J.~Wang, W.~Ge, S.~Song, K.~Dang, P.~Wang, S.~Wang,
  J.~Tang, H.~Zhong, Y.~Zhu, M.~Yang, Z.~Li, J.~Wan, P.~Wang, W.~Ding, Z.~Fu,
  Y.~Xu, J.~Ye, X.~Zhang, T.~Xie, Z.~Cheng, H.~Zhang, Z.~Yang, H.~Xu, and
  J.~Lin.
\newblock Qwen2.5-vl technical report.
\newblock {\em arXiv preprint arXiv:2502.13923}, 2025.

\bibitem{VLM:Fuyu-8b}
R.~Bavishi, E.~Elsen, C.~Hawthorne, M.~Nye, A.~Odena, A.~Somani, and
  S.~Ta\c{s}\i{}rlar.
\newblock Introducing our multimodal models, 2023.

\bibitem{flux2024}
{Black Forest Labs}.
\newblock Flux.
\newblock \url{https://github.com/black-forest-labs/flux}, 2024.

\bibitem{blattmann2023stable}
A.~Blattmann, T.~Dockhorn, S.~Kulal, D.~Mendelevitch, M.~Kilian, D.~Lorenz,
  Y.~Levi, Z.~English, V.~Voleti, A.~Letts, et~al.
\newblock Stable video diffusion: Scaling latent video diffusion models to
  large datasets.
\newblock {\em arXiv preprint arXiv:2311.15127}, 2023.

\bibitem{cai2026hidream}
Q.~Cai, J.~Chen, C.~Gao, Z.~Gong, Y.~Li, Y.~Pan, Y.~Peng, Z.~Qiu, K.~Yu,
  Y.~Zhang, et~al.
\newblock Hidream-o1-image: A natively unified image generative foundation
  model with pixel-level unified transformer.
\newblock {\em arXiv preprint arXiv:2605.11061}, 2026.

\bibitem{cao2025hunyuanimage}
S.~Cao, H.~Chen, P.~Chen, Y.~Cheng, Y.~Cui, X.~Deng, Y.~Dong, K.~Gong, T.~Gu,
  X.~Gu, et~al.
\newblock Hunyuanimage 3.0 technical report.
\newblock {\em arXiv preprint arXiv:2509.23951}, 2025.

\bibitem{caron2021emerging}
M.~Caron, H.~Touvron, I.~Misra, H.~J{\'e}gou, J.~Mairal, P.~Bojanowski, and
  A.~Joulin.
\newblock Emerging properties in self-supervised vision transformers.
\newblock In {\em ICCV}, 2021.

\bibitem{AlignTok}
B.~Chen, S.~Bi, H.~Tan, H.~Zhang, T.~Zhang, Z.~Li, Y.~Xiong, J.~Zhang, and
  K.~Zhang.
\newblock Aligning visual foundation encoders to tokenizers for diffusion
  models.
\newblock {\em arXiv preprint arXiv:2509.25162}, 2025.

\bibitem{chen2025blip3o}
J.~Chen, Z.~Xu, X.~Pan, Y.~Hu, C.~Qin, T.~Goldstein, L.~Huang, T.~Zhou, S.~Xie,
  S.~Savarese, L.~Xue, C.~Xiong, and R.~Xu.
\newblock Blip3-o: A family of fully open unified multimodal
  models-architecture, training and dataset.
\newblock {\em arXiv preprint arXiv:2505.09568}, 2025.

\bibitem{PixelFlow}
S.~Chen, C.~Ge, S.~Zhang, P.~Sun, and P.~Luo.
\newblock Pixelflow: Pixel-space generative models with flow.
\newblock {\em arXiv preprint arXiv:2504.07963}, 2025.

\bibitem{chen2020simpleframeworkcontrastivelearning}
T.~Chen, S.~Kornblith, M.~Norouzi, and G.~Hinton.
\newblock A simple framework for contrastive learning of visual
  representations.
\newblock {\em arXiv preprint arXiv:2002.05709}, 2020.

\bibitem{chen2025vugenvisualunderstandingpriors}
X.~Chen, T.~Vallaeys, M.~Elbayad, J.~Nguyen, and J.~Verbeek.
\newblock Vugen: Visual understanding priors for generation.
\newblock {\em arXiv preprint arXiv:2510.06529}, 2025.

\bibitem{chen2025janus}
X.~Chen, Z.~Wu, X.~Liu, Z.~Pan, W.~Liu, Z.~Xie, X.~Yu, and C.~Ruan.
\newblock Janus-pro: Unified multimodal understanding and generation with data
  and model scaling.
\newblock {\em arXiv preprint arXiv:2501.17811}, 2025.

\bibitem{VLM:SOLO}
Y.~Chen, X.~Wang, H.~Peng, and H.~Ji.
\newblock A single transformer for scalable vision-language modeling.
\newblock {\em Transactions on Machine Learning Research}, 2024.

\bibitem{chen2024internvl}
Z.~Chen, J.~Wu, W.~Wang, W.~Su, G.~Chen, S.~Xing, M.~Zhong, Q.~Zhang, X.~Zhu,
  L.~Lu, et~al.
\newblock Internvl: Scaling up vision foundation models and aligning for
  generic visual-linguistic tasks.
\newblock In {\em Proceedings of the IEEE/CVF Conference on Computer Vision and
  Pattern Recognition,}, pages 24185--24198, 2024.

\bibitem{Dip}
Z.~Chen, J.~Zhu, X.~Chen, J.~Zhang, X.~Hu, H.~Zhao, C.~Wang, J.~Yang, and
  Y.~Tai.
\newblock Dip: Taming diffusion models in pixel space.
\newblock {\em arXiv preprint arXiv:2511.18822}, 2025.

\bibitem{cheng2024videollama}
Z.~Cheng, S.~Leng, H.~Zhang, Y.~Xin, X.~Li, G.~Chen, Y.~Zhu, W.~Zhang, Z.~Luo,
  D.~Zhao, et~al.
\newblock Videollama 2: Advancing spatial-temporal modeling and audio
  understanding in video-llms.
\newblock {\em arXiv preprint arXiv:2406.07476}, 2024.

\bibitem{chern2024anole}
E.~Chern, J.~Su, Y.~Ma, and P.~Liu.
\newblock Anole: An open, autoregressive, native large multimodal models for
  interleaved image-text generation.
\newblock {\em arXiv preprint arXiv:2407.06135}, 2024.

\bibitem{cui2025emu35nativemultimodalmodels}
Y.~Cui, H.~Chen, H.~Deng, X.~Huang, X.~Li, J.~Liu, Y.~Liu, Z.~Luo, J.~Wang,
  W.~Wang, et~al.
\newblock Emu3. 5: Native multimodal models are world learners.
\newblock {\em arXiv preprint arXiv:2510.26583}, 2025.

\bibitem{dao2023flashattention2}
T.~Dao.
\newblock {FlashAttention-2}: Faster attention with better parallelism and work
  partitioning.
\newblock {\em arXiv preprint arXiv:2307.08691}, 2023.

\bibitem{dao2022flashattention}
T.~Dao, D.~Y. Fu, S.~Ermon, A.~Rudra, and C.~R{\'e}.
\newblock {FlashAttention}: Fast and memory-efficient exact attention with
  {IO}-awareness.
\newblock {\em Advances in Neural Information Processing Systems},
  35:16344--16359, 2022.

\bibitem{dehghani2023patch}
M.~Dehghani, B.~Mustafa, J.~Djolonga, J.~Heek, M.~Minderer, M.~Caron,
  A.~Steiner, J.~Puigcerver, R.~Geirhos, I.~M. Alabdulmohsin, et~al.
\newblock Patch n’pack: Navit, a vision transformer for any aspect ratio and
  resolution.
\newblock In {\em NeurIPS}, 2023.

\bibitem{deng2025bagel}
C.~Deng, D.~Zhu, K.~Li, C.~Gou, F.~Li, Z.~Wang, S.~Zhong, W.~Yu, X.~Nie,
  Z.~Song, G.~Shi, and H.~Fan.
\newblock Emerging properties in unified multimodal pretraining.
\newblock {\em arXiv preprint arXiv:2505.14683}, 2025.

\bibitem{diao2024EVE}
H.~Diao, Y.~Cui, X.~Li, Y.~Wang, H.~Lu, and X.~Wang.
\newblock Unveiling encoder-free vision-language models.
\newblock {\em arXiv preprint arXiv:2406.11832}, 2024.

\bibitem{Diao2025NEO}
H.~Diao, M.~Li, S.~Wu, L.~Dai, X.~Wang, H.~Deng, L.~Lu, D.~Lin, and Z.~Liu.
\newblock From pixels to words--towards native vision-language primitives at
  scale.
\newblock {\em arXiv preprint arXiv:2510.14979}, 2025.

\bibitem{diao2025EVEv2}
H.~Diao, X.~Li, Y.~Cui, Y.~Wang, H.~Deng, T.~Pan, W.~Wang, H.~Lu, and X.~Wang.
\newblock Evev2: Improved baselines for encoder-free vision-language models.
\newblock {\em arXiv preprint arXiv:2502.06788}, 2025.

\bibitem{diao2026sensenovau15nativeunifiedvisual}
H.~Diao, J.~Wang, C.~Ding, H.~Deng, J.~Chen, R.~Zhang, R.~Wang, W.~Tong,
  X.~Fan, Y.~Wang, Y.~Zhu, Y.~Niu, Z.~Bai, Z.~Lin, Z.~Yang, Z.~Cai, B.~Yang,
  C.~Feng, C.~Lv, G.~Liu, G.~Wang, H.~Zhang, H.~Yu, H.~Xiao, H.~Wang, H.~Wu,
  H.~Zhong, J.~Fang, J.~Fan, J.~Li, J.~Lu, J.~Zuo, J.~Ni, J.~Xu, L.~Dai, M.~Xu,
  P.~Yan, P.~Wu, R.~Mao, R.~Wang, S.~Bai, S.~Yang, S.~Yang, S.~Zheng, S.~Wu,
  S.~Li, T.~Chu, T.~Zhong, T.~Zhou, W.~Luo, W.~Fan, W.~Jia, W.~Gao, X.~Kong,
  Y.~Li, Y.~Yong, Z.~Wen, Z.~Qian, W.~Sun, R.~Gong, Q.~Wang, L.~Lu, L.~Yang,
  Z.~Liu, and D.~Lin.
\newblock Sensenova-u1.5: Towards native unified visual intelligence.
\newblock {\em arXiv preprint arXiv:2609.11929}, 2026.

\bibitem{diao2026neov}
H.~Diao, J.~Wang, P.~Wu, Y.~Dong, Y.~Niu, Y.~Zhu, Z.~Cai, W.~Fan, L.~Dai,
  S.~Wu, X.~Zheng, M.~Li, Y.~Zhang, B.~Li, H.~Deng, H.~Lu, Q.~Wang, L.~Yang,
  L.~Lu, D.~Lin, and Z.~Liu.
\newblock From pixels to words--towards native one-vision models at scale.
\newblock {\em arXiv preprint arXiv:2605.28820}, 2026.

\bibitem{diao2026sensenovau1}
H.~Diao, P.~Wu, H.~Deng, J.~Wang, S.~Bai, S.~Wu, W.~Fan, W.~Ye, W.~Tong,
  X.~Fan, et~al.
\newblock {SenseNova-U1}: Unifying multimodal understanding and generation with
  {NEO-unify} architecture.
\newblock {\em arXiv preprint arXiv:2605.12500}, 2026.

\bibitem{dream-llm}
R.~Dong, C.~Han, Y.~Peng, Z.~Qi, Z.~Ge, J.~Yang, L.~Zhao, J.~Sun, H.~Zhou,
  H.~Wei, et~al.
\newblock Dreamllm: Synergistic multimodal comprehension and creation.
\newblock In {\em ICLR}, 2024.

\bibitem{ViT}
A.~Dosovitskiy, L.~Beyer, A.~Kolesnikov, D.~Weissenborn, X.~Zhai,
  T.~Unterthiner, M.~Dehghani, M.~Minderer, G.~Heigold, S.~Gelly, et~al.
\newblock An image is worth 16x16 words: Transformers for image recognition at
  scale.
\newblock In {\em Proc. ICLR}, 2021.

\bibitem{SD3}
P.~Esser, S.~Kulal, A.~Blattmann, R.~Entezari, J.~M{\"u}ller, H.~Saini,
  Y.~Levi, D.~Lorenz, A.~Sauer, F.~Boesel, et~al.
\newblock Scaling rectified flow transformers for high-resolution image
  synthesis.
\newblock In {\em Proc. ICML}, 2024.

\bibitem{esser2021taming}
P.~Esser, R.~Rombach, and B.~Ommer.
\newblock Taming transformers for high-resolution image synthesis.
\newblock In {\em Proceedings of the IEEE/CVF conference on computer vision and
  pattern recognition}, pages 12873--12883, 2021.

\bibitem{fan2025scalinglanguagefreevisualrepresentation}
D.~Fan, S.~Tong, J.~Zhu, K.~Sinha, Z.~Liu, X.~Chen, M.~Rabbat, N.~Ballas,
  Y.~LeCun, A.~Bar, and S.~Xie.
\newblock Scaling language-free visual representation learning.
\newblock {\em arXiv preprint arXiv:2504.01017}, 2025.

\bibitem{fan2023motionguidedmaskingspatiotemporalrepresentation}
D.~Fan, J.~Wang, S.~Liao, Y.~Zhu, V.~Bhat, H.~Santos-Villalobos, R.~MV, and
  X.~Li.
\newblock Motion-guided masking for spatiotemporal representation learning.
\newblock {\em arXiv preprint arXiv:2308.12962}, 2023.

\bibitem{fan2025unified}
L.~Fan, L.~Tang, S.~Qin, T.~Li, X.~Yang, S.~Qiao, A.~Steiner, C.~Sun, Y.~Li,
  T.~Zhu, et~al.
\newblock Unified autoregressive visual generation and understanding with
  continuous tokens.
\newblock {\em arXiv preprint arXiv:2503.13436}, 2025.

\bibitem{fan2025prism}
W.~Fan, H.~Diao, Q.~Wang, D.~Lin, and Z.~Liu.
\newblock The prism hypothesis: Harmonizing semantic and pixel representations
  via unified autoencoding.
\newblock {\em arXiv preprint arXiv:2512.19693}, 2025.

\bibitem{fu2026lance}
F.~Fu, M.~Huang, S.~Wu, Y.~Jiang, Y.~Huo, H.~Li, Y.~Song, F.~Ding, J.~Guo,
  Q.~He, Z.~Fu, Z.~Mao, and Y.~Zhang.
\newblock Lance: Unified multimodal modeling by multi-task synergy.
\newblock {\em arXiv preprint arXiv:2605.18678}, 2026.

\bibitem{ge2023making}
Y.~Ge, S.~Zhao, Z.~Zeng, Y.~Ge, C.~Li, X.~Wang, and Y.~Shan.
\newblock Making llama see and draw with seed tokenizer.
\newblock {\em arXiv preprint arXiv:2310.01218}, 2023.

\bibitem{ge2024seed}
Y.~Ge, S.~Zhao, J.~Zhu, Y.~Ge, K.~Yi, L.~Song, C.~Li, X.~Ding, and Y.~Shan.
\newblock Seed-x: Multimodal models with unified multi-granularity
  comprehension and generation.
\newblock {\em arXiv preprint arXiv:2404.14396}, 2024.

\bibitem{geng2025x}
Z.~Geng, Y.~Wang, Y.~Ma, C.~Li, Y.~Rao, S.~Gu, Z.~Zhong, Q.~Lu, H.~Hu,
  X.~Zhang, et~al.
\newblock X-omni: Reinforcement learning makes discrete autoregressive image
  generative models great again.
\newblock {\em arXiv preprint arXiv:2507.22058}, 2025.

\bibitem{han2025vision}
J.~Han, H.~Chen, Y.~Zhao, H.~Wang, Q.~Zhao, Z.~Yang, H.~He, X.~Yue, and
  L.~Jiang.
\newblock Vision as a dialect: Unifying visual understanding and generation via
  text-aligned representations.
\newblock {\em arXiv preprint arXiv:2506.18898}, 2025.

\bibitem{hao2025uni}
J.~Hao, H.~Liu, X.~Xiao, Q.~Huang, and J.~Yu.
\newblock Uni-x: Mitigating modality conflict with a two-end-separated
  architecture for unified multimodal models.
\newblock {\em arXiv preprint arXiv:2509.24365}, 2025.

\bibitem{he2022masked}
K.~He, X.~Chen, S.~Xie, Y.~Li, P.~Doll{\'a}r, and R.~Girshick.
\newblock Masked autoencoders are scalable vision learners.
\newblock In {\em Proceedings of the IEEE/CVF conference on computer vision and
  pattern recognition}, pages 16000--16009, 2022.

\bibitem{he2020momentumcontrastunsupervisedvisual}
K.~He, H.~Fan, Y.~Wu, S.~Xie, and R.~Girshick.
\newblock Momentum contrast for unsupervised visual representation learning.
\newblock {\em arXiv preprint arXiv:1911.05722}, 2019.

\bibitem{he2025emma}
X.~He, L.~Wei, J.~Ouyang, M.~Liao, L.~Xie, and Q.~Tian.
\newblock Emma: Efficient multimodal understanding, generation, and editing
  with a unified architecture.
\newblock {\em arXiv preprint arXiv:2512.04810}, 2025.

\bibitem{DDPM}
J.~Ho, A.~Jain, and P.~Abbeel.
\newblock Denoising diffusion probabilistic models.
\newblock In {\em Proc. NeurIPS}, 2020.

\bibitem{VideoDiffusionModels}
J.~Ho, T.~Salimans, A.~Gritsenko, W.~Chan, M.~Norouzi, and D.~J. Fleet.
\newblock Video diffusion models.
\newblock {\em Proc. NeurIPS}, 2022.

\bibitem{huang2025vipe}
J.~Huang, Q.~Zhou, H.~Rabeti, A.~Korovko, H.~Ling, X.~Ren, T.~Shen, J.~Gao,
  D.~Slepichev, C.-H. Lin, et~al.
\newblock Vipe: Video pose engine for 3d geometric perception.
\newblock {\em arXiv preprint arXiv:2508.10934}, 2025.

\bibitem{huang2025illume+}
R.~Huang, C.~Wang, J.~Yang, G.~Lu, Y.~Yuan, J.~Han, L.~Hou, W.~Zhang, L.~Hong,
  H.~Zhao, et~al.
\newblock Illume+: Illuminating unified mllm with dual visual tokenization and
  diffusion refinement.
\newblock {\em arXiv preprint arXiv:2504.01934}, 2025.

\bibitem{huang2025ming}
Z.~Huang, D.~Zheng, C.~Zou, R.~Liu, X.~Wang, K.~Ji, W.~Chai, J.~Sun, L.~Wang,
  Y.~Lv, et~al.
\newblock Ming-univision: Joint image understanding and generation with a
  unified continuous tokenizer.
\newblock {\em arXiv preprint arXiv:2510.06590}, 2025.

\bibitem{jiang2026empiricalpixelspace}
D.~Jiang, R.~Du, Z.~Chen, D.~Liu, Z.~Wang, M.~Zheng, X.~Yang, H.~Cai, A.~Hao,
  Y.~Jiang, P.~Gao, H.~Yang, and S.~Hoi.
\newblock An empirical study of training pixel-space text-to-image diffusion
  models.
\newblock {\em arXiv preprint arXiv:2608.16887}, 2026.

\bibitem{VAE}
D.~P. Kingma and M.~Welling.
\newblock Auto-encoding variational bayes.
\newblock {\em arXiv preprint arXiv:1312.6114}, 2013.

\bibitem{kong2024hunyuanvideo}
W.~Kong, Q.~Tian, Z.~Zhang, R.~Min, Z.~Dai, J.~Zhou, J.~Xiong, X.~Li, B.~Wu,
  J.~Zhang, et~al.
\newblock Hunyuanvideo: A systematic framework for large video generative
  models.
\newblock {\em arXiv preprint arXiv:2412.03603}, 2024.

\bibitem{lee2022autoregressive}
D.~Lee, C.~Kim, S.~Kim, M.~Cho, and W.-S. Han.
\newblock Autoregressive image generation using residual quantization.
\newblock In {\em CVPR}, 2022.

\bibitem{VLM:SAIL}
W.~Lei, J.~Wang, H.~Wang, X.~Li, J.~H. Liew, J.~Feng, and Z.~Huang.
\newblock The scalability of simplicity: Empirical analysis of vision-language
  learning with a single transformer.
\newblock In {\em Proceedings of the IEEE/CVF International Conference on
  Computer Vision}, pages 20758--20769, 2025.

\bibitem{REPA-E}
X.~Leng, J.~Singh, Y.~Hou, Z.~Xing, S.~Xie, and L.~Zheng.
\newblock Repa-e: Unlocking vae for end-to-end tuning of latent diffusion
  transformers.
\newblock In {\em Proceedings of the IEEE/CVF International Conference on
  Computer Vision}, pages 18262--18272, 2025.

\bibitem{li2026uevalbenchmarkunifiedmultimodal}
B.~Li, Y.~Yin, W.~Chai, X.~Fu, and Z.~Liu.
\newblock Ueval: A benchmark for unified multimodal generation.
\newblock {\em arXiv preprint arXiv:2601.22155}, 2026.

\bibitem{li2025ariaopenmultimodalnative}
D.~Li, Y.~Liu, H.~Wu, Y.~Wang, Z.~Shen, B.~Qu, X.~Niu, F.~Zhou, C.~Huang,
  Y.~Li, C.~Zhu, X.~Ren, C.~Li, Y.~Ye, P.~Liu, L.~Zhang, H.~Yan, G.~Wang,
  B.~Chen, and J.~Li.
\newblock Aria: An open multimodal native mixture-of-experts model.
\newblock {\em arXiv preprint arXiv:2410.05993}, 2024.

\bibitem{li2025onecat}
H.~Li, X.~Peng, Y.~Wang, Z.~Peng, X.~Chen, R.~Weng, J.~Wang, X.~Cai, W.~Dai,
  and H.~Xiong.
\newblock Onecat: Decoder-only auto-regressive model for unified understanding
  and generation.
\newblock {\em arXiv preprint arXiv:2509.03498}, 2025.

\bibitem{li2025synergen}
H.~Li, C.~Tian, J.~Shao, X.~Zhu, Z.~Wang, J.~Zhu, W.~Dou, X.~Wang, H.~Li,
  L.~Lu, et~al.
\newblock Synergen-vl: Towards synergistic image understanding and generation
  with vision experts and token folding.
\newblock In {\em Proceedings of the Computer Vision and Pattern Recognition
  Conference}, pages 29767--29779, 2025.

\bibitem{blip2}
J.~Li, D.~Li, S.~Savarese, and S.~Hoi.
\newblock Blip-2: Bootstrapping language-image pre-training with frozen image
  encoders and large language models.
\newblock In {\em International conference on machine learning}, pages
  19730--19742. PMLR, 2023.

\bibitem{li2025back}
T.~Li and K.~He.
\newblock Back to basics: Let denoising generative models denoise.
\newblock {\em arXiv preprint arXiv:2511.13720}, 2025.

\bibitem{VLM:BREEN}
T.~Li, Y.~Rao, W.~Hu, and Y.~Cheng.
\newblock Breen: bridge data-efficient encoder-free multimodal learning with
  learnable queries.
\newblock In {\em Proceedings of the IEEE/CVF Winter Conference on Applications
  of Computer Vision}, pages 5384--5395, 2026.

\bibitem{li2024videochat}
X.~Li, Y.~Wang, J.~Yu, X.~Zeng, Y.~Zhu, H.~Huang, J.~Gao, K.~Li, Y.~He,
  C.~Wang, et~al.
\newblock Videochat-flash: Hierarchical compression for long-context video
  modeling.
\newblock {\em arXiv preprint arXiv:2501.00574}, 2024.

\bibitem{li2025dual}
Z.~Li, H.~Li, Y.~Shi, A.~B. Farimani, Y.~Kluger, L.~Yang, and P.~Wang.
\newblock Dual diffusion for unified image generation and understanding.
\newblock In {\em Proceedings of the Computer Vision and Pattern Recognition
  Conference}, pages 2779--2790, 2025.

\bibitem{MoT}
W.~Liang, L.~Yu, L.~Luo, S.~Iyer, N.~Dong, C.~Zhou, G.~Ghosh, M.~Lewis, W.-t.
  Yih, L.~Zettlemoyer, et~al.
\newblock Mixture-of-transformers: A sparse and scalable architecture for
  multi-modal foundation models.
\newblock {\em arXiv preprint arXiv:2411.04996}, 2024.

\bibitem{liao2025mogao}
C.~Liao, L.~Liu, X.~Wang, Z.~Luo, X.~Zhang, W.~Zhao, J.~Wu, L.~Li, Z.~Tian, and
  W.~Huang.
\newblock Mogao: An omni foundation model for interleaved multi-modal
  generation.
\newblock {\em arXiv preprint arXiv:2505.05472}, 2025.

\bibitem{lin2025uniworld}
B.~Lin, Z.~Li, X.~Cheng, Y.~Niu, Y.~Ye, X.~He, S.~Yuan, W.~Yu, S.~Wang, Y.~Ge,
  et~al.
\newblock Uniworld: High-resolution semantic encoders for unified visual
  understanding and generation.
\newblock {\em arXiv preprint arXiv:2506.03147}, 2025.

\bibitem{lin2024videollava}
B.~Lin, Y.~Ye, B.~Zhu, J.~Cui, M.~Ning, P.~Jin, and L.~Yuan.
\newblock Video-llava: Learning united visual representation by alignment
  before projection.
\newblock In {\em EMNLP}, 2024.

\bibitem{TokLIP}
H.~Lin, T.~Wang, Y.~Ge, Y.~Ge, Z.~Lu, Y.~Wei, Q.~Zhang, Z.~Sun, and Y.~Shan.
\newblock Toklip: Marry visual tokens to clip for multimodal comprehension and
  generation.
\newblock {\em arXiv preprint arXiv:2505.05422}, 2025.

\bibitem{MOMA}
X.~V. Lin, A.~Shrivastava, L.~Luo, S.~Iyer, M.~Lewis, G.~Ghosh, L.~Zettlemoyer,
  and A.~Aghajanyan.
\newblock Moma: Efficient early-fusion pre-training with mixture of
  modality-aware experts.
\newblock {\em arXiv preprint arXiv:2407.21770}, 2024.

\bibitem{liu2023llava}
H.~Liu, C.~Li, Q.~Wu, and Y.~J. Lee.
\newblock Visual instruction tuning.
\newblock {\em Advances in Neural Information Processing Systems},
  36:34892--34916, 2023.

\bibitem{tuna2}
Z.~Liu, W.~Ren, X.~Huang, S.~Chen, T.~Li, M.~Chen, Y.~Ji, S.~He, J.~Schult,
  B.~Zeng, T.~Xiang, W.~Chen, P.~Luo, L.~Zettlemoyer, and Y.~Cong.
\newblock Tuna-2: Pixel embeddings beat vision encoders for multimodal
  understanding and generation.
\newblock {\em arXiv preprint arXiv:2604.24763}, 2026.

\bibitem{liu2025tuna}
Z.~Liu, W.~Ren, H.~Liu, Z.~Zhou, S.~Chen, H.~Qiu, X.~Huang, Z.~An, F.~Yang,
  A.~Patel, V.~Atliha, T.~Ng, X.~Han, C.~Zhu, C.~Zhang, D.~Liu, J.-M.
  Perez-Rua, S.~He, J.~Schmidhuber, W.~Chen, P.~Luo, W.~Liu, T.~Xiang,
  J.~Schult, and Y.~Cong.
\newblock Tuna: Taming unified visual representations for native unified
  multimodal models.
\newblock {\em arXiv preprint arXiv:2512.02014}, 2025.

\bibitem{io2}
J.~Lu, C.~Clark, S.~Lee, Z.~Zhang, S.~Khosla, R.~Marten, D.~Hoiem, and
  A.~Kembhavi.
\newblock Unified-io 2: Scaling autoregressive multimodal models with vision
  language audio and action.
\newblock In {\em CVPR}, 2024.

\bibitem{lu2022unifiediounifiedmodelvision}
J.~Lu, C.~Clark, R.~Zellers, R.~Mottaghi, and A.~Kembhavi.
\newblock Unified-io: A unified model for vision, language, and multi-modal
  tasks.
\newblock {\em arXiv preprint arXiv:2206.08916}, 2022.

\bibitem{VLM:Mono-InternVL-1.5}
G.~Luo, W.~Dou, W.~Li, Z.~Wang, X.~Yang, C.~Tian, H.~Li, W.~Wang, W.~Wang,
  X.~Zhu, et~al.
\newblock Mono-internvl-1.5: Towards cheaper and faster monolithic multimodal
  large language models.
\newblock {\em arXiv preprint arXiv:2507.12566}, 2025.

\bibitem{VLM:Mono-InternVL}
G.~Luo, X.~Yang, W.~Dou, Z.~Wang, J.~Liu, J.~Dai, Y.~Qiao, and X.~Zhu.
\newblock Mono-internvl: Pushing the boundaries of monolithic multimodal large
  language models with endogenous visual pre-training.
\newblock In {\em Proceedings of the IEEE/CVF Conference on Computer Vision and
  Pattern Recognition}, pages 24960--24971, 2025.

\bibitem{luo2025univid}
J.~Luo, J.~Lin, Z.~Zhang, B.~Wu, M.~Fang, L.~Chen, and H.~Tang.
\newblock Univid: The open-source unified video model.
\newblock {\em arXiv preprint arXiv:2509.24200}, 2025.

\bibitem{luo2025next}
R.~Luo, X.~Xia, L.~Wang, L.~Chen, R.~Shan, J.~Luo, M.~Yang, and T.-S. Chua.
\newblock Next-omni: Towards any-to-any omnimodal foundation models with
  discrete flow matching.
\newblock {\em arXiv preprint arXiv:2510.13721}, 2025.

\bibitem{ma2025unitok}
C.~Ma, Y.~Jiang, J.~Wu, J.~Yang, X.~Yu, Z.~Yuan, B.~Peng, and X.~Qi.
\newblock Unitok: A unified tokenizer for visual generation and understanding.
\newblock {\em arXiv preprint arXiv:2502.20321}, 2025.

\bibitem{ma2025genhancer}
S.~Ma, Y.~Ge, T.~Wang, Y.~Guo, Y.~Ge, and Y.~Shan.
\newblock Genhancer: Imperfect generative models are secretly strong
  vision-centric enhancers.
\newblock {\em arXiv preprint arXiv:2503.19480}, 2025.

\bibitem{ma2024janusflow}
Y.~Ma, X.~Liu, X.~Chen, W.~Liu, C.~Wu, Z.~Wu, Z.~Pan, Z.~Xie, H.~Zhang, X.~Yu,
  et~al.
\newblock Janusflow: Harmonizing autoregression and rectified flow for unified
  multimodal understanding and generation.
\newblock In {\em Proceedings of the IEEE/CVF Conference on Computer Vision and
  Pattern Recognition}, pages 7739--7751, 2025.

\bibitem{mentzer2023finite}
F.~Mentzer, D.~Minnen, E.~Agustsson, and M.~Tschannen.
\newblock Finite scalar quantization: Vq-vae made simple.
\newblock {\em arXiv preprint arXiv:2309.15505}, 2023.

\bibitem{nguyen2025oneflowconcurrentmixedmodalinterleaved}
J.~Nguyen, M.~Havasi, T.~Berrada, L.~Zettlemoyer, and R.~T.~Q. Chen.
\newblock Oneflow: Concurrent mixed-modal and interleaved generation with edit
  flows.
\newblock {\em arXiv preprint arXiv:2510.03506}, 2025.

\bibitem{niu2025does}
Y.~Niu, W.~Jin, J.~Liao, C.~Feng, P.~Jin, B.~Lin, Z.~Li, B.~Zhu, W.~Yu, and
  L.~Yuan.
\newblock Does understanding inform generation in unified multimodal models?
  from analysis to path forward.
\newblock {\em arXiv preprint arXiv:2511.20561}, 2025.

\bibitem{niu2025wise}
Y.~Niu, M.~Ning, M.~Zheng, W.~Jin, B.~Lin, P.~Jin, J.~Liao, C.~Feng, K.~Ning,
  B.~Zhu, et~al.
\newblock Wise: A world knowledge-informed semantic evaluation for
  text-to-image generation.
\newblock {\em arXiv preprint arXiv:2503.07265}, 2025.

\bibitem{nvidia2025cosmospredict2}
{NVIDIA}.
\newblock {Cosmos-Predict2}: World simulation model for physical {AI}.
\newblock \url{https://github.com/nvidia-cosmos/cosmos-predict2}, 2025.

\bibitem{agarwal2025cosmos}
{NVIDIA}.
\newblock {Cosmos} world foundation model platform for physical {AI}.
\newblock {\em arXiv preprint arXiv:2501.03575}, 2025.

\bibitem{nvidia2026cosmos3}
{NVIDIA}.
\newblock Cosmos 3: Omnimodal world models for physical ai.
\newblock {\em arXiv preprint arXiv:2606.02800}, 2026.

\bibitem{VQ_VAE}
A.~v.~d. Oord, O.~Vinyals, and K.~Kavukcuoglu.
\newblock Neural discrete representation learning.
\newblock {\em arXiv preprint arXiv:1711.00937}, 2017.

\bibitem{oquab2023dinov2}
M.~Oquab, T.~Darcet, T.~Moutakanni, H.~Vo, M.~Szafraniec, V.~Khalidov,
  P.~Fernandez, D.~Haziza, F.~Massa, A.~El-Nouby, et~al.
\newblock Dinov2: Learning robust visual features without supervision.
\newblock {\em arXiv preprint arXiv:2304.07193}, 2023.

\bibitem{pan2025transfer}
X.~Pan, S.~N. Shukla, A.~Singh, Z.~Zhao, S.~K. Mishra, J.~Wang, Z.~Xu, J.~Chen,
  K.~Li, F.~Juefei-Xu, et~al.
\newblock Transfer between modalities with metaqueries.
\newblock {\em arXiv preprint arXiv:2504.06256}, 2025.

\bibitem{DiT}
W.~Peebles and S.~Xie.
\newblock Scalable diffusion models with transformers.
\newblock In {\em Proc. ICCV}, 2023.

\bibitem{SDXL}
D.~Podell, Z.~English, K.~Lacey, A.~Blattmann, T.~Dockhorn, J.~M{\"u}ller,
  J.~Penna, and R.~Rombach.
\newblock Sdxl: Improving latent diffusion models for high-resolution image
  synthesis.
\newblock {\em arXiv preprint arXiv:2307.01952}, 2023.

\bibitem{MovieGen}
A.~Polyak, A.~Zohar, A.~Brown, A.~Tjandra, A.~Sinha, A.~Lee, A.~Vyas, B.~Shi,
  C.-Y. Ma, C.-Y. Chuang, et~al.
\newblock Movie gen: A cast of media foundation models.
\newblock {\em arXiv preprint arXiv:2410.13720}, 2024.

\bibitem{qu2025tokenflow}
L.~Qu, H.~Zhang, Y.~Liu, X.~Wang, Y.~Jiang, Y.~Gao, H.~Ye, D.~K. Du, Z.~Yuan,
  and X.~Wu.
\newblock Tokenflow: Unified image tokenizer for multimodal understanding and
  generation.
\newblock In {\em Proceedings of the Computer Vision and Pattern Recognition
  Conference}, pages 2545--2555, 2025.

\bibitem{radford2021learning}
A.~Radford, J.~W. Kim, C.~Hallacy, A.~Ramesh, G.~Goh, S.~Agarwal, G.~Sastry,
  A.~Askell, P.~Mishkin, J.~Clark, et~al.
\newblock Learning transferable visual models from natural language
  supervision.
\newblock In {\em Proc. ICML}, 2021.

\bibitem{razavi2019generatingdiversehighfidelityimages}
A.~Razavi, A.~van~den Oord, and O.~Vinyals.
\newblock Generating diverse high-fidelity images with vq-vae-2.
\newblock {\em arXiv preprint arXiv:1906.00446}, 2019.

\bibitem{LatentDiffusion}
R.~Rombach, A.~Blattmann, D.~Lorenz, P.~Esser, and B.~Ommer.
\newblock High-resolution image synthesis with latent diffusion models.
\newblock In {\em Proc. CVPR}, 2022.

\bibitem{sensenova2026neounify}
SenseNova.
\newblock Neo-unify: Building native multimodal unified models end to end,
  2026.

\bibitem{shen2025mammothmoda2}
T.~Shen, X.~Wan, T.~Chen, R.~Zhang, J.~Pan, D.~Lu, F.~Lei, Z.~Lu, Y.~Yang,
  C.~Cheng, et~al.
\newblock Mammothmoda2: A unified ar-diffusion framework for multimodal
  understanding and generation.
\newblock {\em arXiv preprint arXiv:2511.18262}, 2025.

\bibitem{shi2025latent}
M.~Shi, H.~Wang, W.~Zheng, Z.~Yuan, X.~Wu, X.~Wang, P.~Wan, J.~Zhou, and J.~Lu.
\newblock Latent diffusion model without variational autoencoder.
\newblock {\em arXiv preprint arXiv:2510.15301}, 2025.

\bibitem{shi2025muddit}
Q.~Shi, J.~Bai, Z.~Zhao, W.~Chai, K.~Yu, J.~Wu, S.~Song, Y.~Tong, X.~Li, X.~Li,
  et~al.
\newblock Muddit: Liberating generation beyond text-to-image with a unified
  discrete diffusion model.
\newblock {\em arXiv preprint arXiv:2505.23606}, 2025.

\bibitem{shi2024llamafusion}
W.~Shi, X.~Han, C.~Zhou, W.~Liang, X.~V. Lin, L.~Zettlemoyer, and L.~Yu.
\newblock Llamafusion: Adapting pretrained language models for multimodal
  generation.
\newblock {\em arXiv preprint arXiv:2412.15188}, 2024.

\bibitem{shi2025realunify}
Y.~Shi, Y.~Dong, Y.~Ding, Y.~Wang, X.~Zhu, S.~Zhou, W.~Liu, H.~Tian, R.~Wang,
  H.~Wang, et~al.
\newblock Realunify: Do unified models truly benefit from unification? a
  comprehensive benchmark.
\newblock {\em arXiv preprint arXiv:2509.24897}, 2025.

\bibitem{simeoni2025dinov3}
O.~Sim{\'e}oni, H.~V. Vo, M.~Seitzer, F.~Baldassarre, M.~Oquab, C.~Jose,
  V.~Khalidov, M.~Szafraniec, S.~Yi, M.~Ramamonjisoa, et~al.
\newblock Dinov3.
\newblock {\em arXiv preprint arXiv:2508.10104}, 2025.

\bibitem{Dualtoken}
W.~Song, Y.~Wang, Z.~Song, Y.~Li, H.~Sun, W.~Chen, Z.~Zhou, J.~Xu, J.~Wang, and
  K.~Yu.
\newblock Dualtoken: Towards unifying visual understanding and generation with
  dual visual vocabularies.
\newblock {\em arXiv preprint arXiv:2503.14324}, 2025.

\bibitem{emu2}
Q.~Sun, Y.~Cui, X.~Zhang, F.~Zhang, Q.~Yu, Y.~Wang, Y.~Rao, J.~Liu, T.~Huang,
  and X.~Wang.
\newblock Generative multimodal models are in-context learners.
\newblock In {\em Proceedings of the IEEE/CVF Conference on Computer Vision and
  Pattern Recognition}, pages 14398--14409, 2024.

\bibitem{sun2023eva}
Q.~Sun, Y.~Fang, L.~Wu, X.~Wang, and Y.~Cao.
\newblock Eva-clip: Improved training techniques for clip at scale.
\newblock {\em arXiv preprint arXiv:2303.15389}, 2023.

\bibitem{sun2024evaclip18bscalingclip18}
Q.~Sun, J.~Wang, Q.~Yu, Y.~Cui, F.~Zhang, X.~Zhang, and X.~Wang.
\newblock Eva-clip-18b: Scaling clip to 18 billion parameters.
\newblock {\em arXiv preprint arXiv:2402.04252}, 2024.

\bibitem{emu1}
Q.~Sun, Q.~Yu, Y.~Cui, F.~Zhang, X.~Zhang, Y.~Wang, H.~Gao, J.~Liu, T.~Huang,
  and X.~Wang.
\newblock Emu: Generative pretraining in multimodality.
\newblock In {\em ICLR}, 2024.

\bibitem{tan2025omni}
Z.~Tan, H.~Yang, L.~Qin, J.~Gong, M.~Yang, and H.~Li.
\newblock Omni-video: Democratizing unified video understanding and generation.
\newblock {\em arXiv preprint arXiv:2507.06119}, 2025.

\bibitem{tang2025unilip}
H.~Tang, C.~Xie, X.~Bao, T.~Weng, P.~Li, Y.~Zheng, and L.~Wang.
\newblock Unilip: Adapting clip for unified multimodal understanding,
  generation and editing.
\newblock {\em arXiv preprint arXiv:2507.23278}, 2025.

\bibitem{VLM:HoVLE}
C.~Tao, S.~Su, X.~Zhu, C.~Zhang, Z.~Chen, J.~Liu, W.~Wang, L.~Lu, G.~Huang,
  Y.~Qiao, et~al.
\newblock Hovle: Unleashing the power of monolithic vision-language models with
  holistic vision-language embedding.
\newblock In {\em Proceedings of the Computer Vision and Pattern Recognition
  Conference}, pages 14559--14569, 2025.

\bibitem{Chameleon}
C.~Team.
\newblock Chameleon: Mixed-modal early-fusion foundation models.
\newblock {\em arXiv preprint arXiv:2405.09818}, 2024.

\bibitem{team2026longcat}
M.~L. Team, B.~Xiao, C.~Wang, C.~Li, C.~Zhang, C.~Peng, H.~Yu, H.~Yang, H.~Yan,
  H.~Sun, et~al.
\newblock Longcat-next: Lexicalizing modalities as discrete tokens.
\newblock {\em arXiv preprint arXiv:2603.27538}, 2026.

\bibitem{flexattention}
P.~Team.
\newblock Flexattention: The flexibility of pytorch with the performance of
  flashattention.
\newblock {\em Pytorch Blog}, 2024.

\bibitem{Wan2.2}
W.-V. team.
\newblock Wan: Open and advanced large-scale video generative models (wan2.2),
  2025.
\newblock GitHub repository, Apache-2.0 License.

\bibitem{tian2026internvludemocratizingunifiedmultimodal}
C.~Tian, D.~Yang, G.~Chen, E.~Cui, Z.~Wang, Y.~Duan, P.~Yin, S.~Chen, G.~Yang,
  M.~Liu, et~al.
\newblock Internvl-u: Democratizing unified multimodal models for
  understanding, reasoning, generation and editing.
\newblock {\em arXiv preprint arXiv:2603.09877}, 2026.

\bibitem{tong2026beyond}
S.~Tong, D.~Fan, J.~Nguyen, E.~Brown, G.~Zhou, S.~Qian, B.~Zheng, T.~Vallaeys,
  J.~Han, R.~Fergus, N.~Murray, M.~Ghazvininejad, M.~Lewis, N.~Ballas, A.~Bar,
  M.~Rabbat, J.~Verbeek, L.~Zettlemoyer, K.~Sinha, Y.~LeCun, and S.~Xie.
\newblock Beyond language modeling: An exploration of multimodal pretraining.
\newblock {\em arXiv preprint arXiv:2603.03276}, 2026.

\bibitem{tong2024metamorph}
S.~Tong, D.~Fan, J.~Zhu, Y.~Xiong, X.~Chen, K.~Sinha, M.~Rabbat, Y.~LeCun,
  S.~Xie, and Z.~Liu.
\newblock Metamorph: Multimodal understanding and generation via instruction
  tuning.
\newblock {\em arXiv preprint arXiv:2412.14164}, 2024.

\bibitem{tong2026scalingtexttoimagediffusiontransformers}
S.~Tong, B.~Zheng, Z.~Wang, B.~Tang, N.~Ma, E.~Brown, J.~Yang, R.~Fergus,
  Y.~LeCun, and S.~Xie.
\newblock Scaling text-to-image diffusion transformers with representation
  autoencoders.
\newblock {\em arXiv preprint arXiv:2601.16208}, 2026.

\bibitem{tschannen2025siglip}
M.~Tschannen, A.~Gritsenko, X.~Wang, M.~F. Naeem, I.~Alabdulmohsin,
  N.~Parthasarathy, T.~Evans, L.~Beyer, Y.~Xia, B.~Mustafa, et~al.
\newblock Siglip 2: Multilingual vision-language encoders with improved
  semantic understanding, localization, and dense features.
\newblock {\em arXiv preprint arXiv:2502.14786}, 2025.

\bibitem{wan2025wan}
T.~Wan, A.~Wang, B.~Ai, B.~Wen, C.~Mao, C.-W. Xie, D.~Chen, F.~Yu, H.~Zhao,
  J.~Yang, et~al.
\newblock Wan: Open and advanced large-scale video generative models.
\newblock {\em arXiv preprint arXiv:2503.20314}, 2025.

\bibitem{wang2024illume}
C.~Wang, G.~Lu, J.~Yang, R.~Huang, J.~Han, L.~Hou, W.~Zhang, and H.~Xu.
\newblock Illume: Illuminating your llms to see, draw, and self-enhance.
\newblock {\em arXiv preprint arXiv:2412.06673}, 2024.

\bibitem{wang2025autoregressive}
D.~Wang, W.~Song, Y.~Wang, S.~Wang, K.~Yu, Z.~Wei, and J.~Wang.
\newblock Autoregressive semantic visual reconstruction helps vlms understand
  better.
\newblock {\em arXiv preprint arXiv:2506.09040}, 2025.

\bibitem{wang2025ovis}
G.-H. Wang, S.~Zhao, X.~Zhang, L.~Cao, P.~Zhan, L.~Duan, S.~Lu, M.~Fu, X.~Chen,
  J.~Zhao, et~al.
\newblock Ovis-u1 technical report.
\newblock {\em arXiv preprint arXiv:2506.23044}, 2025.

\bibitem{VLM:VoRA}
H.~Wang, Y.~Ye, B.~Li, Y.~Nie, J.~Lu, J.~Tang, Y.~Wang, and C.~Huang.
\newblock Vision as lora.
\newblock {\em arXiv preprint arXiv:2503.20680}, 2025.

\bibitem{wang2024reconstructive}
H.~Wang, A.~Zheng, Y.~Zhao, T.~Wang, Z.~Ge, X.~Zhang, and Z.~Zhang.
\newblock Reconstructive visual instruction tuning.
\newblock {\em arXiv preprint arXiv:2410.09575}, 2024.

\bibitem{wang2025pixnerd}
S.~Wang, Z.~Gao, C.~Zhu, W.~Huang, and L.~Wang.
\newblock Pixnerd: Pixel neural field diffusion.
\newblock {\em arXiv preprint arXiv:2507.23268}, 2025.

\bibitem{wang2024emu3}
X.~Wang, X.~Zhang, Z.~Luo, Q.~Sun, Y.~Cui, J.~Wang, F.~Zhang, Y.~Wang, Z.~Li,
  Q.~Yu, et~al.
\newblock Emu3: Next-token prediction is all you need.
\newblock {\em arXiv preprint arXiv:2409.18869}, 2024.

\bibitem{wang2025hbridge}
X.~Wang, Z.~Zhang, H.~Zhang, Z.~Lin, Y.~Zhou, Q.~Liu, S.~Zhang, Y.~Li, S.~Liu,
  H.~Zheng, et~al.
\newblock Hbridge: H-shape bridging of heterogeneous experts for unified
  multimodal understanding and generation.
\newblock {\em arXiv preprint arXiv:2511.20520}, 2025.

\bibitem{minit2i2026}
X.~Wang, H.~Zhao, Y.~Lu, K.~Zhou, L.~Ma, and K.~He.
\newblock {MiniT2I}: A minimalist baseline for text-to-image generation, 2026.

\bibitem{wang2025lightfusion}
Z.~Wang, Z.~Chen, C.~Gou, F.~Li, C.~Deng, D.~Zhu, K.~Li, W.~Yu, H.~Tu, H.~Fan,
  et~al.
\newblock Lightfusion: A light-weighted, double fusion framework for unified
  multimodal understanding and generation.
\newblock {\em arXiv preprint arXiv:2510.22946}, 2025.

\bibitem{wei2025univideo}
C.~Wei, Q.~Liu, Z.~Ye, Q.~Wang, X.~Wang, P.~Wan, K.~Gai, and W.~Chen.
\newblock Univideo: Unified understanding, generation, and editing for videos.
\newblock {\em arXiv preprint arXiv:2510.08377}, 2025.

\bibitem{wiedmann2025finevision}
L.~Wiedmann, O.~Zohar, A.~Mahla, X.~Wang, R.~Li, T.~Frere, L.~von Werra, A.~R.
  Gosthipaty, and A.~Marafioti.
\newblock Finevision: Open data is all you need.
\newblock {\em arXiv preprint arXiv:2510.17269}, 2025.

\bibitem{wu2024janus}
C.~Wu, X.~Chen, Z.~Wu, Y.~Ma, X.~Liu, Z.~Pan, W.~Liu, Z.~Xie, X.~Yu, C.~Ruan,
  et~al.
\newblock Janus: Decoupling visual encoding for unified multimodal
  understanding and generation.
\newblock {\em arXiv preprint arXiv:2410.13848}, 2024.

\bibitem{wu2024liquid}
J.~Wu, Y.~Jiang, C.~Ma, Y.~Liu, H.~Zhao, Z.~Yuan, S.~Bai, and X.~Bai.
\newblock Liquid: Language models are scalable multi-modal generators.
\newblock {\em arXiv e-prints}, pages arXiv--2412, 2024.

\bibitem{wunext}
S.~Wu, H.~Fei, L.~Qu, W.~Ji, and T.-S. Chua.
\newblock {NExT-GPT}: {Any-to-Any Multimodal LLM}.
\newblock In {\em Proceedings of the International Conference on Machine
  Learning}, 2024.

\bibitem{wu2025openuni}
S.~Wu, Z.~Wu, Z.~Gong, Q.~Tao, S.~Jin, Q.~Li, W.~Li, and C.~C. Loy.
\newblock Openuni: A simple baseline for unified multimodal understanding and
  generation.
\newblock {\em arXiv preprint arXiv:2505.23661}, 2025.

\bibitem{wu2025harmonizing}
S.~Wu, W.~Zhang, L.~Xu, S.~Jin, Z.~Wu, Q.~Tao, W.~Liu, W.~Li, and C.~C. Loy.
\newblock Harmonizing visual representations for unified multimodal
  understanding and generation.
\newblock {\em arXiv preprint arXiv:2503.21979}, 2025.

\bibitem{wu2024vila}
Y.~Wu, Z.~Zhang, J.~Chen, H.~Tang, D.~Li, Y.~Fang, L.~Zhu, E.~Xie, H.~Yin,
  L.~Yi, et~al.
\newblock Vila-u: a unified foundation model integrating visual understanding
  and generation.
\newblock {\em arXiv preprint arXiv:2409.04429}, 2024.

\bibitem{xiao2025mindomniunleashingreasoninggeneration}
Y.~Xiao, L.~Song, Y.~Chen, Y.~Luo, Y.~Chen, Y.~Gan, W.~Huang, X.~Li, X.~Qi, and
  Y.~Shan.
\newblock Mindomni: Unleashing reasoning generation in vision language models
  with rgpo.
\newblock {\em arXiv preprint arXiv:2505.13031}, 2025.

\bibitem{xie2025sana15efficientscaling}
E.~Xie, J.~Chen, Y.~Zhao, J.~Yu, L.~Zhu, C.~Wu, Y.~Lin, Z.~Zhang, M.~Li,
  J.~Chen, H.~Cai, B.~Liu, D.~Zhou, and S.~Han.
\newblock Sana 1.5: Efficient scaling of training-time and inference-time
  compute in linear diffusion transformer.
\newblock {\em arXiv preprint arXiv:2501.18427}, 2025.

\bibitem{xie2025reconstruction}
J.~Xie, T.~Darrell, L.~Zettlemoyer, and X.~Wang.
\newblock Reconstruction alignment improves unified multimodal models.
\newblock {\em arXiv preprint arXiv:2509.07295}, 2025.

\bibitem{xie2024show}
J.~Xie, W.~Mao, Z.~Bai, D.~J. Zhang, W.~Wang, K.~Q. Lin, Y.~Gu, Z.~Chen,
  Z.~Yang, and M.~Z. Shou.
\newblock Show-o: One single transformer to unify multimodal understanding and
  generation.
\newblock {\em arXiv preprint arXiv:2408.12528}, 2024.

\bibitem{xie2025show}
J.~Xie, Z.~Yang, and M.~Z. Shou.
\newblock Show-o2: Improved native unified multimodal models.
\newblock {\em arXiv preprint arXiv:2506.15564}, 2025.

\bibitem{xie2024muse}
R.~Xie, C.~Du, P.~Song, and C.~Liu.
\newblock Muse-vl: Modeling unified vlm through semantic discrete encoding.
\newblock {\em arXiv preprint arXiv:2411.17762}, 2024.

\bibitem{xin2025lumina}
Y.~Xin, Q.~Qin, S.~Luo, K.~Zhu, J.~Yan, Y.~Tai, J.~Lei, Y.~Cao, K.~Wang,
  Y.~Wang, et~al.
\newblock Lumina-dimoo: An omni diffusion large language model for multi-modal
  generation and understanding.
\newblock {\em arXiv preprint arXiv:2510.06308}, 2025.

\bibitem{xu2025demystifyingclipdata}
H.~Xu, S.~Xie, X.~E. Tan, P.-Y. Huang, R.~Howes, V.~Sharma, S.-W. Li, G.~Ghosh,
  L.~Zettlemoyer, and C.~Feichtenhofer.
\newblock Demystifying clip data.
\newblock {\em arXiv preprint arXiv:2309.16671}, 2023.

\bibitem{xu2025tbacuniimageunifiedunderstandinggeneration}
J.~Xu, Y.~Yin, and X.~Chen.
\newblock Tbac-uniimage: Unified understanding and generation by ladder-side
  diffusion tuning.
\newblock {\em arXiv preprint arXiv:2508.08098}, 2025.

\bibitem{xu2025nextembeddingpredictionmakesstrong}
S.~Xu, Z.~Ma, W.~Chai, X.~Chen, W.~Jin, J.~Chai, S.~Xie, and S.~X. Yu.
\newblock Next-embedding prediction makes strong vision learners.
\newblock {\em arXiv preprint arXiv:2512.16922}, 2025.

\bibitem{yang2025qwen3}
A.~Yang, A.~Li, B.~Yang, B.~Zhang, B.~Hui, B.~Zheng, B.~Yu, C.~Gao, C.~Huang,
  C.~Lv, et~al.
\newblock Qwen3 technical report.
\newblock {\em arXiv preprint arXiv:2505.09388}, 2025.

\bibitem{yang2026ureasonbenchmarkingreasoningtogenerationalignment}
C.~Yang, C.~Shi, B.~Shui, Y.~Wu, M.~Tao, H.~Wang, I.~Y. Lee, Y.~Liu, X.~Ma, and
  T.~Berg-Kirkpatrick.
\newblock Ureason: Benchmarking reasoning-to-generation alignment in unified
  multimodal models.
\newblock {\em arXiv preprint arXiv:2602.08336}, 2026.

\bibitem{yang2025latent}
J.~Yang, T.~Li, L.~Fan, Y.~Tian, and Y.~Wang.
\newblock Latent denoising makes good tokenizers.
\newblock In {\em The Fourteenth International Conference on Learning
  Representations}, 2026.

\bibitem{yang2025mmada}
L.~Yang, Y.~Tian, B.~Li, X.~Zhang, K.~Shen, Y.~Tong, and M.~Wang.
\newblock Mmada: Multimodal large diffusion language models.
\newblock {\em arXiv preprint arXiv:2505.15809}, 2025.

\bibitem{VLM:HaploVL}
R.~Yang, L.~Song, Y.~Xiao, R.~Huang, Y.~Ge, Y.~Shan, and H.~Zhao.
\newblock Haplovl: A single-transformer baseline for multi-modal understanding.
\newblock {\em arXiv preprint arXiv:2503.14694}, 2025.

\bibitem{CogVideoX}
Z.~Yang, J.~Teng, W.~Zheng, M.~Ding, S.~Huang, J.~Xu, Y.~Yang, W.~Hong,
  X.~Zhang, G.~Feng, et~al.
\newblock Cogvideox: Text-to-video diffusion models with an expert transformer.
\newblock {\em arXiv preprint arXiv:2408.06072}, 2024.

\bibitem{vavae}
J.~Yao, B.~Yang, and X.~Wang.
\newblock Reconstruction vs. generation: Taming optimization dilemma in latent
  diffusion models.
\newblock In {\em Proceedings of the Computer Vision and Pattern Recognition
  Conference}, pages 15703--15712, 2025.

\bibitem{you2026llada}
Z.~You, X.~Zhang, J.~Zhou, C.~Li, and J.-R. Wen.
\newblock Llada-o: An effective and length-adaptive omni diffusion model.
\newblock {\em arXiv preprint arXiv:2603.01068}, 2026.

\bibitem{yu2022vectorquantized}
J.~Yu, X.~Li, J.~Y. Koh, H.~Zhang, R.~Pang, J.~Qin, A.~Ku, Y.~Xu, J.~Baldridge,
  and Y.~Wu.
\newblock Vector-quantized image modeling with improved {VQGAN}.
\newblock In {\em ICLR}, 2022.

\bibitem{yu2024representation}
S.~Yu, S.~Kwak, H.~Jang, J.~Jeong, J.~Huang, J.~Shin, and S.~Xie.
\newblock Representation alignment for generation: Training diffusion
  transformers is easier than you think.
\newblock {\em arXiv preprint arXiv:2410.06940}, 2024.

\bibitem{yu2025pixeldit}
Y.~Yu, W.~Xiong, W.~Nie, Y.~Sheng, S.~Liu, and J.~Luo.
\newblock Pixeldit: Pixel diffusion transformers for image generation.
\newblock {\em arXiv preprint arXiv:2511.20645}, 2025.

\bibitem{yue2025uniflow}
Z.~Yue, H.~Zhang, X.~Zeng, B.~Chen, C.~Wang, S.~Zhuang, L.~Dong, K.~Du,
  Y.~Wang, L.~Wang, et~al.
\newblock Uniflow: A unified pixel flow tokenizer for visual understanding and
  generation.
\newblock {\em arXiv preprint arXiv:2510.10575}, 2025.

\bibitem{VLP:SigLIP}
X.~Zhai, B.~Mustafa, A.~Kolesnikov, and L.~Beyer.
\newblock Sigmoid loss for language image pre-training.
\newblock In {\em Proceedings of the IEEE/CVF international conference on
  computer vision}, pages 11975--11986, 2023.

\bibitem{zhang2026nextflow}
H.~Zhang, L.~Qu, Y.~Liu, H.~Chen, Y.~Song, Y.~Dong, S.~Sun, X.~Li, X.~Wang,
  Y.~Jiang, et~al.
\newblock Nextflow: Unified sequential modeling activates multimodal
  understanding and generation.
\newblock {\em arXiv preprint arXiv:2601.02204}, 2026.

\bibitem{zhang2026crosstaskgeneralizationunderstandinggeneration}
J.~Zhang, T.~Li, L.~Li, Z.~Yang, and Y.~Cheng.
\newblock Cross-task generalization between understanding and generation in
  unified vision-language models: A controlled study.
\newblock {\em arXiv preprint arXiv:2505.23043}, 2025.

\bibitem{zhang2026openvision}
L.~Zhang, S.~Ren, Y.~Liu, X.~Li, Z.~Wang, Y.~Zhou, H.~Yao, Z.~Zheng, W.~Nie,
  G.~Liu, et~al.
\newblock Openvision 3: A family of unified visual encoder for both
  understanding and generation.
\newblock {\em arXiv preprint arXiv:2601.15369}, 2026.

\bibitem{zhang2024video}
Y.~Zhang, J.~Wu, W.~Li, B.~Li, Z.~Ma, Z.~Liu, and C.~Li.
\newblock Video instruction tuning with synthetic data.
\newblock {\em arXiv preprint arXiv:2410.02713}, 2024.

\bibitem{QLIP}
Y.~Zhao, F.~Xue, S.~Reed, L.~Fan, Y.~Zhu, J.~Kautz, Z.~Yu,
  P.~Kr{\"a}henb{\"u}hl, and D.-A. Huang.
\newblock Qlip: Text-aligned visual tokenization unifies auto-regressive
  multimodal understanding and generation.
\newblock {\em arXiv preprint arXiv:2502.05178}, 2025.

\bibitem{zheng2025diffusion}
B.~Zheng, N.~Ma, S.~Tong, and S.~Xie.
\newblock Diffusion transformers with representation autoencoders.
\newblock {\em arXiv preprint arXiv:2510.11690}, 2025.

\bibitem{zheng2023minigpt}
K.~Zheng, X.~He, and X.~E. Wang.
\newblock {MiniGPT-5}: Interleaved vision-and-language generation via
  generative vokens.
\newblock {\em arXiv preprint arXiv:2310.02239}, 2023.

\bibitem{zhou2024transfusion}
C.~Zhou, L.~Yu, A.~Babu, K.~Tirumala, M.~Yasunaga, L.~Shamis, J.~Kahn, X.~Ma,
  L.~Zettlemoyer, and O.~Levy.
\newblock Transfusion: Predict the next token and diffuse images with one
  multi-modal model.
\newblock {\em arXiv preprint arXiv:2408.11039}, 2024.

\end{thebibliography}
\end{document}